\documentclass[conference]{IEEEtran}
\IEEEoverridecommandlockouts
\usepackage{cite}
\usepackage{amsmath,amssymb,amsfonts}
\usepackage{algorithmic}
\usepackage{graphicx}
\usepackage{textcomp}
\usepackage{xcolor}
\def\BibTeX{{\rm B\kern-.05em{\sc i\kern-.025em b}\kern-.08em
    T\kern-.1667em\lower.7ex\hbox{E}\kern-.125emX}}

\usepackage[utf8]{inputenc} % allow utf-8 input
\usepackage[T1]{fontenc}    % use 8-bit T1 fonts
\usepackage{booktabs}       % professional-quality tables
\usepackage{amsmath,amssymb,amsfonts,amstext,amsthm,mathrsfs}       % blackboard math symbols
\usepackage{nicefrac}       % compact symbols for 1/2, etc.
\usepackage{microtype}      % microtypography
\usepackage{graphicx,enumitem}
\usepackage{epstopdf}
\usepackage{cleveref}
\usepackage{algorithmic,algorithm}
\usepackage[algo2e]{algorithm2e}

\newtheorem{theorem}{Theorem}
\newtheorem{definition}{Definition}

\usepackage{xcolor}

\newcommand{\inner}[2]{\langle #1, #2 \rangle}
\allowdisplaybreaks  %Allow equations to cross multiple pages
\begin{document}

\title{Neural Network Training Techniques Regularize Optimization Trajectory: An Empirical Study\\
%\thanks{}
}

\author{\IEEEauthorblockN{Cheng Chen}
\IEEEauthorblockA{\textit{Department of ECE} \\
\textit{University of Utah}\\
%Salt lake city, US \\
u0952128@utah.edu}
\and
\IEEEauthorblockN{Junjie Yang}
\IEEEauthorblockA{\textit{Department of ECE} \\
\textit{The Ohio State University}\\
%Columbus, US \\
yang.4972@osu.edu}
\and
\IEEEauthorblockN{Yi Zhou}
\IEEEauthorblockA{\textit{Department of ECE} \\
\textit{University of Utah}\\
%Salt lake city, US \\
yi.zhou@utah.edu}
}
%\and
%\and
%\and
%email address}\left( 

\maketitle

\begin{abstract}
	Modern deep neural network (DNN) trainings utilize various training techniques, e.g., nonlinear activation functions, batch normalization, skip-connections, etc. Despite their effectiveness, it is still mysterious how they help accelerate DNN trainings in practice. In this paper, we provide an empirical study of the regularization effect of these training techniques on DNN optimization. Specifically, we find that the optimization trajectories of successful DNN trainings consistently obey a certain regularity principle that regularizes the model update direction to be aligned with the trajectory direction. Theoretically, we show that such a regularity principle leads to a convergence guarantee in nonconvex optimization and the convergence rate depends on a regularization parameter. 
	Empirically, we find that DNN trainings that apply the training techniques achieve a fast convergence and obey the regularity principle with a large regularization parameter, implying that the model updates are well aligned with the trajectory. On the other hand, DNN trainings without the training techniques have slow convergence and obey the regularity principle with a small regularization parameter, implying that the model updates are not well aligned with the trajectory. Therefore, different training techniques regularize the model update direction via the regularity principle to facilitate the convergence.
	
	%Therefore, the training techniques impose a regularization on the optimization trajectory of DNN training via the regularity principle.
	%for nonconvex optimization with over-parameterized models, we propose a regularity principle that regularizes the optimization trajectory with a regularization parameter $\gamma$. We show that optimization trajectories that obey the regularity principle with a larger $\gamma$ achieve faster convergence. Through extensive experiments, we observe that practical DNN optimization trajectories consistently obey the regularity principle. In specific, DNN trainings that apply the training techniques achieve an accelerated convergence and obey the regularity principle with a large $\gamma$, whereas DNN trainings without the training techniques have slow convergence and obey the regularity principle with a small $\gamma$. Therefore, the training techniques impose a regularization on the optimization trajectory of DNN training via the regularity principle.
\end{abstract}

\begin{IEEEkeywords}
	Neural network, training techniques, nonconvex optimization, optimization trajectories, regularity principle 
\end{IEEEkeywords}

\section{Introduction}

Deep learning has been successfully applied to various domains such as computer vision, natural language processing, etc, and has achieved state-of-art performance in solving challenging tasks. Although deep neural networks (DNNs) have been well-known for decades to have great expressive power \cite{Cybenko1989}, the empirical success of training DNNs postponed to recent years when sufficient computation power is accessible and effective DNN {\em training techniques} are developed. 

The milestone developments of DNN training techniques can be roughly divided into two categories. First, various techniques have been developed at different levels of {\em neural network design}. Specifically, at the neuron level, various functions have been applied to activate the neurons, e.g., sigmoid function, hyperbolic tangent (tanh) function and the more popular rectified linear unit (ReLU) function \cite{Nair2010}. At the layer level, batch normalization (BN) has been widely applied to the hidden layers of DNNs to stabilize the training \cite{Ioffe2015}. Moreover, at the architecture level, skip-connections have been introduced to enable successful training of deep networks \cite{He2015,Huang2016,Srivastava2015,Szegedy2015}. Second, various efficient stochastic  {\em optimization algorithms} have been developed for DNN training, e.g., stochastic gradient descent (SGD) \cite{robbins1951,Rumelhart1986}, SGD with momentum \cite{Nesterov2014,Qian1999} and Adam \cite{kingma2015}, etc. \Cref{table: 1} provides a summary of these important DNN training techniques.

\begin{table}[H]
	\setlength{\tabcolsep}{4pt}
		\vspace{-2mm}
	\caption{Summary of DNN training techniques}\label{table: 1}
	\vspace{-2mm}
	\center
	\begin{tabular}{ccccc}
		\toprule
		\begin{tabular}{@{}c@{}} Neuron \\ activation \end{tabular}  
		& \begin{tabular}{@{}c@{}} Layer \\ normalization\end{tabular} 
		& \begin{tabular}{@{}c@{}} Network \\ architecture \end{tabular} 
		& \begin{tabular}{@{}c@{}} Optimization \\ algorithm \end{tabular} \\ \midrule
		\begin{tabular}{@{}c@{}} sigmoid, \\ tanh, \\ ReLU \end{tabular}
		& \begin{tabular}{@{}c@{}} Batch\\normalization \end{tabular}   
		& \begin{tabular}{@{}c@{}} Skip-\\connection \end{tabular}
		& \begin{tabular}{@{}c@{}} SGD,\\ SGD-\\momemtum, \\ Adam \end{tabular}\\
		\bottomrule
	\end{tabular}
\vspace{-2mm}
\end{table}
\vspace{-2mm}
%However, it is difficult to train DNNs with these activation functions as they may saturate and cause the vanishing gradient problem. Instead, rectified linear unit (ReLU) function has been a popular choice for activating the neurons due to its simplicity and robustness to the vanishing gradients \cite{Nair2010,glorot11a}. 

Although these training techniques have been widely applied in practical DNN training, there is limited understanding of how they help facilitate the training to achieve a global minimum of the network. In the existing literature, it is known that the sigmoid and tanh activation functions can cause the vanishing gradient problem, and the ReLU activation function is a popular replacement that avoids this issue \cite{glorot11a,Nair2010}. On the other hand, the batch normalization was originally proposed to reduce the internal covariance shift \cite{Ioffe2015}, and more recent studies show that it allows to use a large learning rate \cite{Bjorck2018} and improves the loss landscape \cite{Santurkar2018}. The skip-connection has been shown to help eliminate singularities and degeneracies \cite{orhan2018} and improve the loss landscape \cite{Hardt2016}. Moreover, regarding the optimization algorithm, the momentum scheme has been well-known to accelerate convex optimization \cite{Nesterov2014} and is also widely applied to accelerate nonconvex optimization \cite{Ghadimi2016}, whereas the Adam algorithm normalizes the update in each dimension to accelerate deep learning optimization \cite{kingma2015}. While these existing studies provide partial explanations to the effectiveness of DNN training techniques, their reasonings are from very different perspectives and are lack of a principled understanding. In particular, these studies do not fully explain why these diverse types of DNN training techniques can facilitate the training in practice. %and they do not quantify the effectiveness of the training techniques in accelerating DNN training.

%several points still remain unclear. First, these explainations do not guarantee DNN training to achieve global minimum, which is often observed in practice. Second, the explainations do not quantify the improvement of these techniques in DNN training. Moreover, it is unclear what optimization principle does the training techniques follow to facilitate the optimization process. 

In this paper, we take a step toward understanding DNN training techniques by providing a systematic empirical study of their regularization effect in the perspective of optimization. We empirically show that the training techniques regularize the optimization trajectory \footnote{Optimization trajectory is the sequence of model parameters generated in the iterative DNN training process.} in practical DNN training following a regularity principle, which quantifies the regularization effect that further determines the convergence rate in nonconvex optimization. We summarize our contributions as follows.

%attempts to propose an optimization principle that characterizes and quantifies the impacts of various training techniques on DNN training. The proposed principle is applicable to general stochastic algorithms in the nonconvex and over-parameterized regime and leads to guaranteed convergence to a global minimum. 

%\vspace{-1mm}
\subsection{Our Contributions}
%\vspace{-1mm}
We first propose and study a regularity principle (see \Cref{def: principle}) for stochastic algorithms in nonconvex optimization. 
 The regularity principle regularizes the model updates to be aligned with the optimization trajectory of the stochastic optimization process and is parameterized by a regularization parameter $\gamma>0$.  Intuitively, an optimization trajectory satisfying the regularity principle with a large parameter $\gamma$ implies that the model updates are well aligned with the trajectory direction and is close to a straight line, and hence the convergence is fast along this short trajectory.
 Theoretically, we show that the regularity principle with parameter $\gamma$ guarantees the optimization trajectory to converge at a sublinear rate $\mathcal{O}(1/\gamma T)$, which is faster for a larger parameter $\gamma$ of the regularity principle. 
	
 We conduct extensive experiments to examine the validity of the regularity principle in training popular deep models such as AlexNet, VGG, ResNet and U-Net. In specific, we find that all the optimization trajectories in the tested DNN trainings obey the regularity principle throughout the training process. Moreover, we explore the impacts of the training techniques that are listed in \Cref{table: 1} on the regularization parameter $\gamma$ of the regularity principle. We find that DNN trainings that apply the training techniques achieve a fast convergence and the corresponding optimization trajectories obey the regularity principle with a large $\gamma$, implying that the model updates are well aligned with the trajectory and facilitate the convergence. As a comparison, DNN trainings without the training techniques converge slowly and their optimization trajectories obey the regularity principle with a small $\gamma$, implying that the model updates are not well aligned with the trajectory and slows down the convergence. These observations are consistent with the theoretical implication of the regularity principle, in which a larger $\gamma$ leads to faster convergence. Therefore, the regularity principle captures and quantifies the regularization effect of these training techniques on DNN training and sheds light on the underlying mechanism that leads to successful DNN training.

%Moreover, in all our experiments, the sequence of model parameters generated along the optimization path converges to the global minimum with a monotonically diminishing distance, which is consistent with the theoretical implications of our proposed principle. 

%\vspace{-1mm}
\subsection{Related Work}
%\vspace{-1mm}
\textbf{DNN training techniques:} Various training techniques have been developed for DNN training. Examples include piece-wise linear activation functions, e.g., ReLU \cite{Nair2010}, ELU \cite{Clevert2015}, leaky ReLU \cite{Maas13}, batch normalization \cite{Ioffe2015}, skip-connection \cite{He2015,Srivastava2015,Szegedy2015} and advanced optimizers such as SGD with momentum \cite{Qian1999}, Adagrad \cite{Duchi2011}, Adam \cite{kingma2015}, AMSgrad \cite{Reddi2018}. The ReLU activation function and skip connection have been shown to help avoid the vanishing gradient problem and improve the loss landscape \cite{Hardt2016,zhang2019,Zhou2017,Zou2018}. The batch normalization has been shown to help avoid the internal covariance shift problem \cite{Ioffe2015}. More recent studies show that batch normalization allows to adopt a large learning rate \cite{Bjorck2018} and improves the loss landscape in the training \cite{Santurkar2018}. The convergence properties of the advanced optimizers have been studied in nonconvex optimization \cite{chen2018}.

\textbf{Optimization properties of nonconvex ML:}
Many nonconvex ML models have amenable properties for accelerating the optimization. For example, nonconvex problems such as phase retrieval \cite{zhang_2018}, low-rank matrix recovery, blind deconvolution \cite{Li_2018} and neural network sensing \cite{zhong_2017} satisfy the local regularity geometry around the global minimum \cite{Li_2018,zhong_2017,Zhou2017,Zhou2016b}, which guarantees the linear convergence of gradient-based algorithms. Recently, DNN trainings that use SGD have been shown to follow a star-convex optimization path \cite{zhou2018sgd}. Moreover, there are many works that study the convergence of gradient methods in training neural networks, e.g., \cite{arora2018a,du19c,Li2017}

%On the algorithm side, SGD has been shown to converge to a stationary point in nonconvex optimization \cite{Bottou_2016,Ghadimi_2016}. Recently studies show that SGD has the capability to escape strict saddle points \cite{Ge_2015,Chi_2017,Reddi_2018,Daneshmand_2018}.
%which is also achieved by cubic-regularization methods \cite{Nesterov2006,Zhou2018b,Wang2018}.  

%\vspace{-1mm}
\section{Regularity Principle for Nonconvex Optimization}
%\vspace{-1mm}
In this section, we introduce and study a regularity principle that regularizes the optimization trajectory of stochastic algorithms in nonconvex optimization. We show empirically in the subsequent sections that practical DNN trainings obey this principle.

%\vspace{-1mm}
\subsection{Optimization Trajectory of Stochastic Algorithm}
%\vspace{-1mm}
The goal of a machine learning task is to search for a good ML model $\theta$ that minimizes the total loss $f$ on a set of training data samples $\mathcal{Z}:\{z_i\}_{i=1}^n$. The problem is formally written as
\begin{align*}
	\min_{\theta\in \mathbb{R}^d}~f(\theta;\mathcal{Z}):=\frac{1}{n}\sum_{i=1}^{n}\ell(\theta;z_i), \tag{P}
\end{align*}
where $\ell(\cdot;z_i):\mathbb{R}^d\to \mathbb{R}$ corresponds to the loss on the $i$-th data sample $z_i$. 
Consider a generic stochastic algorithm (SA) that is initialized with certain model $\theta_0$. In each iteration $k$, the SA samples a data sample $z_{\xi_k}\in \mathcal{Z}$, where $\xi_k \in \{1,...,n\}$ is obtained via random sampling with reshuffle. Based on the current model $\theta_k$ and the sampled data $z_{\xi_k}$, the SA generates a stochastic update $U(\theta_k; z_{\xi_k})$ and applies it to update the model with a learning rate $\eta>0$ according to the update rule
\begin{align}
\text{(SA)}:~~	\theta_{k+1} = \theta_k - \eta U(\theta_{k}; z_{\xi_k}),~ k=0,1,2,...\label{eq: update}
\end{align}
\Cref{eq: update} covers the update rule of many existing optimizers for DNN training. For example, the stochastic gradient descent (SGD) algorithm chooses the update $U(\theta_{k}; z_{\xi_k})$ to be the stochastic gradient $\nabla \ell(\theta_k;z_{\xi_k})$. In comparison, the SGD with momentum algorithm generates the update using an extra momentum step, and the Adam algorithm generates the update as a moving average of the stochastic gradients normalized by the moving average of their second moments.  We formally define the optimization trajectory of SA as follows.
\begin{definition}[Optimization trajectory]
	The optimization trajectory of SA is the generated sequence of model parameters $\{\theta_k \}_k$ in the stochastic optimization process.
\end{definition}
We note that the stochastic optimization trajectory generally depends on the specific update rule $U$, which is affected by the training techniques. For example, neuron activation functions, batch normalization and skip-connections affect the back-propagation process in the computation of the model update $U$. On the other hand, optimization algorithms such as SGD with momentum and Adam specify the model update $U$ in different forms. Therefore, in order to understand the impact of these training techniques on optimization, it is natural to understand {\em how the training techniques affect the optimization trajectory}.

%\vspace{-1mm}
\subsection{Regularity Principle for Optimization Trajectory}
%\vspace{-1mm}

We next propose a regularity principle for the optimization trajectory generated by SA. %As we show later, the principle guarantees the SA to achieve the global minimum of the problem (P) is obeyed by practical DNN trainings.

\begin{definition}[Regularity principle for SA]\label{def: principle}
	Apply SA to solve the problem (P) for $T$ iterations and generate an optimization trajectory $\{\theta_0, \theta_1, ..., \theta_T \}$. We say that the trajectory satisfies the regularity principle with parameter $\gamma>0$ if for all $k=0,1,...T-1$,
	\begin{align}
		\langle \theta_k-\theta_T, U(\theta_{k};z_{\xi_k}) \rangle &\ge  \frac{\eta}{2}\|U(\theta_{k};z_{\xi_k})\|^2 \nonumber\\
		&\quad+ \gamma \big(\ell(\theta_k;z_{\xi_k}) - \inf_{\theta\in \mathbb{R}^d} \ell(\theta) \big). \label{eq: principle}
	\end{align}
\end{definition}

\begin{figure}[htbp]%[bth]
%	\vspace{-2mm}
	\centering
	\includegraphics[width=0.18\textwidth]{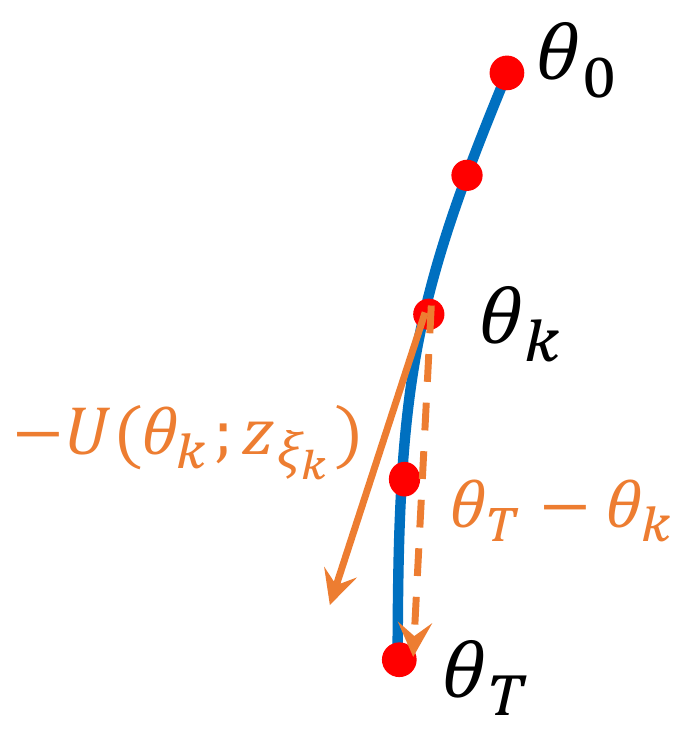}
	\hspace{30pt}
	\includegraphics[width=0.22\textwidth]{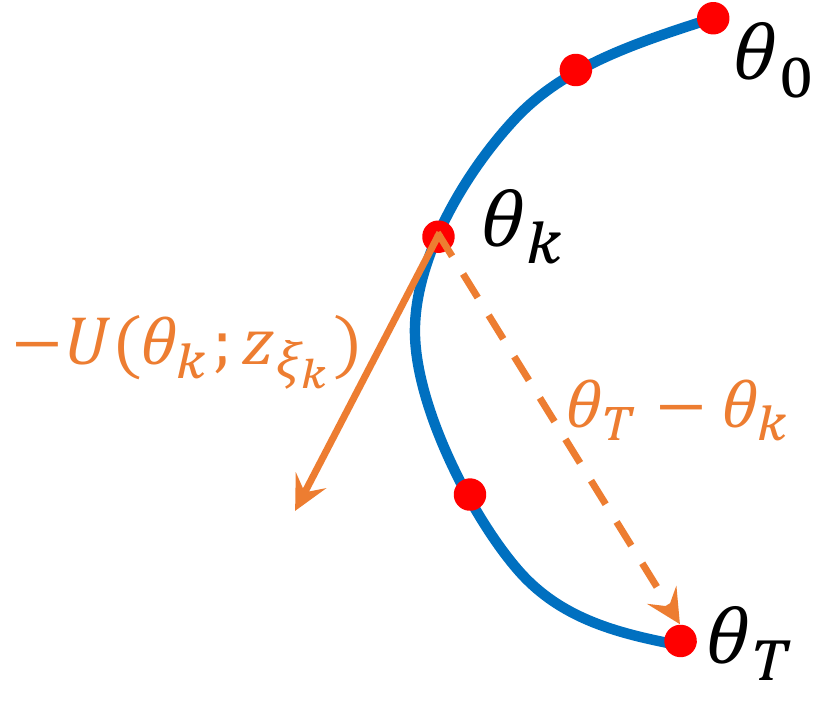}
%	\vspace{-2mm}
	\caption{\small{Illustration of optimization trajectories. 
		The left trajectory satisfies regularity principle with large $\gamma$ and the right trajectory satisfies regularity principle with small $\gamma$.}}\label{fig: 01} 
	\vspace{-2mm}
\end{figure}

To elaborate, the left hand side of \eqref{eq: principle} measures the coherence between $\theta_k-\theta_T$ and the corresponding model update $U(\theta_{k};z_{\xi_k})$. Intuitively, a positive coherence implies that the model update is well aligned with the  corresponding trajectory direction $\theta_k-\theta_T$. On the other hand, the right hand side of \eqref{eq: principle} regularizes the coherence by two non-negative terms: the square norm of the model update scaled by the learning rate $\eta$ and the optimality gap of the loss scaled by a parameter $\gamma$. Hence, a larger value of $\gamma$ implies that the update $U(\theta_{k};z_{\xi_k})$ is more coherent with the trajectory direction $\theta_k-\theta_T$. Fig. \ref{fig: 01} illustrates two optimization trajectories that satisfy the regularity principle with large and small $\gamma$, respectively. Intuitively, the left trajectory is close to a straight line where the update direction $-U(\theta_{k};z_{\xi_k})$ is well aligned with the trajectory direction $\theta_k-\theta_T$, implying a large $\gamma$. As a comparison, the right trajectory is curved and the update direction $-U(\theta_{k};z_{\xi_k})$ diverges from the trajectory direction $\theta_k-\theta_T$, implying a small $\gamma$. Therefore, the left trajectory is well regularized by the regularity principle and the trajectory length is much shorter than the curved trajectory, implying a faster convergence. Indeed, as we show next, the regularity principle with a lager $\gamma$ implies a faster convergence in nonconvex optimization.  

Next, we analyze the convergence of SA under the regularity principle in over-parameterized nonconvex optimization. In specific, we assume the model $\theta$ is over-parameterized so that the global minimum of the total loss $f(\theta; \mathcal{Z})$ interpolates all the sample losses $\{\ell(\theta; z_i)\}_{i=1}^n$, i.e., $\inf_{\theta\in \mathbb{R}^d} f(\theta) = \frac{1}{n} \sum_{i=1}^{n} \inf_{\theta\in \mathbb{R}^d} \ell(\theta; z_i)$. Such a scenario is common in deep learning where the deep neural networks typically have more parameters than samples and the training can overfit all the data samples to achieve near-zero loss.

\begin{theorem}[Convergence under Regularity Principle]\label{thm: convergence}
	Apply SA to solve the over-parameterized problem (P) and generate an optimization trajectory $\{\theta_0, \theta_1, ..., \theta_T \}$. If the optimization trajectory satisfies the regularity principle with parameter $\gamma>0$, then after $T=nB, B\in\mathbb{N}$ iterations (i.e., $B$ epochs), the average loss converges to the global minimum at the rate
	\begin{align}
	\frac{1}{T} \sum_{k=0}^{T-1} \ell(\theta_k;z_{\xi_k}) - \inf_{\theta\in \mathbb{R}^d} f(\theta; \mathcal{Z}) \le \frac{1}{2\eta T}\frac{\|\theta_0-\theta_T\|^2}{\gamma }. \label{eq: thm}
	\end{align}
%	\begin{enumerate}[leftmargin=*,topsep=0pt]
%		%\item The optimization path $\{\theta_k\}_k$ approaches the global minimizer $\theta^*$ with a monotonically diminishing distance, i.e., $\|\theta_{k+1} - \theta^*\| \le \|\theta_{k} - \theta^*\|$ for all $k=0,1,2,...,$.
%		%\item For all $i$, each $\ell(\cdot; z_i)$ converges to its global minimum along the optimization path, i.e., denote $\{i(T)\}_{T\in \mathbb{N}}$ as the sequence of iterations that sample $z_i$, then, $\lim_{T\to\infty}\ell(\theta_{i(T)}; z_i) = \ell(\theta^*; z_i)$.
%		\item 
%	\end{enumerate}
\end{theorem}

\Cref{thm: convergence} shows that the average loss convergences to the global minimum of the total loss at a sub-linear convergence rate. In particular, the convergence rate involves the factor $\gamma^{-1}$, which depends on the parameterization of the regularity principle. Intuitively, if the optimization trajectory is well-regularized by the regularity principle with a  large $\gamma$, then the trajectory is close to a straight line and the loss achieves a fast convergence to the global minimum. Based on the convergence rate in \eqref{eq: thm}, we can quantify the regularization effect of the regularity principle on the optimization trajectory by evaluating the problem-dependent constant factor $\|\theta_0 - \theta_T\|^2\gamma^{-1}$. In the subsequent sections, we empirically compute such a factor to explore the regularization effect of training techniques in deep learning optimization.   

%For many nonconvex ML models, e.g., deep neural networks, a prominent feature is the over-parameterization of the model, i.e., the model capacity is sufficient to over-fit all the training data samples. In another word, the global minimizers of the total loss $f$ under an over-parameterized model are common minimizers of all the individual loss functions $\ell(\cdot;z_i), i=1,...,n$. We summarize this fact formally as follows. 

%\textbf{Remark:} Under the $\gamma$-optimization principle, the sub-linear convergence rate in item 3 depends on the learning rate $\eta$ and the parameter $\gamma$ only, which are problem-independent parameters. Therefore, given a fixed learning rate, the parameter $\gamma$ of the optimization principle provides a universal quantification of the optimization quality. Inspired by this idea, in the subsequent sections, we conduct extensive experiments to examine the validity of the $\gamma$-optimization principle in DNN training. In particular, we empirically quantify the impacts of different training techniques on the DNN training by evaluating the parameter $\gamma$ of the optimization principle in each optimization process.

\section{Experiments on Network-level Training Techniques}

In this section, we examine the validity of the regularity principle in training DNNs with different neural network-level training techniques, i.e., activation function, batch normalization and skip-connection. We outline the exploration plan below and provide the details of the experiment setup in the corresponding subsections. 

\textbf{Exploration plan:} In all DNN trainings, we train the network for a sufficient number of epochs to achieve an approximate global minimum. We store the optimization trajectory $\{\theta_k \}_k$, loss $\{\ell(\theta_k;z_{\xi_k})\}_k$ and update $\{U(\theta_k;z_{\xi_k})\}_k$ that are generated in each DNN training. Then, we compute the upper bound for $\gamma$ in each iteration according to  \eqref{eq: principle}, where $\theta_T$ corresponds to the network parameters produced in the last training iteration. We report the convergence rate factor $\gamma/\|\theta_0-\theta_T\|^2$ that is involved in \eqref{eq: thm}.

\subsection{Effect of Neuron Activation Function on Regularity Principle}

\textbf{Experiment setup:} We train a variant of the AlexNet \cite{Zhang_2017} and the ResNet-18 \cite{He2015} with different choices of activation functions for all the nonlinear neurons. The activation functions that we explore include sigmoid, tanh, ReLU and leaky ReLU (with slope $10^{-2}$).   We apply the standard SGD optimizer with a fixed initialization point, a mini-batch size 128 and a constant learning rate $\eta=0.05$ to train these networks for 150 epochs on the CIFAR-10 and CIFAR-100 datasets \cite{Krizhevsky09}, respectively, and we use the cross-entropy loss.

\begin{figure}[htbp]%[bth]
		\vspace{-2mm}
	\centering
	\includegraphics[width=0.24\textwidth,height=0.2\textwidth]{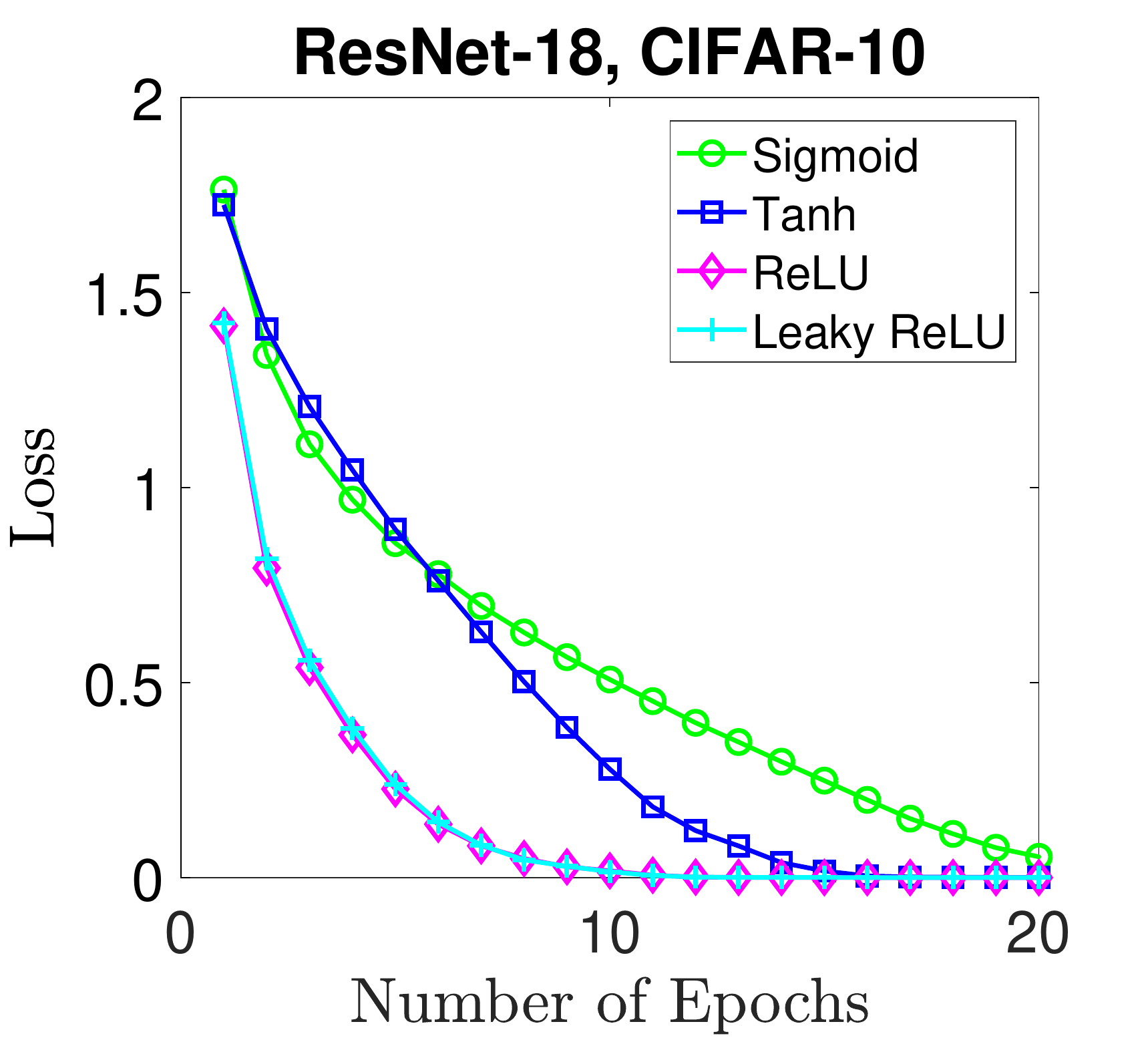}
	\includegraphics[width=0.24\textwidth,height=0.2\textwidth]{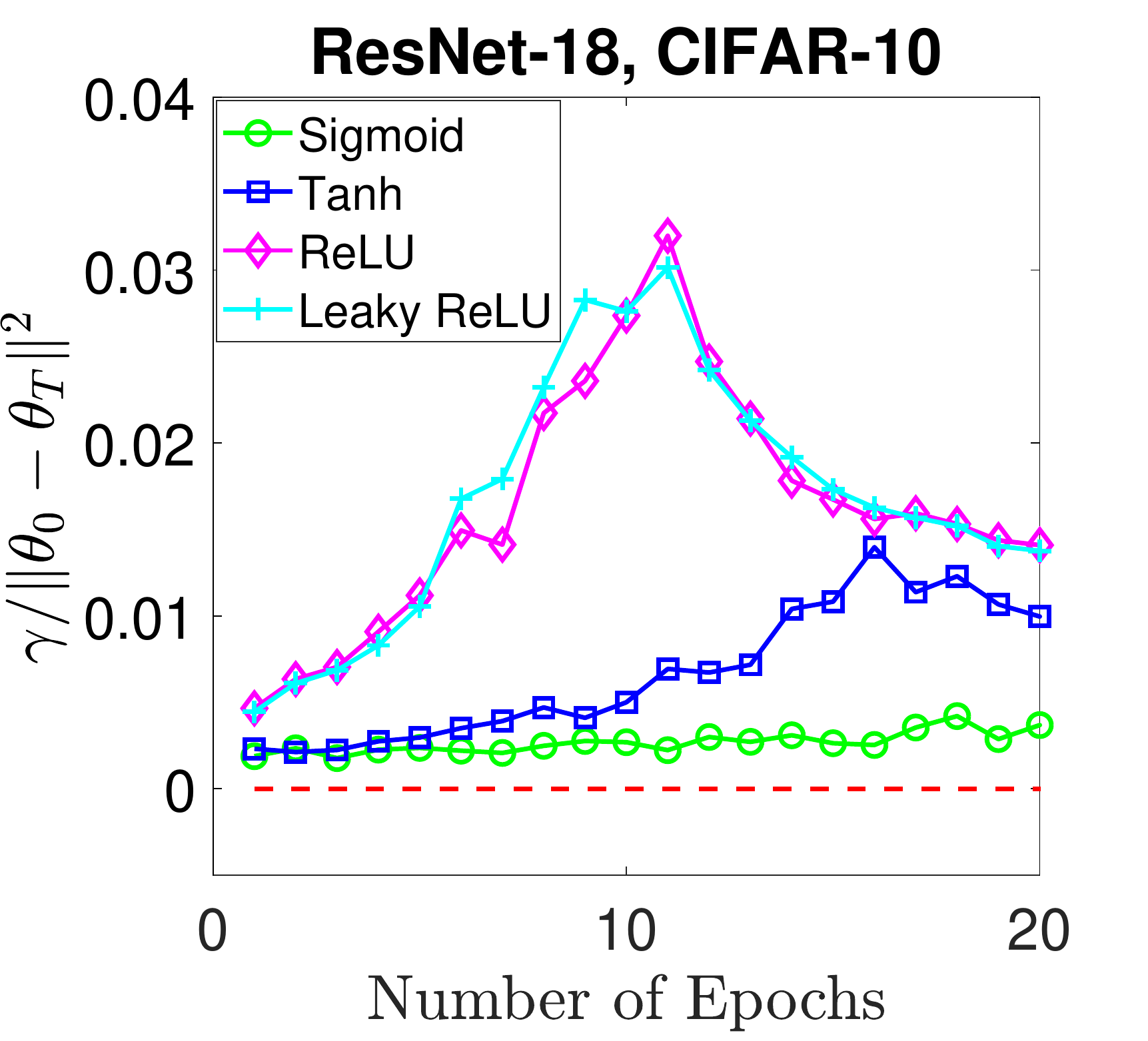}
	\includegraphics[width=0.24\textwidth,height=0.2\textwidth]{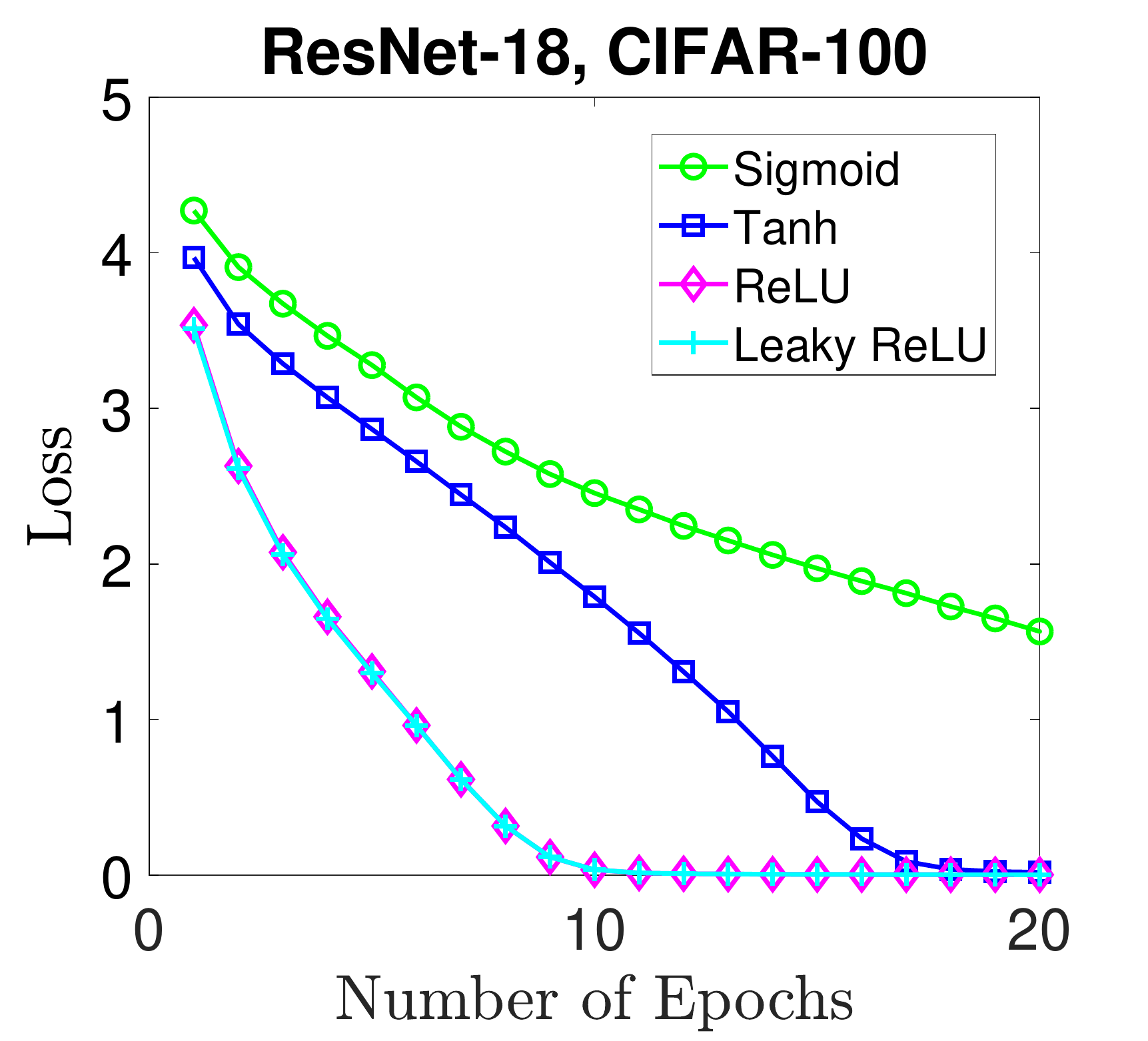}
	\includegraphics[width=0.24\textwidth,height=0.2\textwidth]{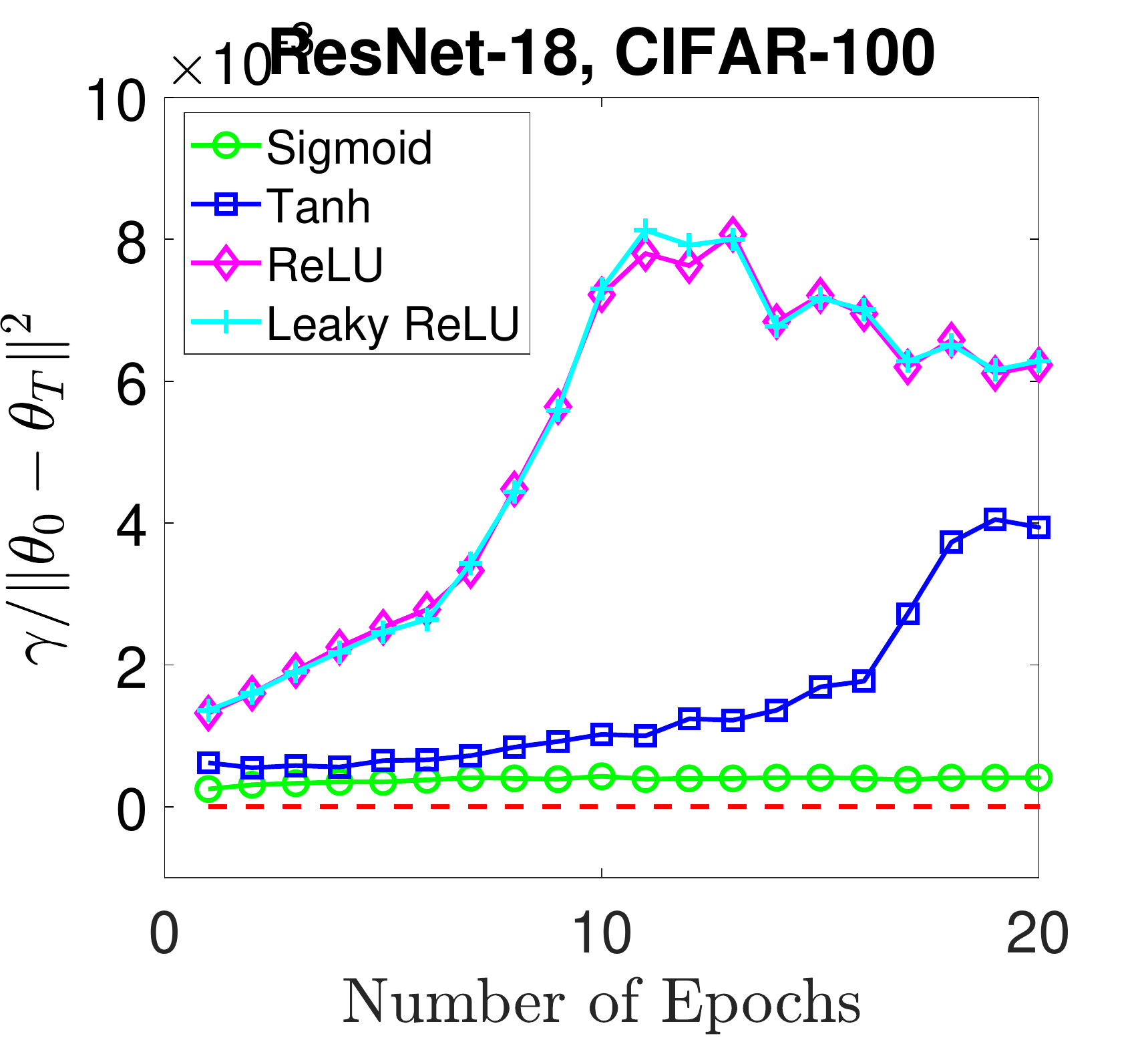}
		\vspace{-6mm}
	\caption{\small{Training ResNet-18 with different activation functions.}}\label{fig: 1} 
%		\vspace{-2mm}
\end{figure}

Fig. \ref{fig: 1} presents the ResNet training results of the average training loss and the factor $\gamma/\|\theta_0-\theta_T\|^2$ involved in the convergence result \eqref{eq: thm} along the optimization trajectory. It can be seen that all the trainings that adopt different activation functions obey the regularity principle with $\gamma>0$ in the training process. 
%Also, regarding the trainings of the Alexnet (first two columns), we observe that the trainings with the sigmoid activation function converge much slower than those with the other activation functions, and the parameter $\gamma$ of the regularity principle under the sigmoid activation is much smaller than those under the other activation functions. These observations are consistent with \Cref{thm: convergence}, where a well-regularized optimization trajectory with a larger $\gamma$ implies faster convergence. 
In particular, observe that the trainings with ReLU types of activation functions converge faster than those with tanh activation function, which is further faster than the trainings with sigmoid activation function. Moreover, one can see  that the trainings with ReLU types of activation functions satisfy the regularity principle with the largest $\gamma$, whereas the trainings with sigmoid activation function have the smallest $\gamma$. These observations are consistent with \Cref{thm: convergence}, where a well-regularized optimization trajectory with a larger $\gamma$ implies faster convergence. We present the AlexNet training results in  \Cref{sec: app: 1}, where one can make very similar observations.

%the figures in the first and second rows respectively show that the distance-to-minimizer diminishes monotonically and the training loss converges to the global minimum in these trainings. These observations are consistent with the theoretical implications of the $\gamma$-optimization principle in items 1 and 2 of \Cref{thm: convergence}. 

%of the proposed optimization principle, this explains why training with sigmoid activation function is slower than training with other activation functions.

%In these experiments, we observe that the convergence speed of the trainings with ReLU types of activation functions is the fastest, and is followed by that of the trainings with tanh activation function and sigmoid activation function, respectively. Moreover, from the figures in the last column, one can see  that the $\gamma$ in the trainings with ReLU types of activation functions has the largest value, and is followed by that in the trainings with tanh activation function and sigmoid activation function, respectively. Such an observation is consistent with the convergence rate result under the $\gamma$-optimization principle in item 3 of \Cref{thm: convergence}, i.e., a larger $\gamma$ leads to faster convergence.

%\vspace{-2mm}
%\begin{figure}[bth]
%	\centering
%	
%	\hspace{2mm}
%
%	\hspace{2mm}
%	
%	\\
%	
%	\hspace{2mm}
%	
%	\hspace{2mm}
%	
%	\vspace{-2mm}
%	\caption{Training Resnet-18 with different activation functions.}\label{fig: 2} 
%\end{figure}
%\vspace{-2mm}

{ \textbf{Experiment setup:}  Next, we train a U-Net \cite{unet} with different choices of activation functions on the CIFAR-10 and CIFAR-100 datasets. The activation functions that we explore include sigmoid, tanh, ReLU and leaky ReLU (with slope $10^{-2}$). We apply the standard SGD optimizer with a fixed initialization point, a mini-batch size 128 and a constant learning rate $\eta=0.005$ to train these networks for 150 epochs, and we use the MSE loss.}

\begin{figure}[htbp]%[bth]
	\vspace{-2mm}
	\centering
	\includegraphics[width=0.24\textwidth,height=0.2\textwidth]{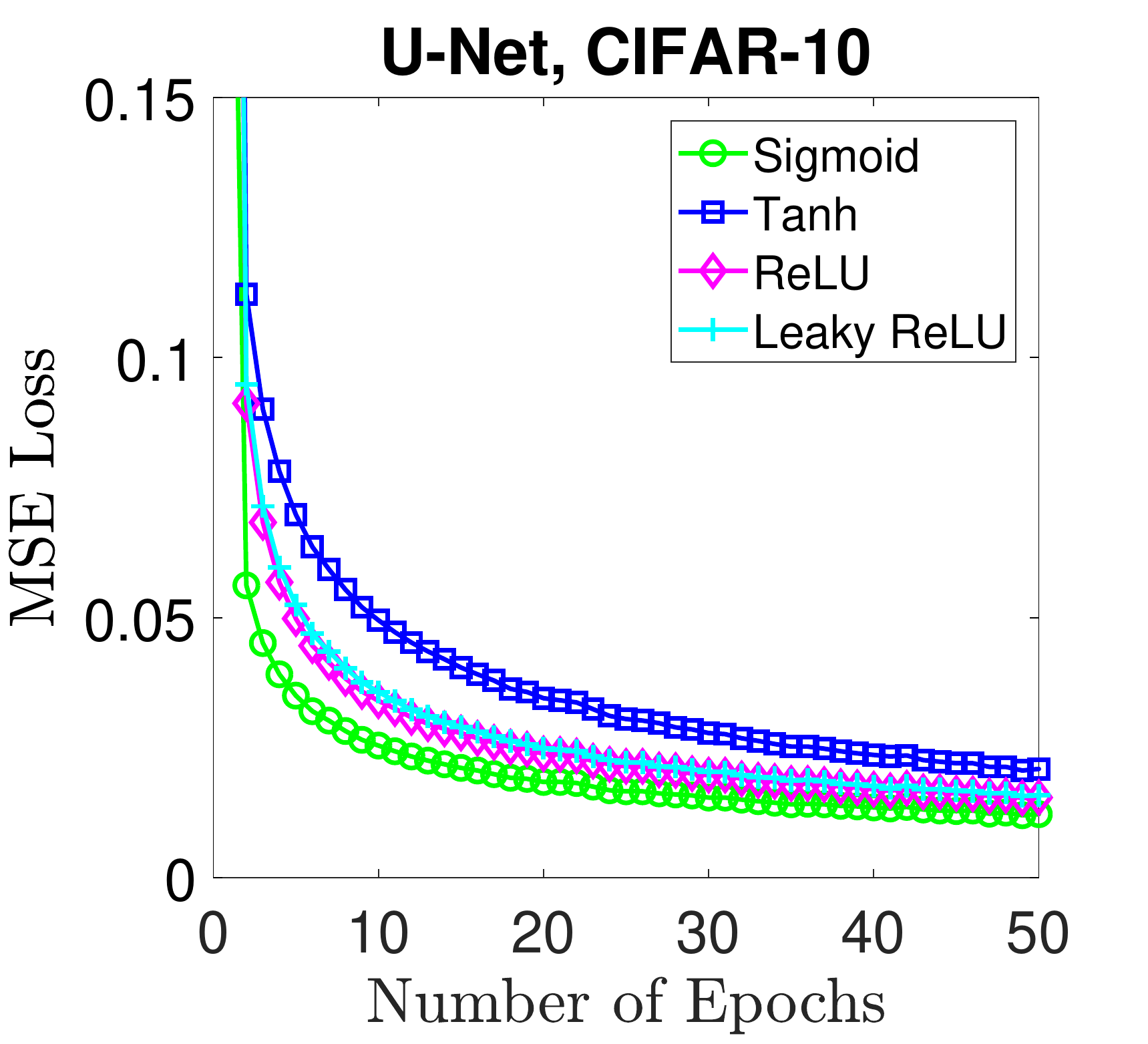}
	\includegraphics[width=0.24\textwidth,height=0.2\textwidth]{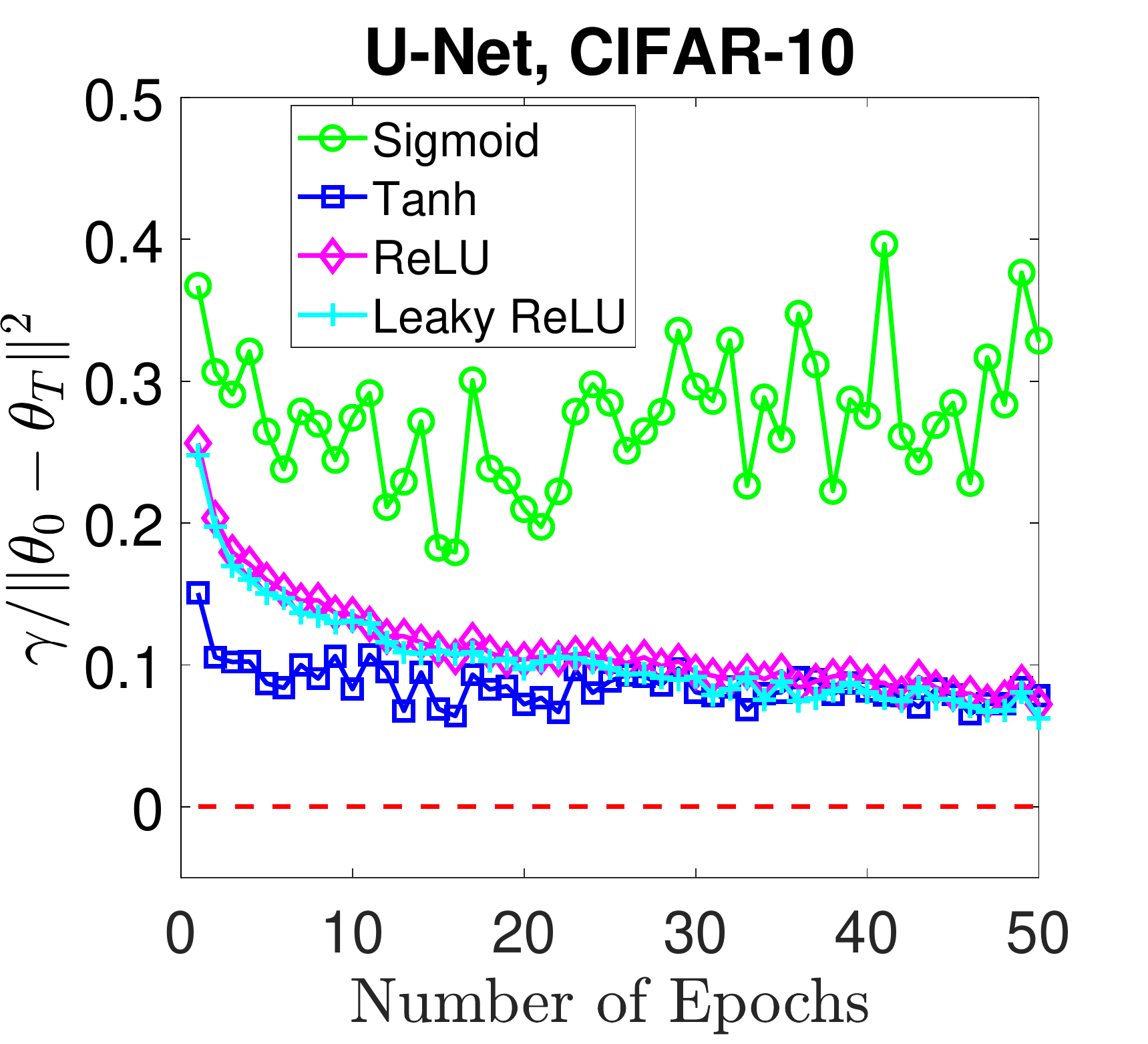}
	\includegraphics[width=0.24\textwidth,height=0.2\textwidth]{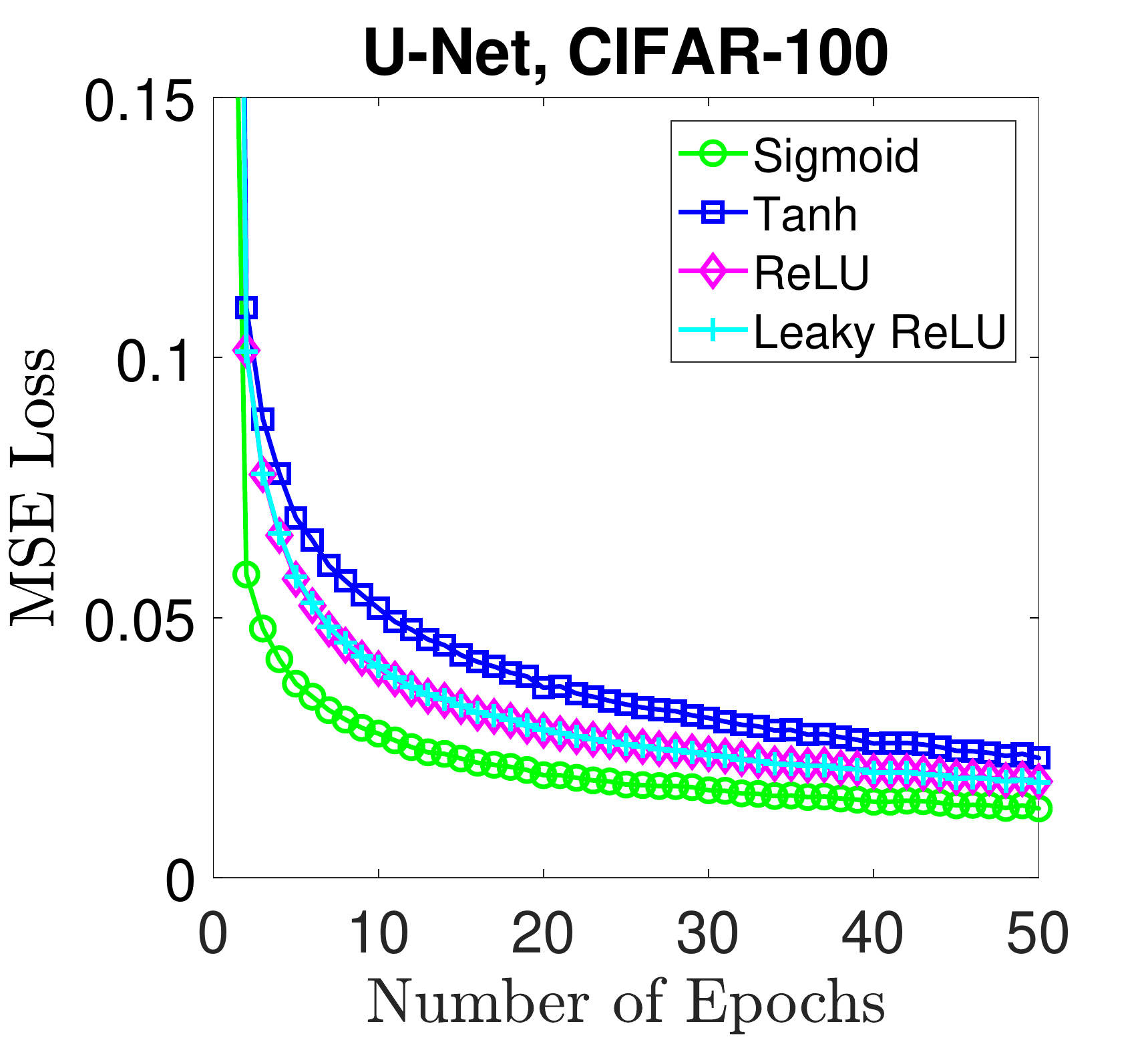}
	\includegraphics[width=0.24\textwidth,height=0.2\textwidth]{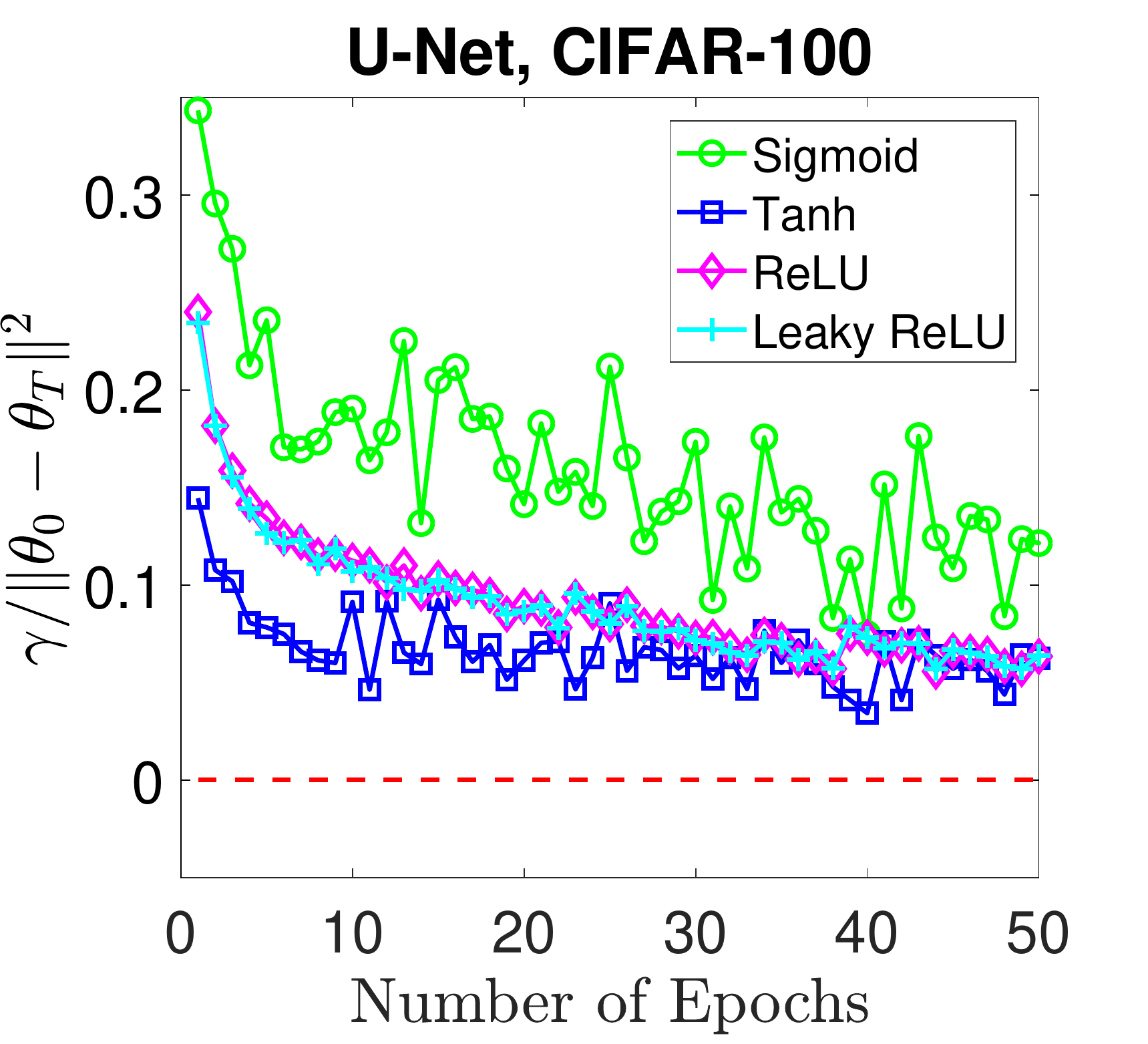}
	\vspace{-6mm}
	\caption{\small{Training U-Net with different activation functions.}}\label{fig: 101} 
	%	\vspace{-2mm}
\end{figure}

Fig. \ref{fig: 101} shows the training loss curves and their corresponding $\gamma/\|\theta_0-\theta_T\|^2$ of the regularity principle. In these U-Net trainings, it can be seen that the trainings with the sigmoid activation function converge faster than those with the ReLU types of activation functions, and the trainings with the tanh activation function converge the slowest. Furthermore, one can observe that the training trajectories with the sigmoid activation function obey the regularity principle with the largest $\gamma$, whereas the training trajectories with the tanh activation function satisfy the regularity principle with the smallest $\gamma$. This is consistent with the theoretical convergence rate characterized in \Cref{thm: convergence}.

The experiments in this subsection show that DNN optimization trajectories are regularized by the regularity principle, and the regularization parameter affects the convergence speed. It seems that ReLU type of activation functions help regularize the optimization trajectory in the tested classification tasks and sigmoid activation function help regularize the trajectory in the tested regression tasks.

\subsection{Effect of Batch Normalization on Regularity Principle}

\textbf{Experiment setup:} We train the ResNet-18, 34 and the VGG-11, 16 networks with the settings: 1) keep all the BN layers; 2) keep the first BN layer in each block; and 3) remove all the BN layers. We remove all dropout layers in the VGG networks. We apply SGD with a fixed initialization, a constant learning rate ($\eta=0.05$ for ResNet, 0.01 for VGG) and batch size 128 to train these networks on the CIFAR-10 and CIFAR-100 datasets, respectively, and we use the cross-entropy loss. 

\begin{figure}[htbp]%[bth]
	\vspace{-2mm}
	\centering
	\includegraphics[width=0.24\textwidth,height=0.2\textwidth]{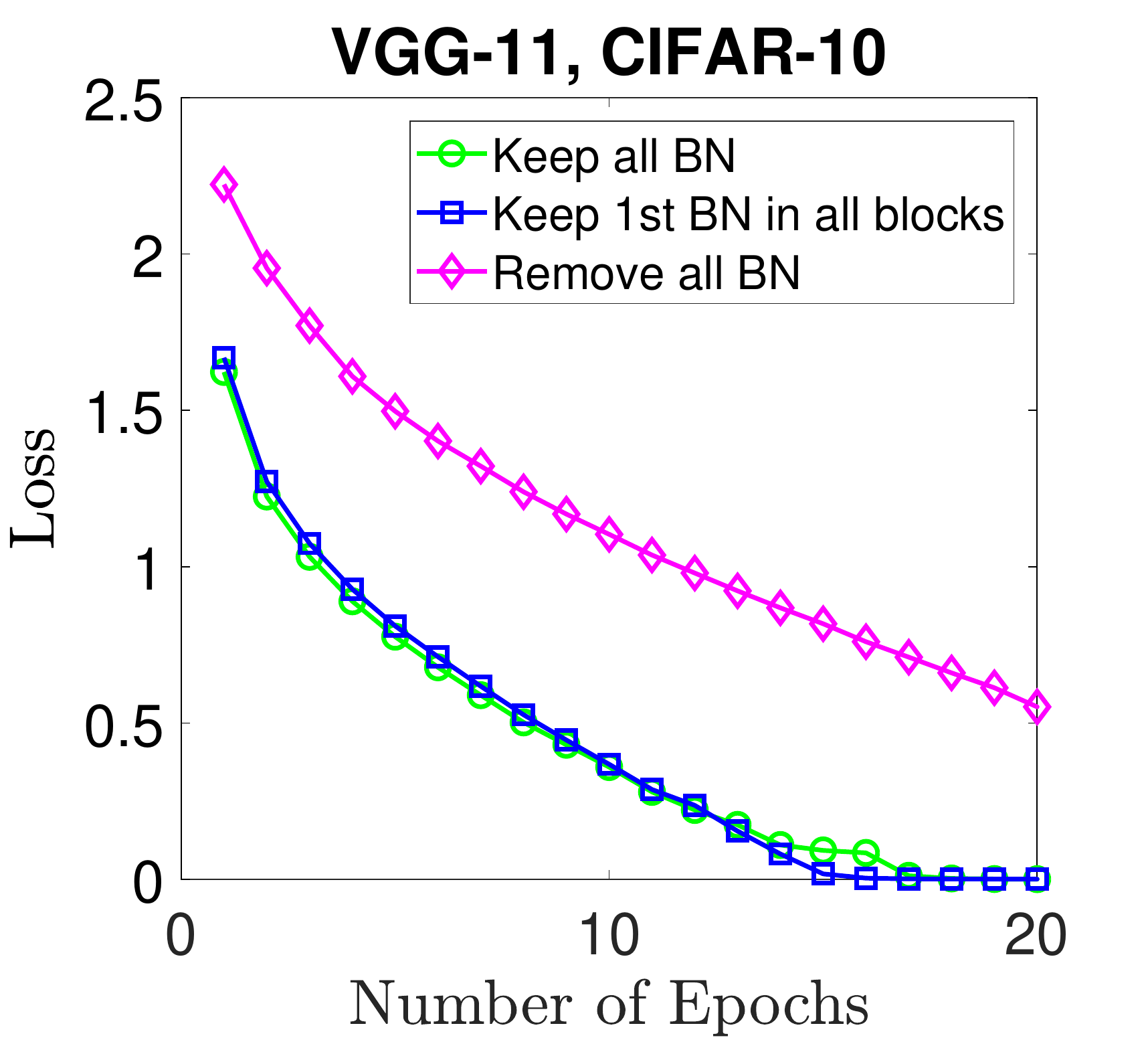}
	\includegraphics[width=0.24\textwidth,height=0.2\textwidth]{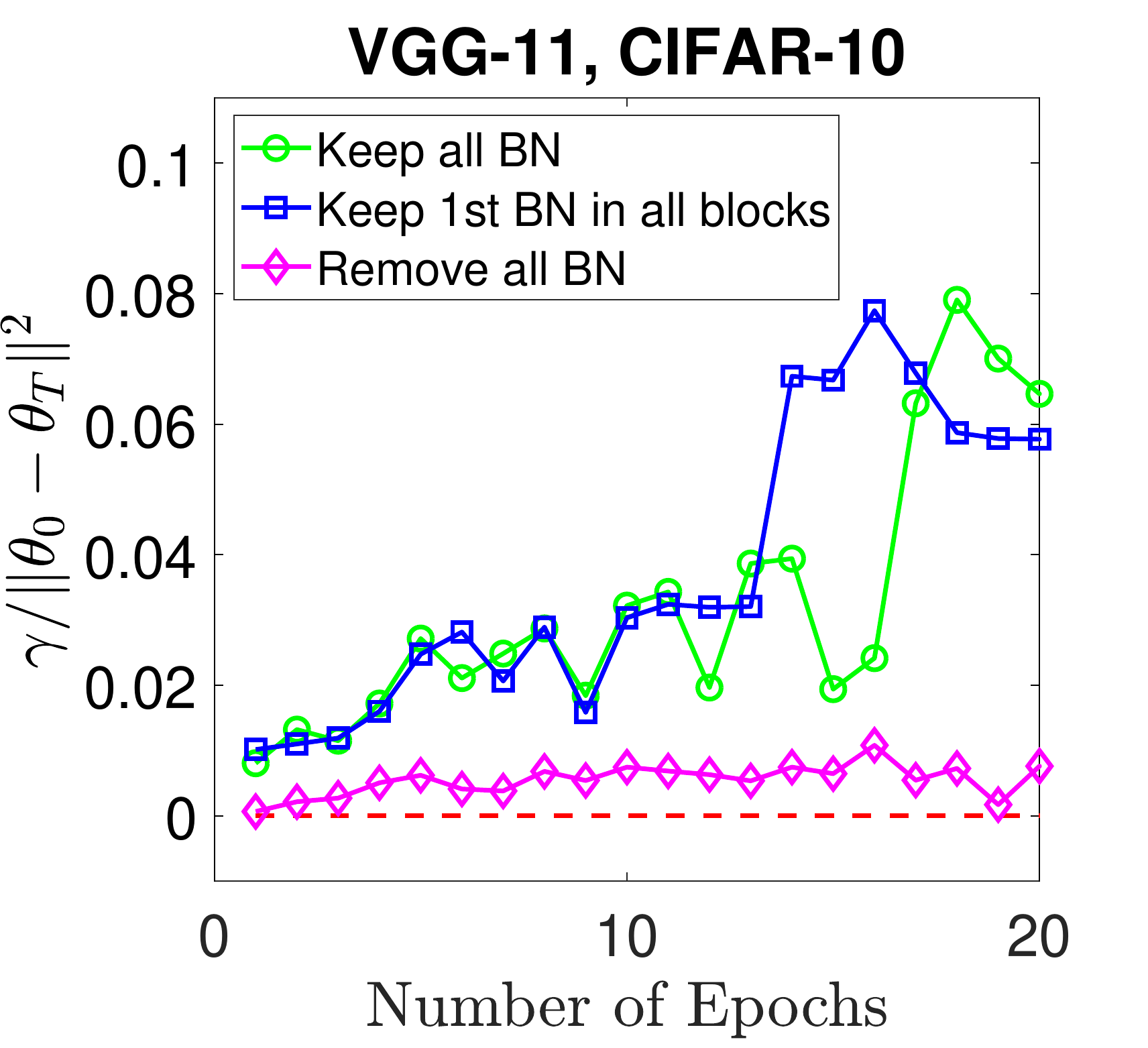}
	\includegraphics[width=0.24\textwidth,height=0.2\textwidth]{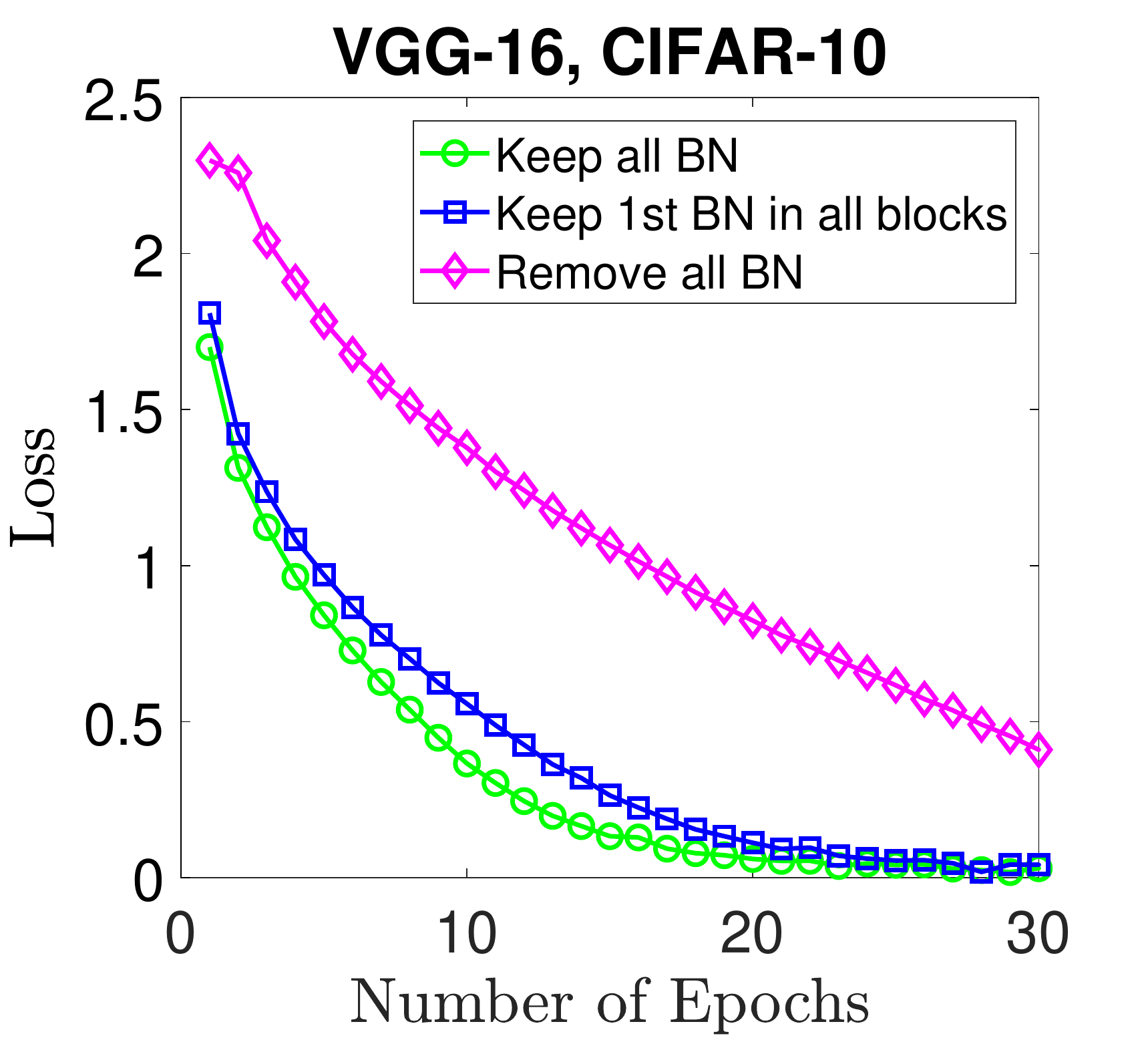}
	\includegraphics[width=0.24\textwidth,height=0.2\textwidth]{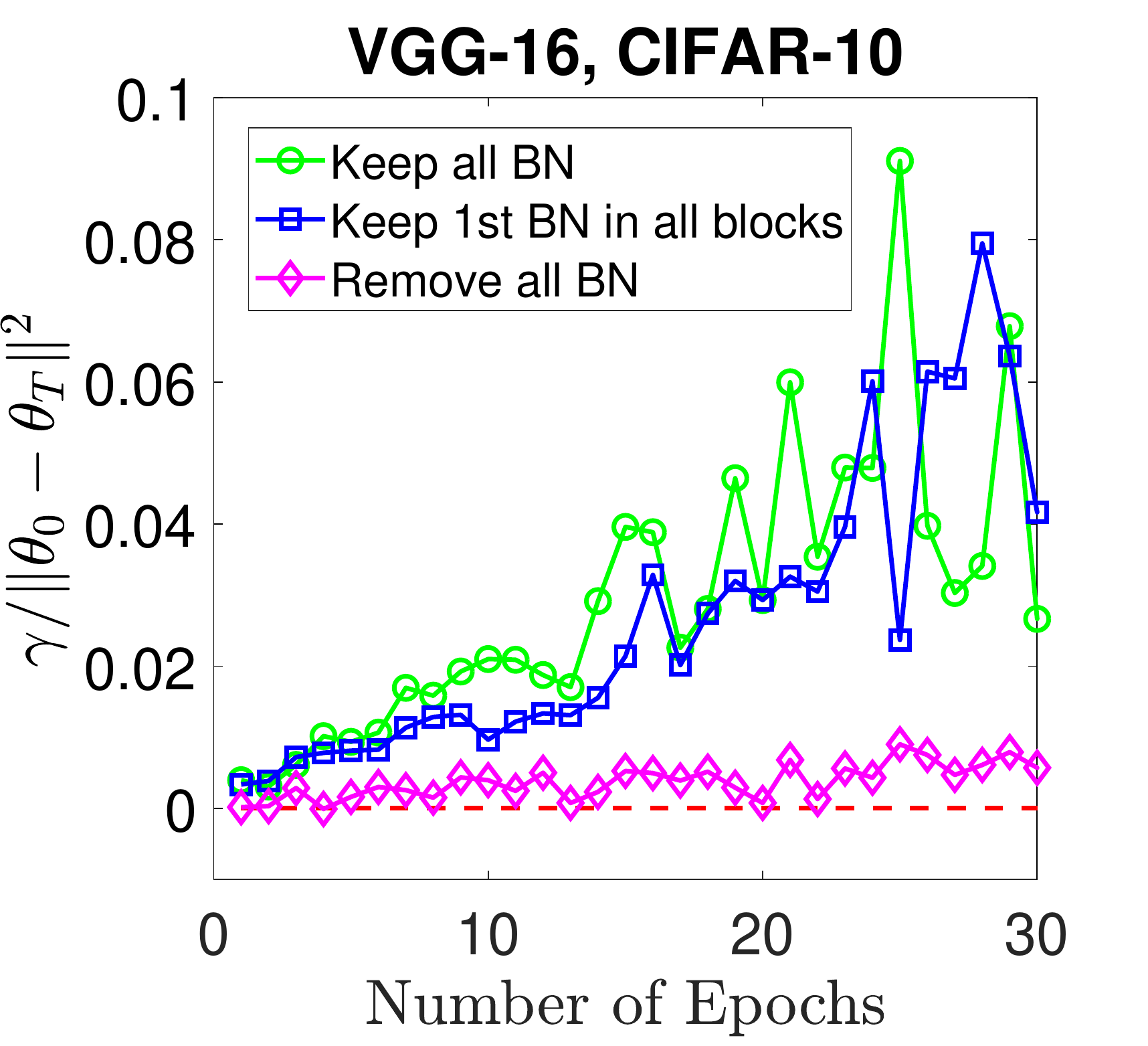}
	\vspace{-6mm}
	\caption{\small{Training VGGs with and without batch normalization on CIFAR-10.}}\label{fig: 3} 
%\vspace{-2mm}
\end{figure}

Fig. \ref{fig: 3} shows the VGG training results on the CIFAR-10 dataset. It can be seen that the trainings with all BN layers removed suffer from a significant convergence slow down, and  the corresponding optimization trajectories obey the regularity principle with a very small $\gamma$. 
On the other hand, the trainings that keep the first BN layer in each block converge as fast as those that keep all the BN layers, and their optimization trajectories obey the regularity principle with a large $\gamma$. These empirical observations corroborate the theoretical implication of the regularity principle in \Cref{thm: convergence}. We also present the ResNet training results in  \Cref{sec: app: 2}, where one can make very similar observations.
 
%Regarding the $\gamma$ values,  they are all positive throughout these trainings, which demonstrates the validity of the $\gamma$-optimization principle. Moreover, the $\gamma$ in the trainings without any BN layer is considerably smaller than that in the trainings with either one BN layer or all BN layers. Such an observation is consistent with the convergence rate result in item 3 of \Cref{thm: convergence}, where a larger $\gamma$ implies faster convergence. %Therefore, the principle  quantifies the  effectiveness of batch normalization in terms of the parameter $\gamma$.

\begin{figure}[htbp]%[bth]
%	\vspace{-2mm}
	\centering
	\includegraphics[width=0.24\textwidth,height=0.2\textwidth]{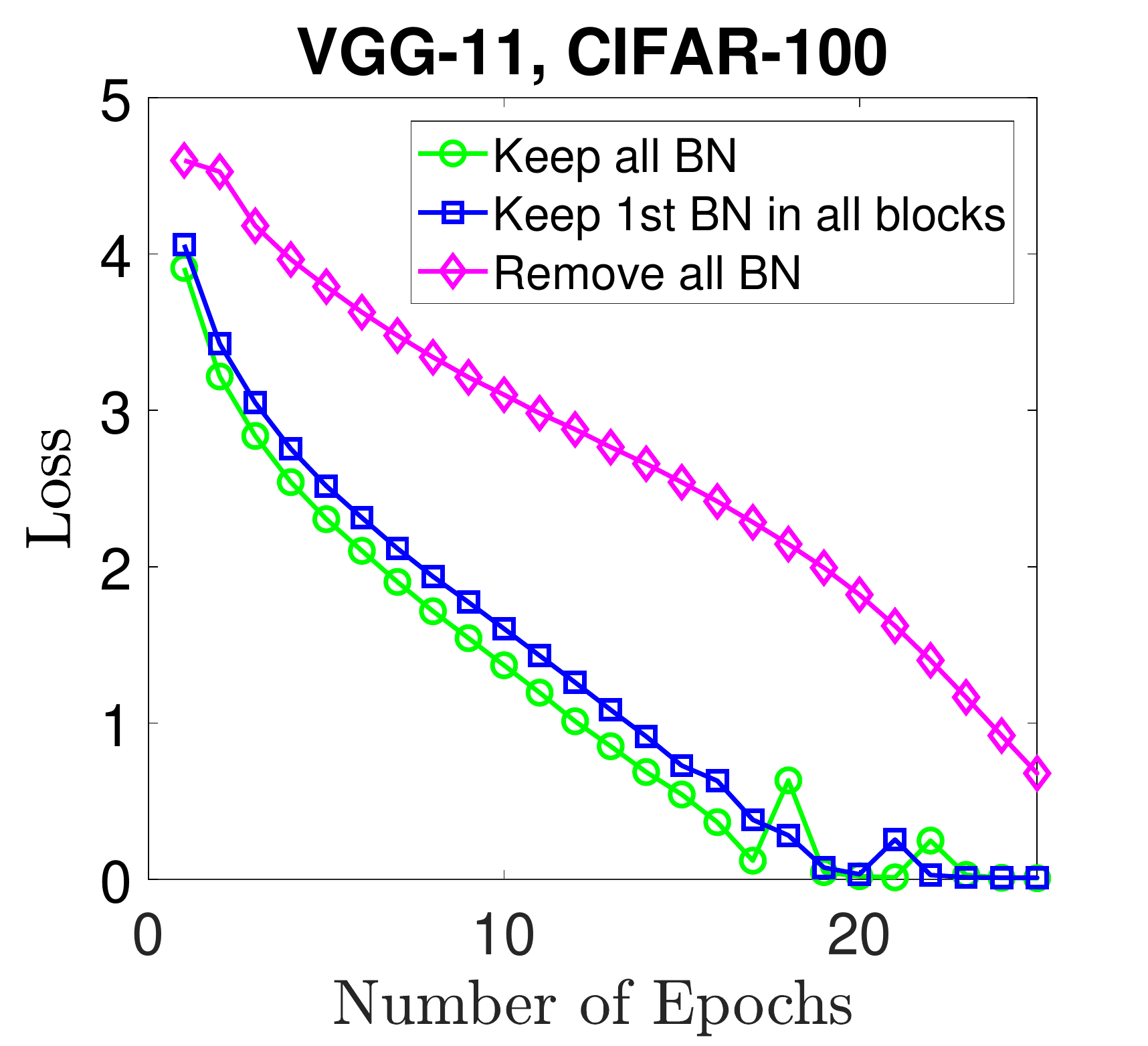}
	\includegraphics[width=0.24\textwidth,height=0.2\textwidth]{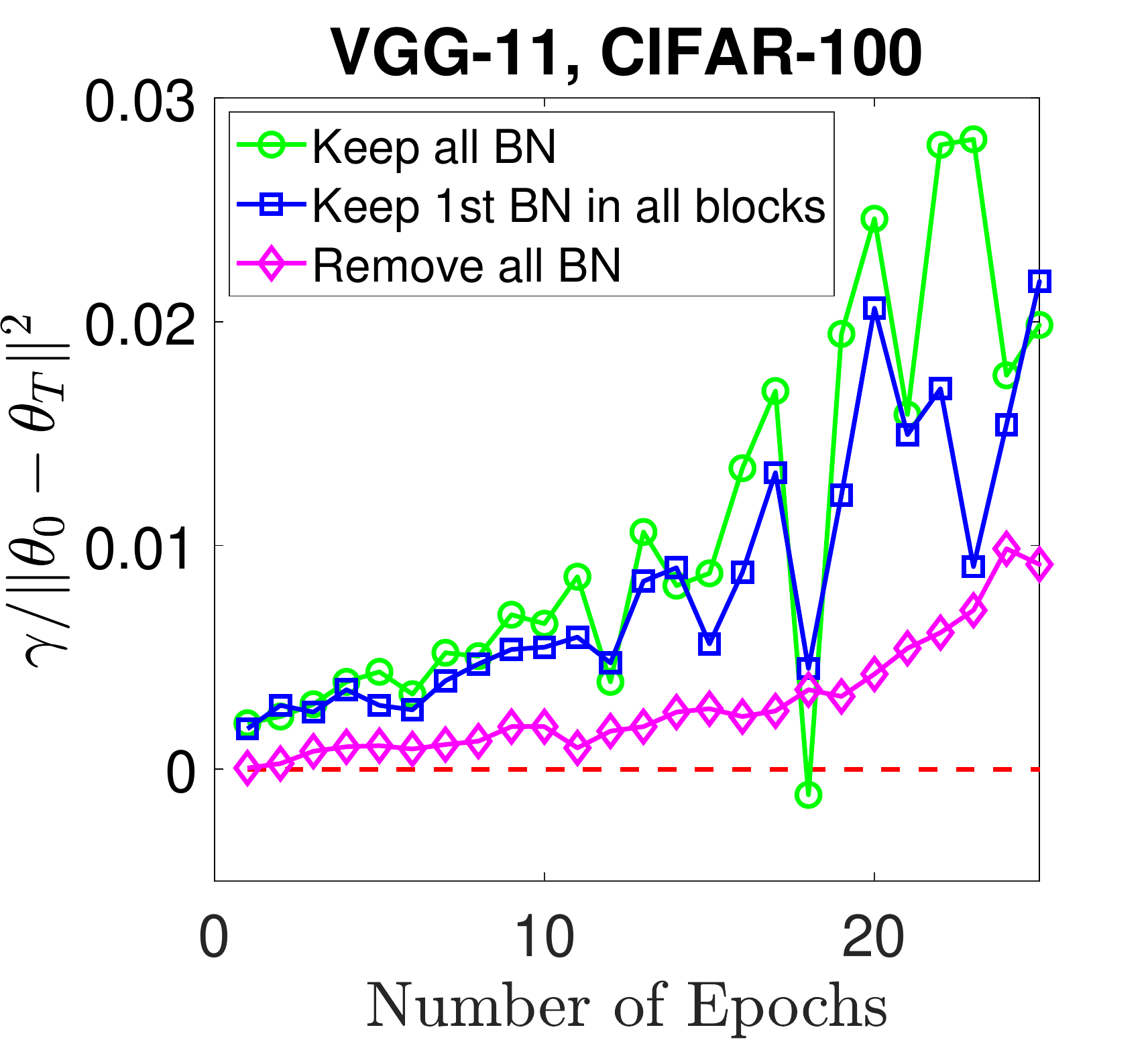}
	\includegraphics[width=0.24\textwidth,height=0.2\textwidth]{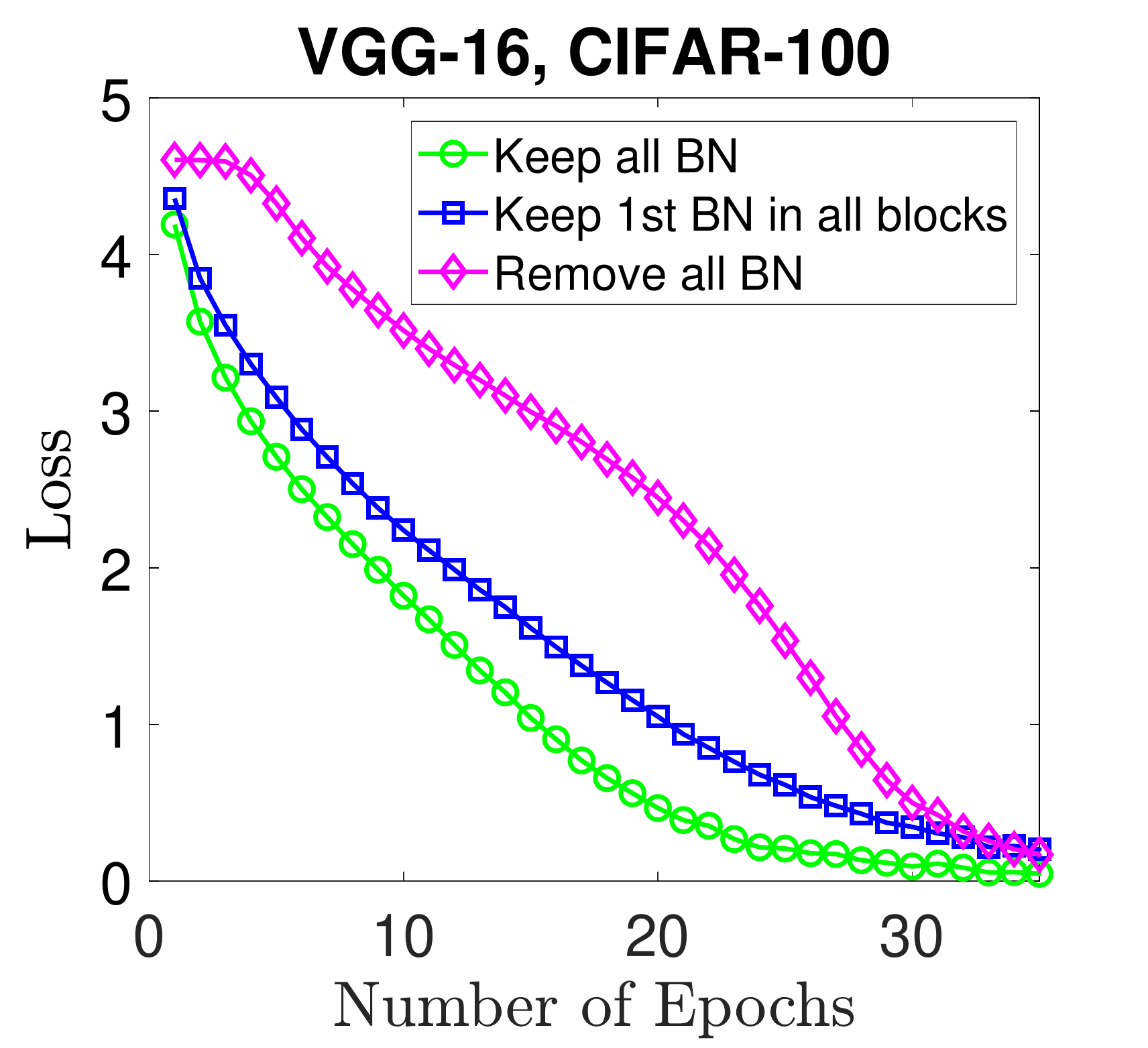}
	\includegraphics[width=0.24\textwidth,height=0.2\textwidth]{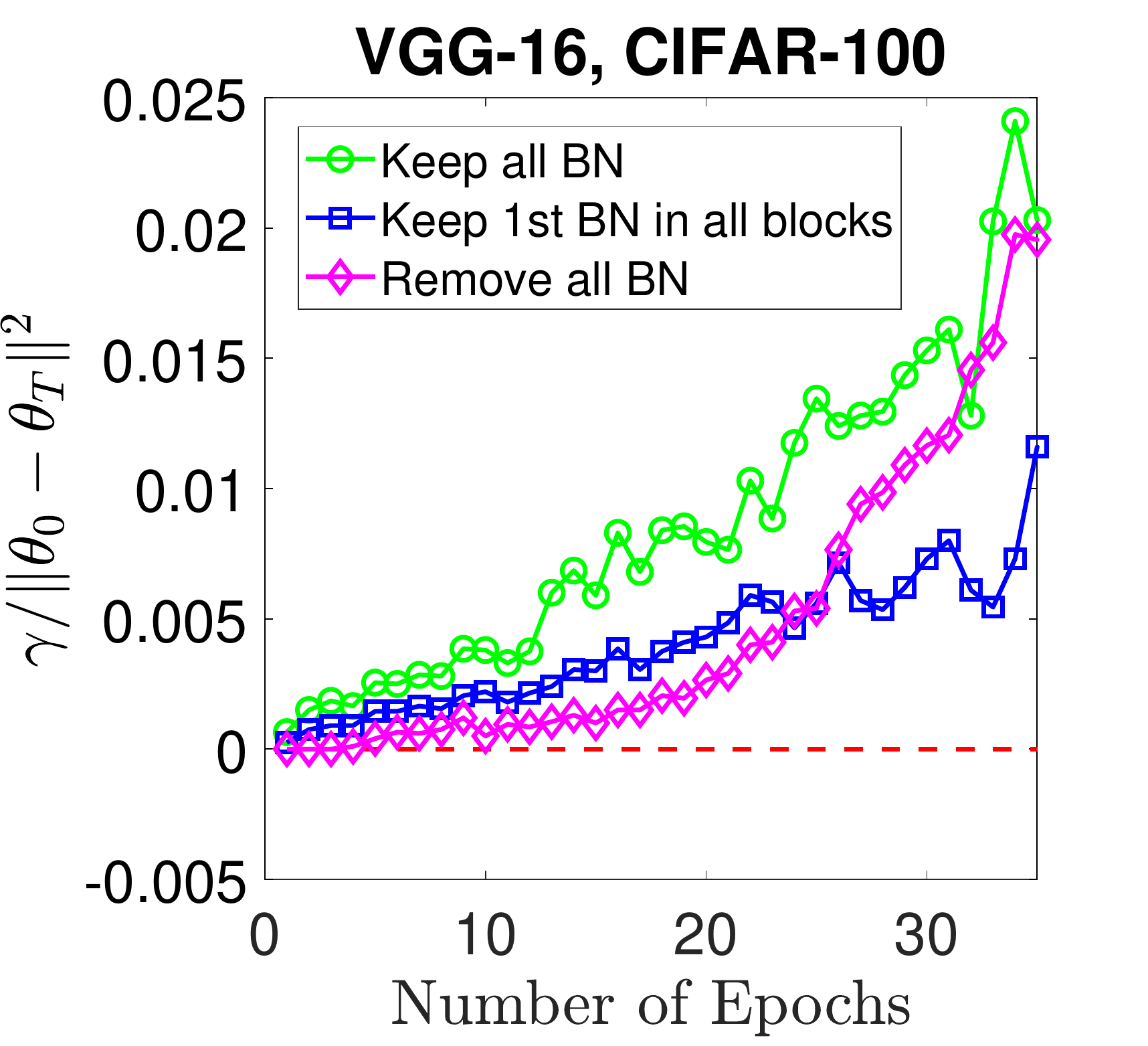}
	\vspace{-6mm}
	\caption{\small{Training VGGs with and without batch normalization on CIFAR-100.}}\label{fig: 4} 
\end{figure}

We also report in Fig. \ref{fig: 4} the results of training VGG networks on the more complex CIFAR-100 dataset, where one can make similar observations as those in the above CIFAR-10 experiment. To elaborate, the trainings that keep all the BN layers are slightly faster than those that keep the first BN layer in each block, and both of them converge much faster than the trainings that remove all the BN layers. On the other hand, the training trajectories with all BN layers kept obey the regularity principle with a larger $\gamma$ than that with only the first BN layer kept, and the training trajectories with all BN layers removed obey the regularity principle with the smallest $\gamma$. This shows that BN layers help to regularize the optimization trajectory via the regularity principle. We also present the ResNet training results in  \Cref{sec: app: 2}, where one can make very similar observations.

%{ \textbf{Experiment setup:}  Next, we train a Unet on the Cifar10 and Cifar100 datasets and consider the settings: 1) keep all the skip connections; 2) keep the last 3,2,1 skip connections, respectively and 3) remove all the skip connections. We apply the standard SGD optimizer with a fixed initialization point, a mini-batch size 128 and a constant learning rate $\eta=0.005$ to train these networks for 150 epochs, and we use the MSE loss.}

%\begin{figure*}[bth]
%	\vspace{-2mm}
%	\centering
%%	\includegraphics[width=0.24\textwidth,height=0.2\textwidth]{fig_unet/BN/loss/mnist_unet.eps}	
%	\includegraphics[width=0.24\textwidth,height=0.2\textwidth]{fig_unet/BN/loss/cifar10_unet.eps}
%	\includegraphics[width=0.24\textwidth,height=0.2\textwidth]{fig_unet/BN/loss/cifar100_unet.eps}
%%	\includegraphics[width=0.24\textwidth,height=0.2\textwidth]{fig_unet/BN/gamma/mnist_unet.eps}
%	\includegraphics[width=0.24\textwidth,height=0.2\textwidth]{fig_unet/BN/gamma/cifar10_unet.eps}
%	\includegraphics[width=0.24\textwidth,height=0.2\textwidth]{fig_unet/BN/gamma/cifar100_unet.eps}
%	\vspace{-2mm}
%	\caption{Training Unet with and without batchnormalization.}\label{fig: 102} 
%	\vspace{-2mm}
%\end{figure*}

%The  experiments in this subsection demonstrate that the regularity principle is able to capture and quantify the impact of batchnormalization on DNN training in terms of the regularization parameter $\gamma$.

\subsection{Effect of Skip-connection on Regularity Principle}

\textbf{Experiment setup:} We train the ResNet-18 with the settings: 1) keep all the skip-connections; and 2) keep the first skip-connection in each block. For the ResNet-34, we consider an additional setting where we keep the first two skip-connections in each block. We apply SGD  with a fixed initialization point, a constant learning rate $\eta=0.05$ and batch size 128 to train these networks for 150 epochs on the CIFAR-10 and CIFAR-100 datasets, respectively, and use the cross-entropy loss.

\begin{figure}[htbp]%[bth]
	\vspace{-2mm}
	\centering
	\includegraphics[width=0.24\textwidth,height=0.2\textwidth]{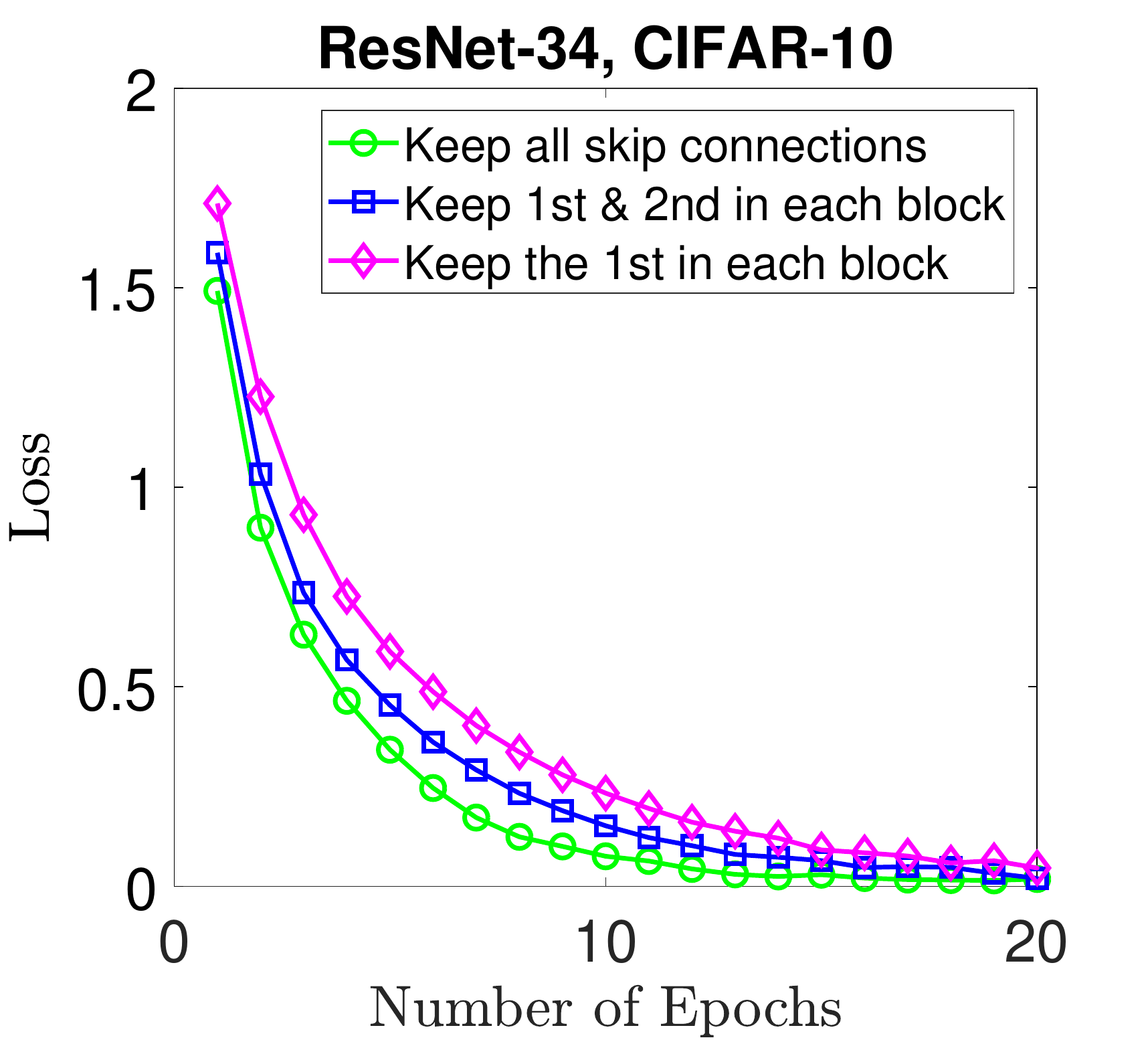}
	\includegraphics[width=0.24\textwidth,height=0.2\textwidth]{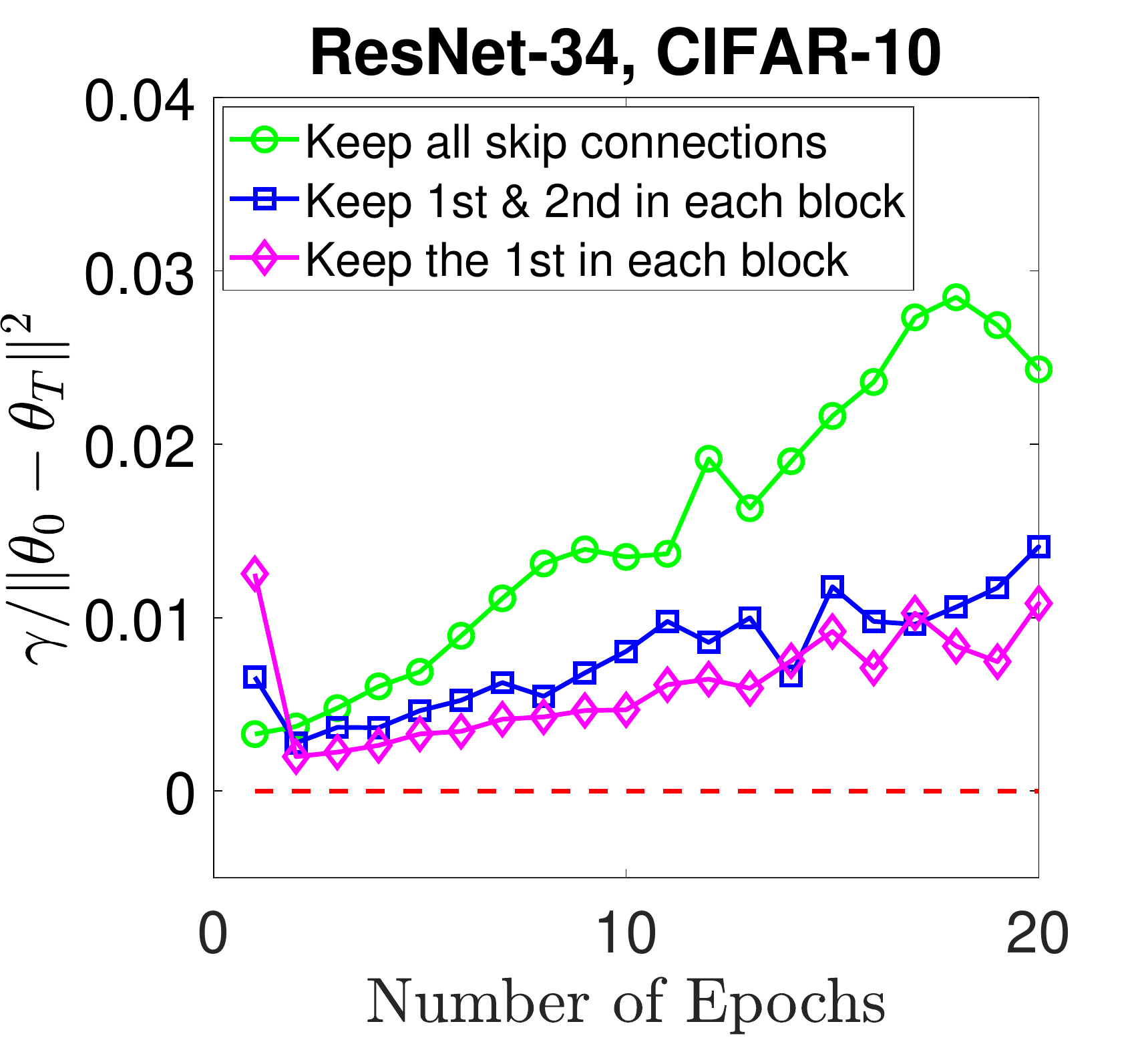}
	\includegraphics[width=0.24\textwidth,height=0.2\textwidth]{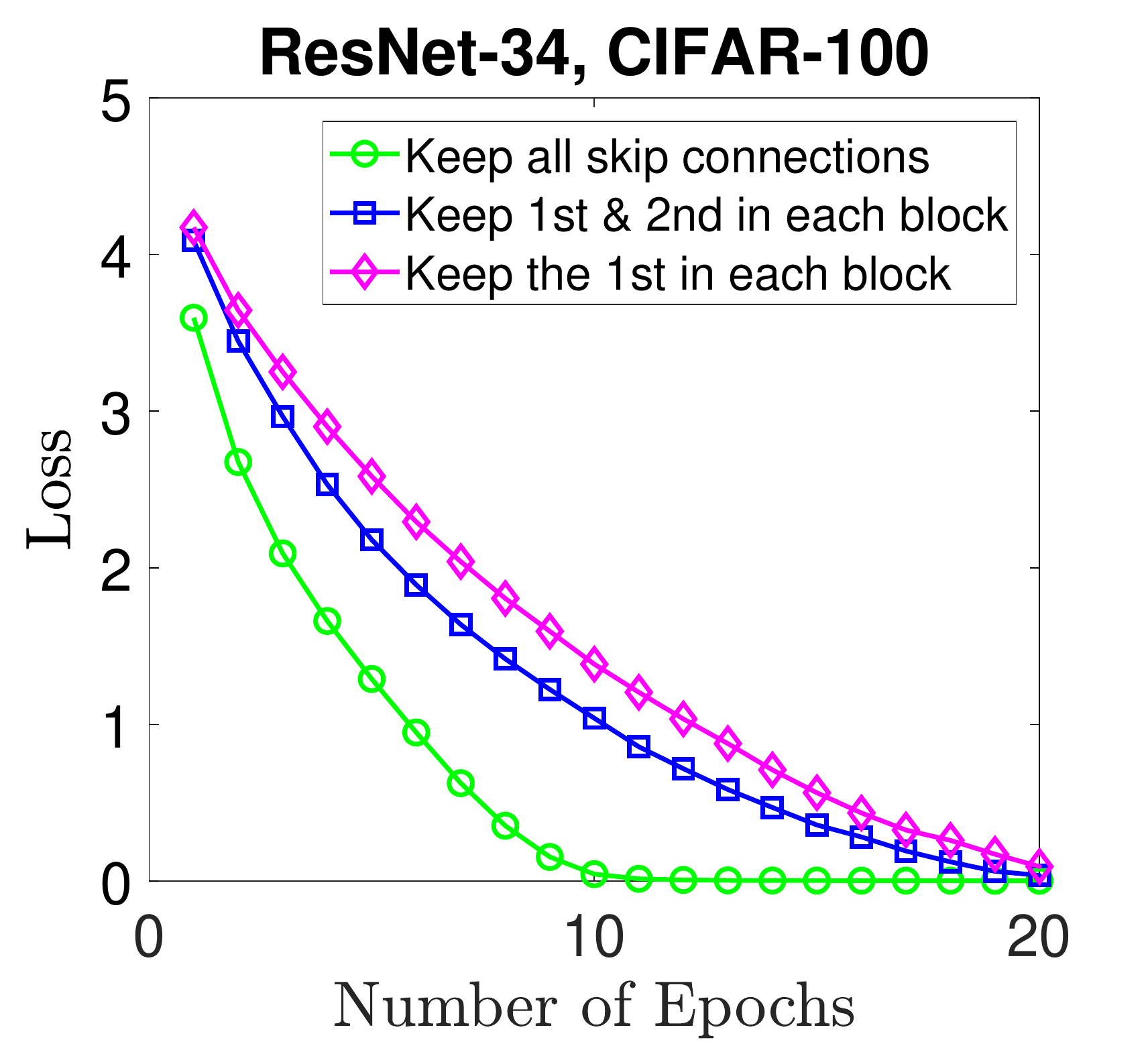}
	\includegraphics[width=0.24\textwidth,height=0.2\textwidth]{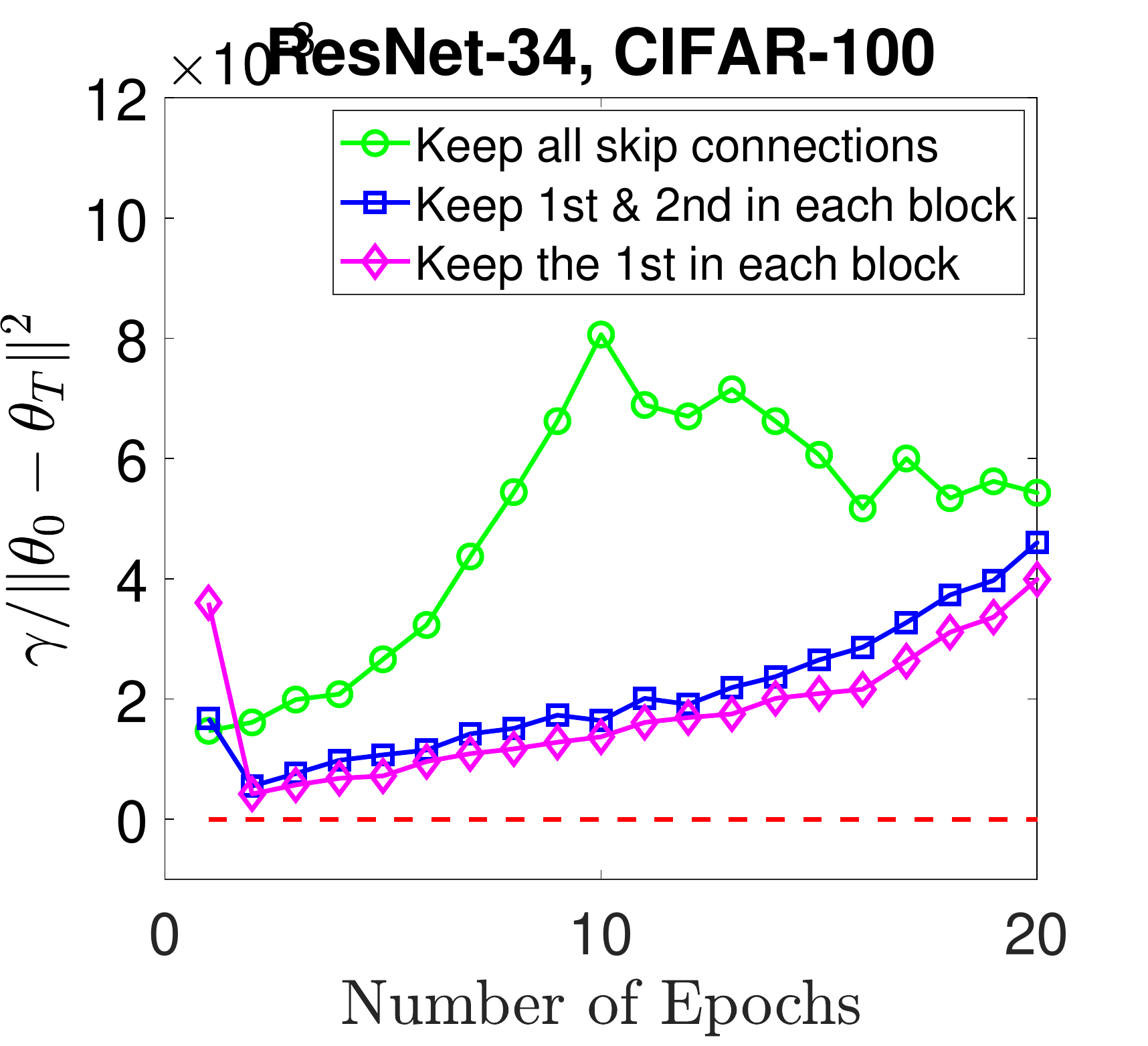}
	\vspace{-6mm}
	\caption{\small{Training ResNet-34 with and without skip-connections.}}\label{fig: 5} 
%	\vspace{-2mm}
\end{figure}

Fig. \ref{fig: 5} shows the ResNet-34 training results. One can see that the trainings that keep more skip-connections achieve a faster convergence, and the corresponding optimization trajectories obey the regularity principle with a larger $\gamma$. These observations are consistent with our theoretical understanding of the regularity principle and they imply that skip-connection helps to regularize the optimization trajectory. We also present the ResNet-18 training results in  \Cref{sec: app: 3}, where one can make very similar observations.

{ \textbf{Experiment setup:}  We also train a U-Net on the CIFAR-10 and CIFAR-100 datasets and consider the settings: 1) keep all the skip-connections; 2) keep the last 3,2,1 skip-connections, respectively and 3) remove all the skip-connections. We apply the standard SGD optimizer with a fixed initialization point, a mini-batch size 128 and a constant learning rate $\eta=0.005$ to train these networks for 150 epochs, and we use the MSE loss.}

\begin{figure}[htbp]%[bth]
	\vspace{-2mm}
	\centering
	\includegraphics[width=0.24\textwidth,height=0.2\textwidth]{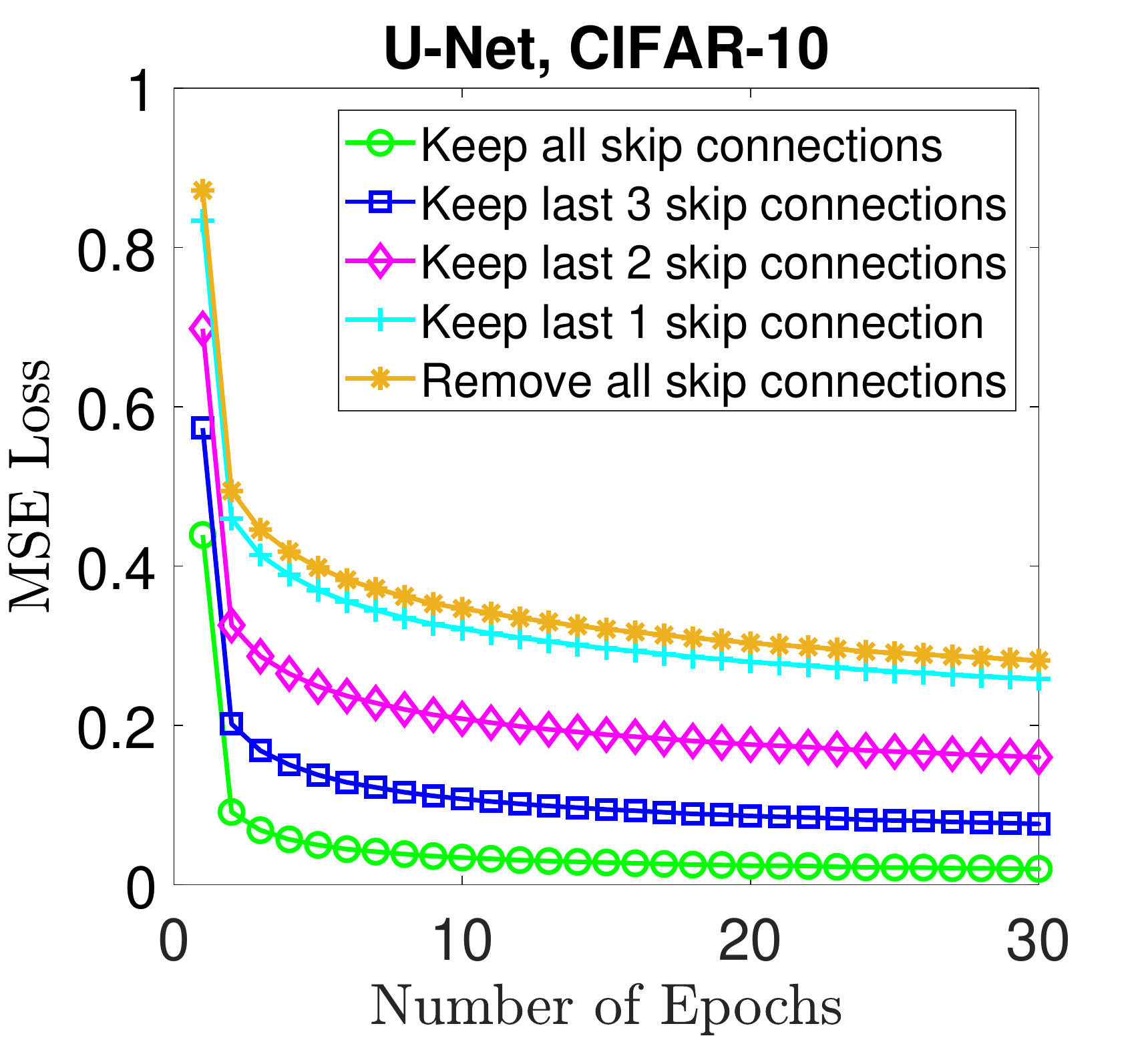}
	\includegraphics[width=0.24\textwidth,height=0.2\textwidth]{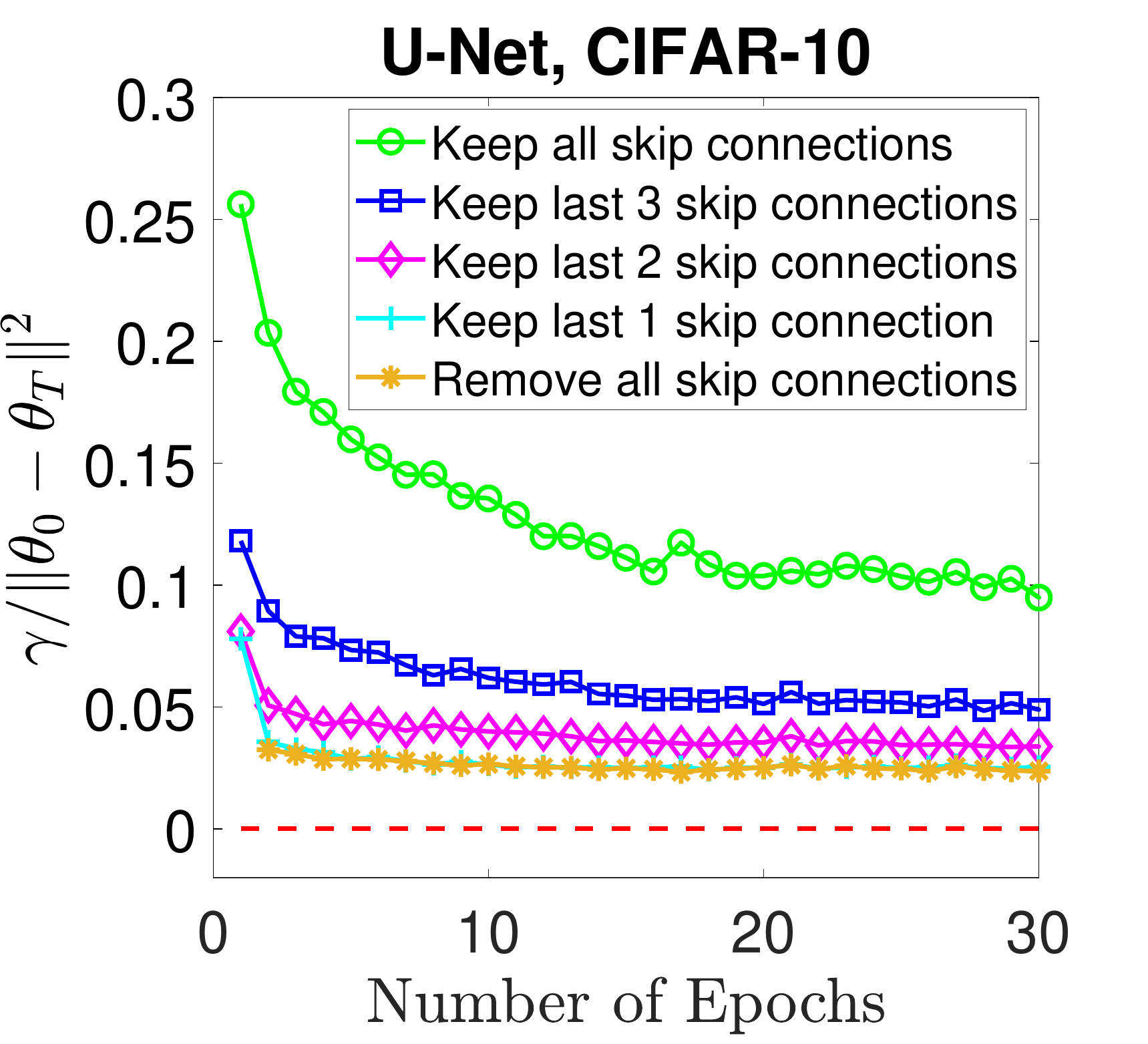}
	\includegraphics[width=0.24\textwidth,height=0.2\textwidth]{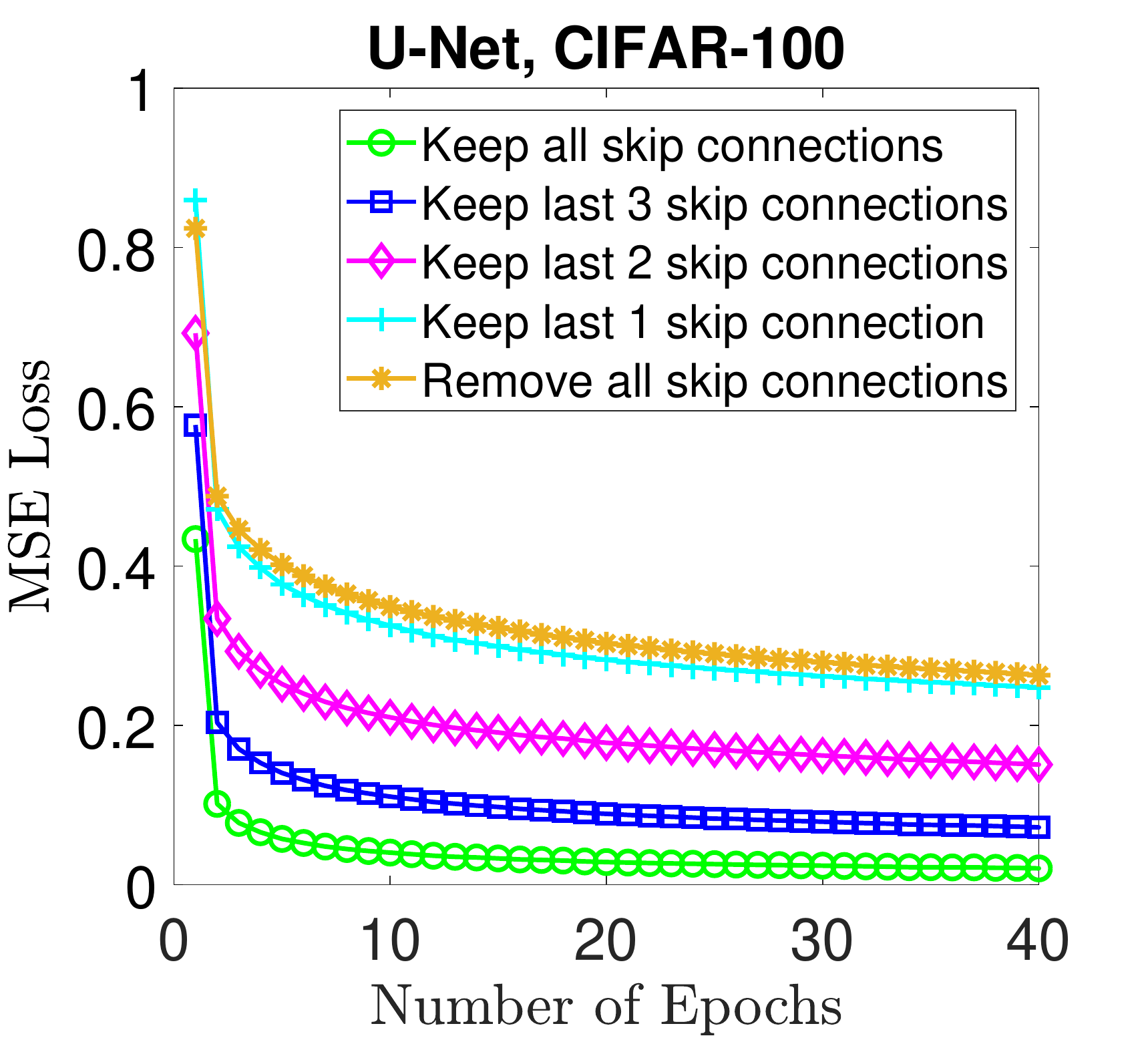}
	\includegraphics[width=0.24\textwidth,height=0.2\textwidth]{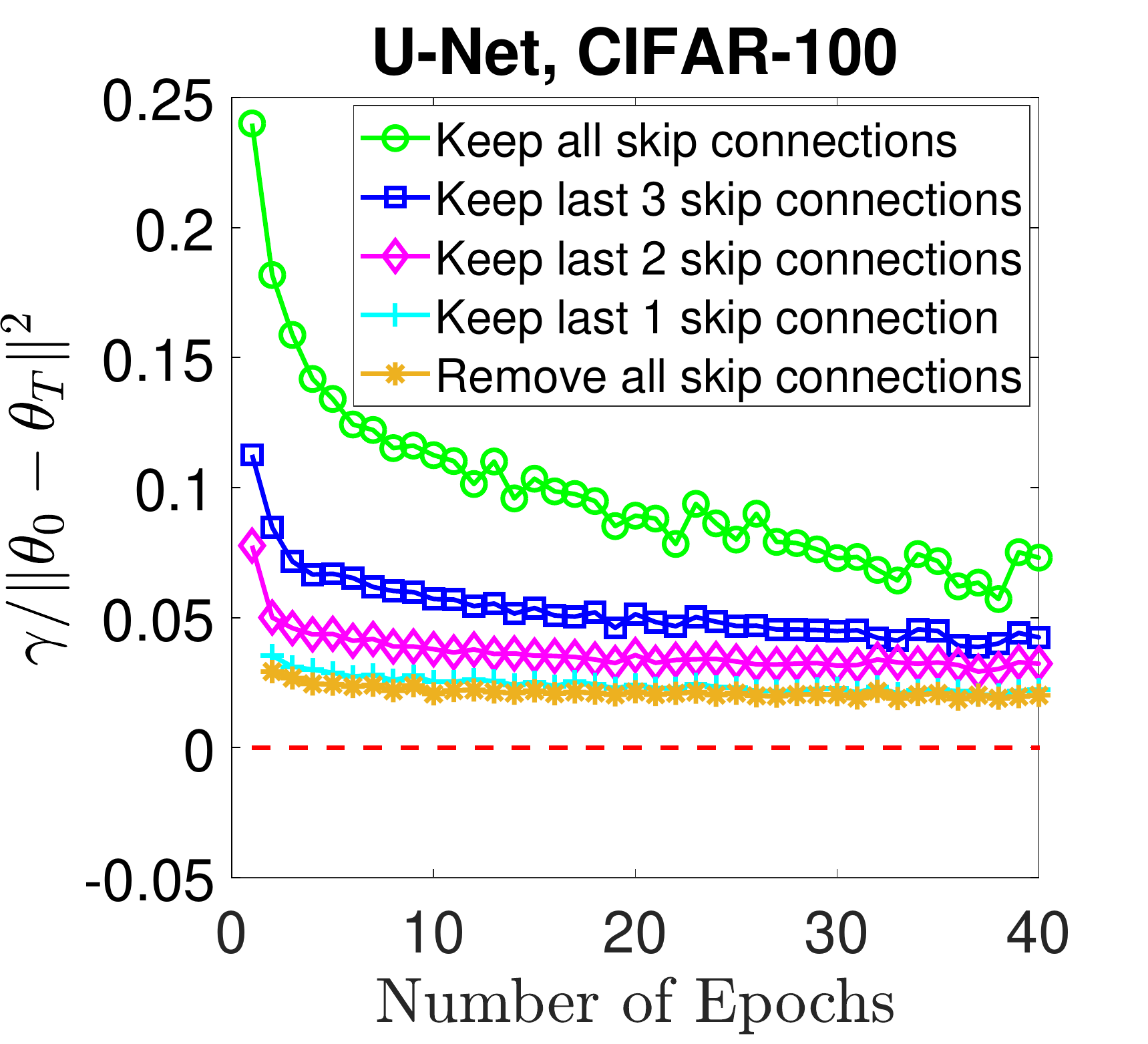}
	\vspace{-6mm}
	\caption{\small{Training U-Net with and without skip-connections.}}\label{fig: 103} 
%	\vspace{-2mm}
\end{figure}

The training results are shown in Fig. \ref{fig: 103}, where one can make similar observations as those above: the trainings that keep more skip-connections achieve a faster convergence and the corresponding optimization trajectories obey the regularity principle with a larger $\gamma$. Therefore, the regularity principle is able to characterize and quantify the regularization effect of skip-connections on DNN training trajectories. 

%In specific, regarding the trainings of the Resnet-18 on Cifar10, the training with one skip connection in each block has comparable performance to that of the training with all skip connections, and both trainings have comparable $\gamma$ values. In the trainings of the Resnet-18 on the more complex Cifar100 dataset, the training with more skip connections achieves a slightly faster convergence, and the corresponding $\gamma$ value  is larger than that in the training with one skip connection in each block. 
%Regarding the trainings of the deeper Resnet-34, it can be seen that skip connections significantly accelerate the training on both datasets. In particular, trainings are faster with more number of skip connections in each block, and the $\gamma$ values are larger in the trainings with more skip connections in each block. All these observations are consistent with  the theoretical implications of the $\gamma$-optimization principle in \Cref{thm: convergence}. 

%\vspace{-1mm}
\section{Experiments on Optimization-level Training Techniques}
%\vspace{-1mm}
In this section, we explore the effect of stochastic optimizers on the regularity principle. 

\textbf{Experiment setup:} We train the ResNet-18, 34 and the VGG-11, 16 networks  using SGD, SGD with momentum and Adam, respectively. We remove all the dropout layers in the VGG networks. We apply a fixed initialization point, a constant learning rate $\eta=0.001$ and batch size 128 to all the optimizers and train these networks on the CIFAR-10 and CIFAR-100 datasets, respectively, and we use the cross-entropy loss. We set the momentum to be 0.5 for the SGD with momentum and set $\beta_1 = 0.9, \beta_2 = 0.999, \epsilon= 10^{-2}$ for the Adam. 

\begin{figure}[htbp]%[bth]
	\vspace{-2mm}
	\centering
	\includegraphics[width=0.24\textwidth,height=0.2\textwidth]{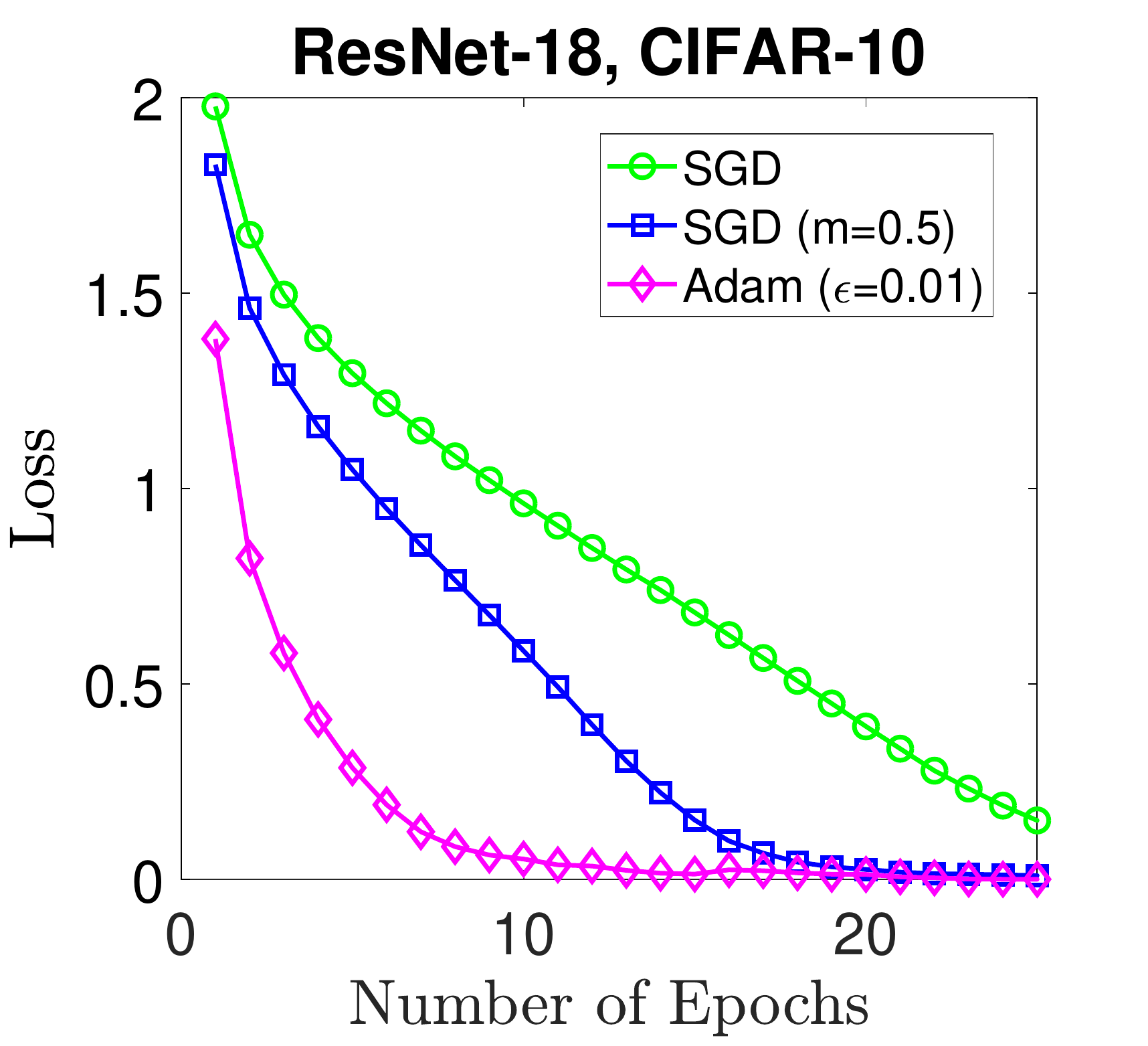}
	\includegraphics[width=0.24\textwidth,height=0.2\textwidth]{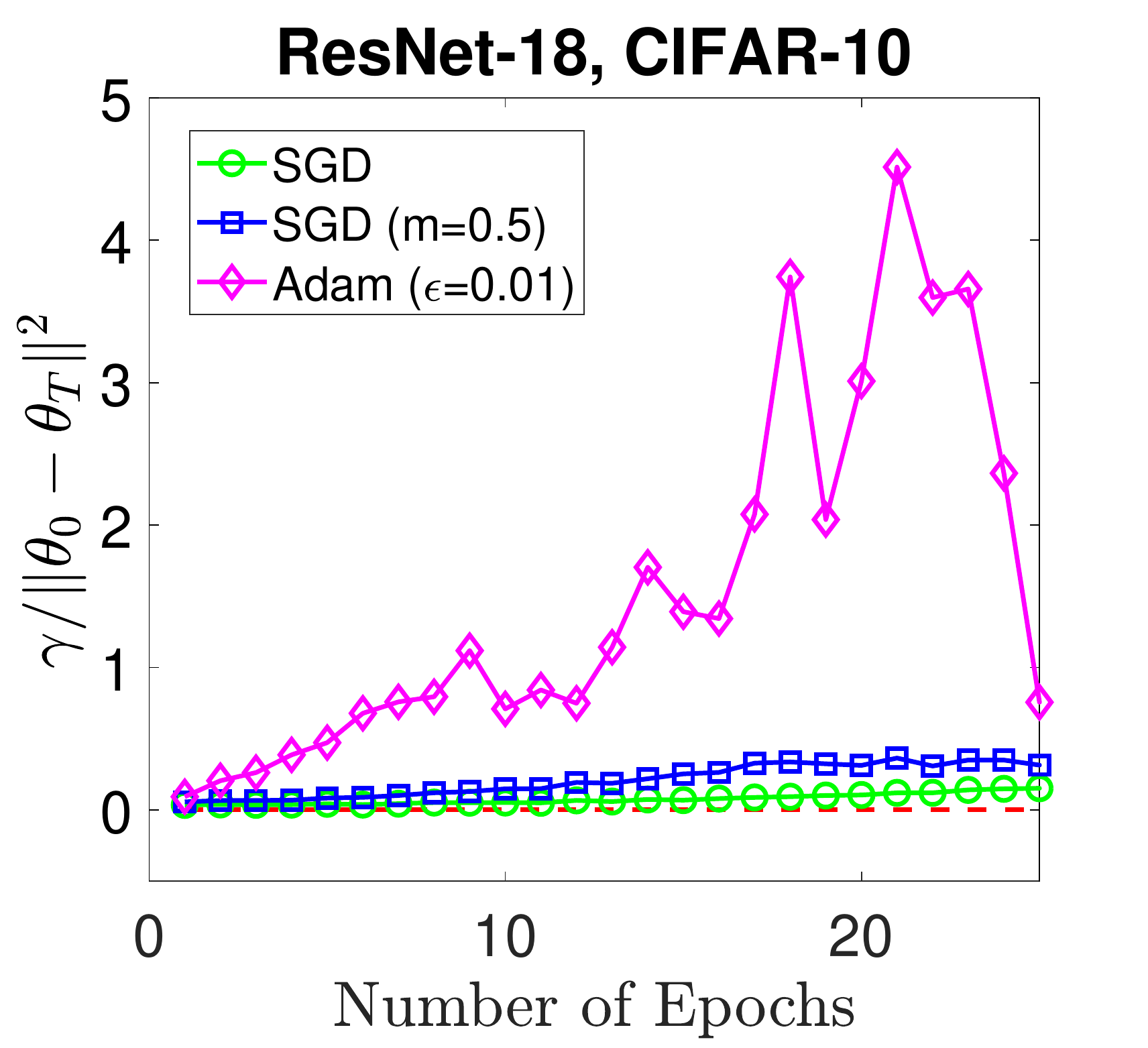}
	\includegraphics[width=0.24\textwidth,height=0.2\textwidth]{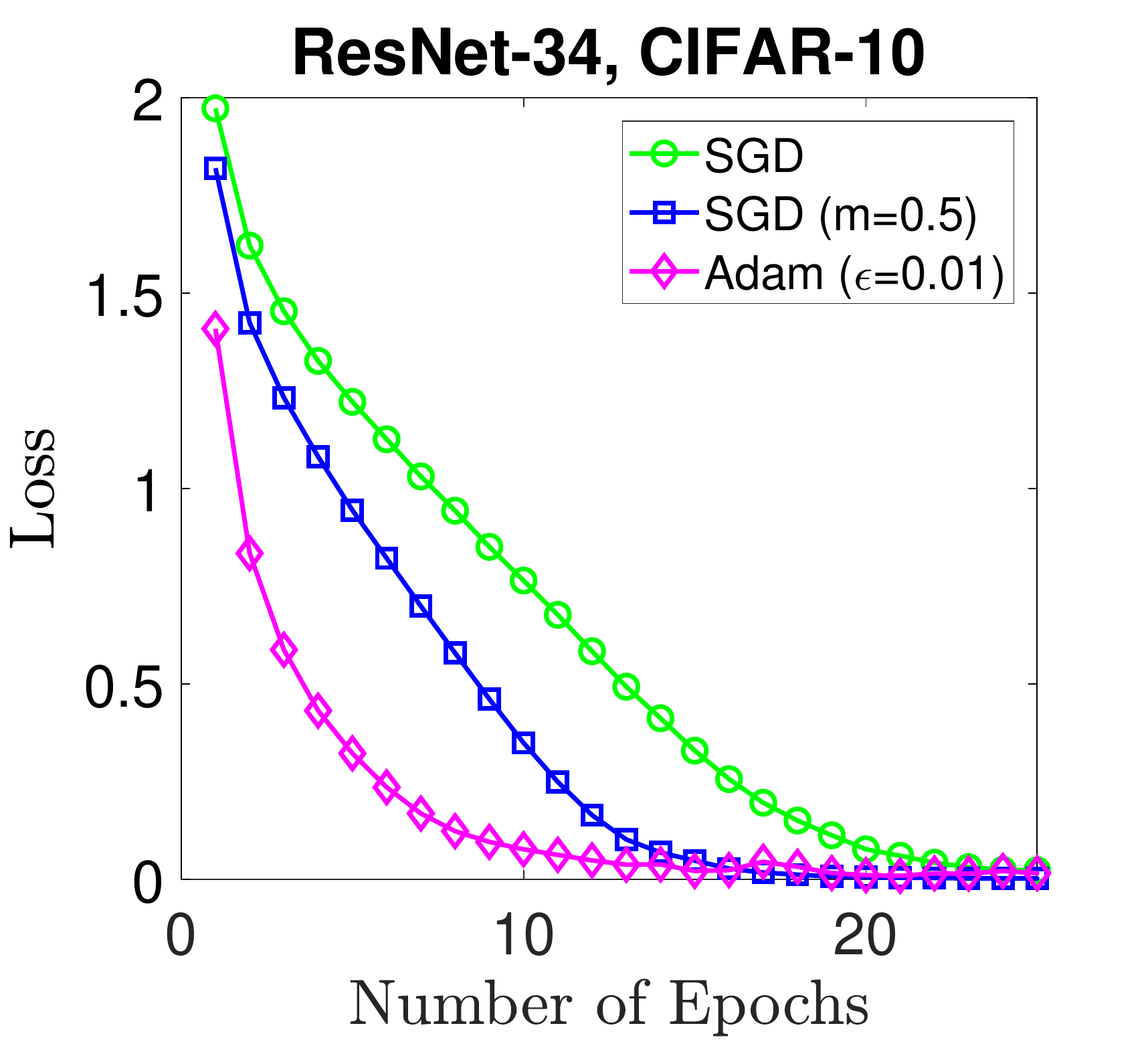}
	\includegraphics[width=0.24\textwidth,height=0.2\textwidth]{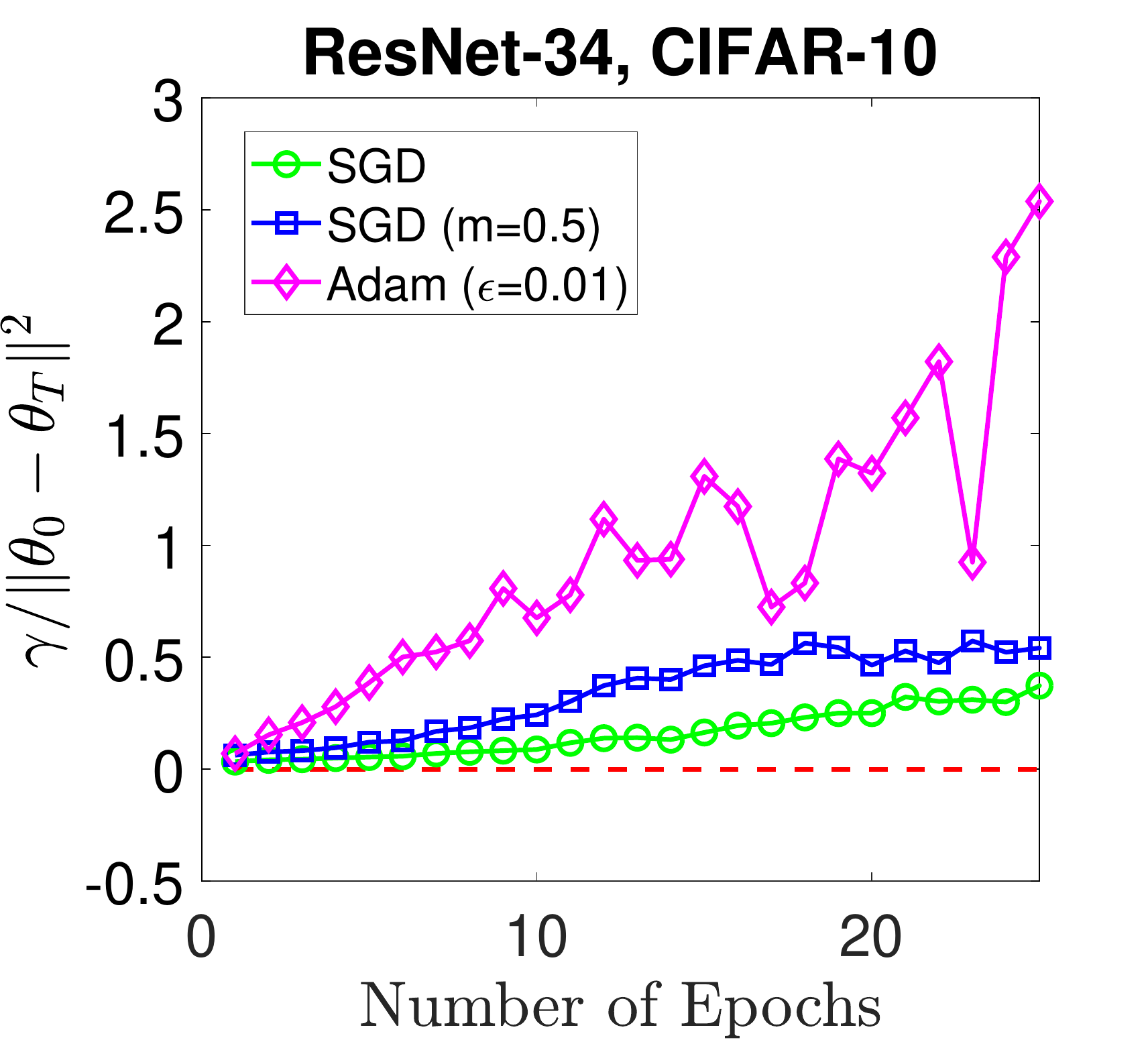}
	\vspace{-6mm}
	\caption{\small{Training ResNets with different optimizers on CIFAR-10.}}\label{fig: 6} 
\end{figure}

\begin{figure}[htbp]%[bth]
	\vspace{-2mm}
	\centering
	\includegraphics[width=0.24\textwidth]{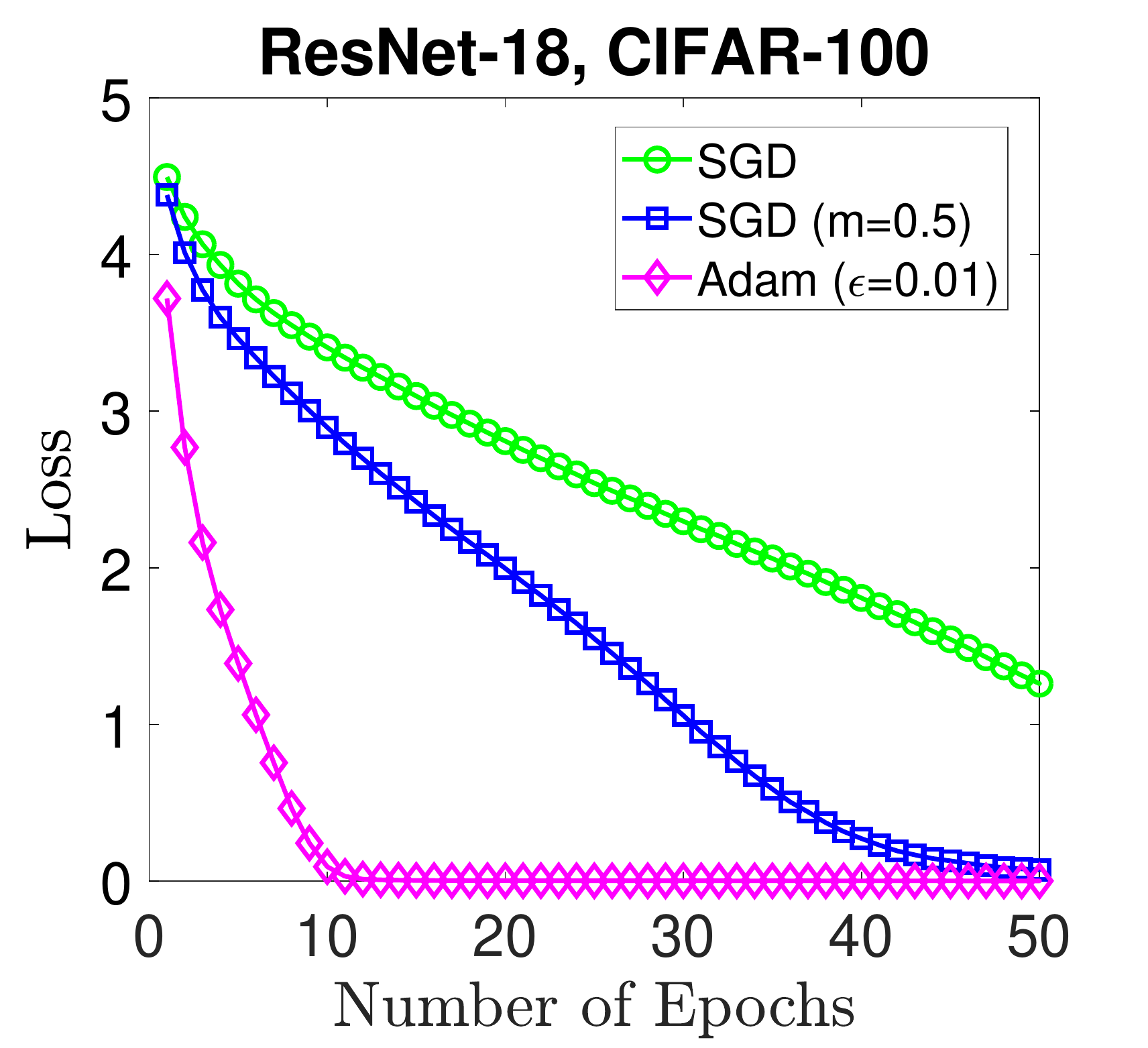}
	\includegraphics[width=0.24\textwidth]{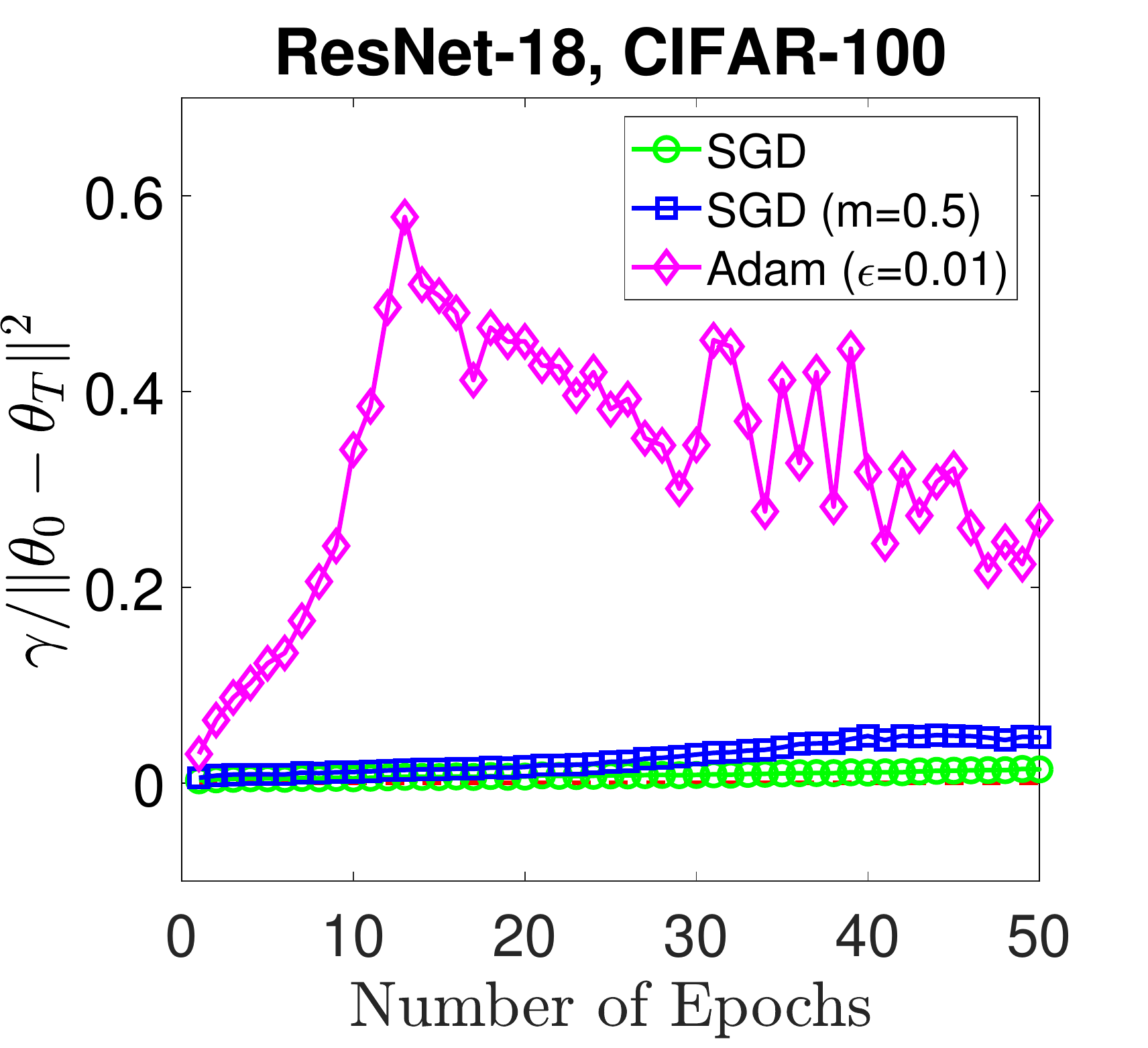}
	\includegraphics[width=0.24\textwidth]{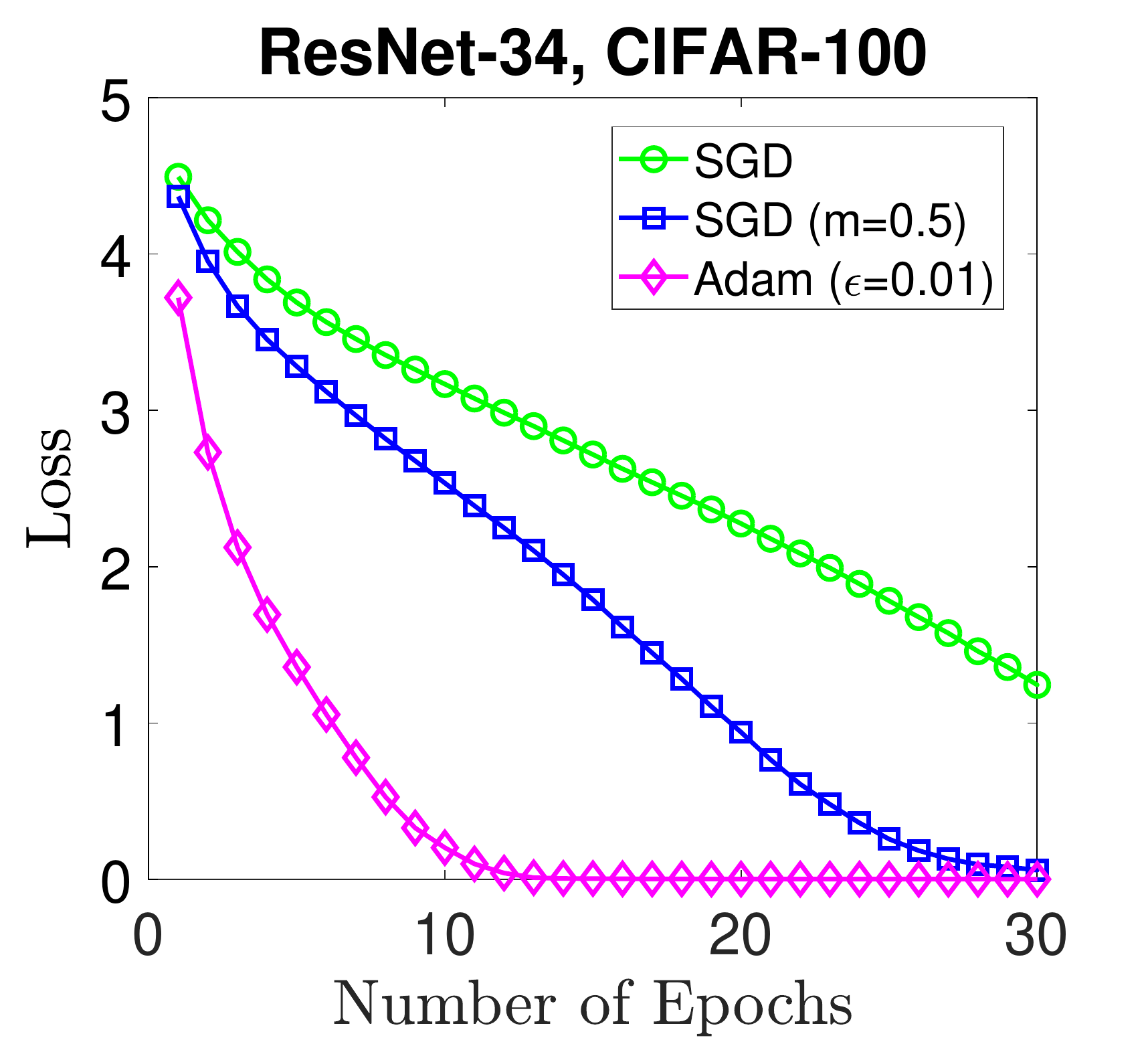}
	\includegraphics[width=0.24\textwidth]{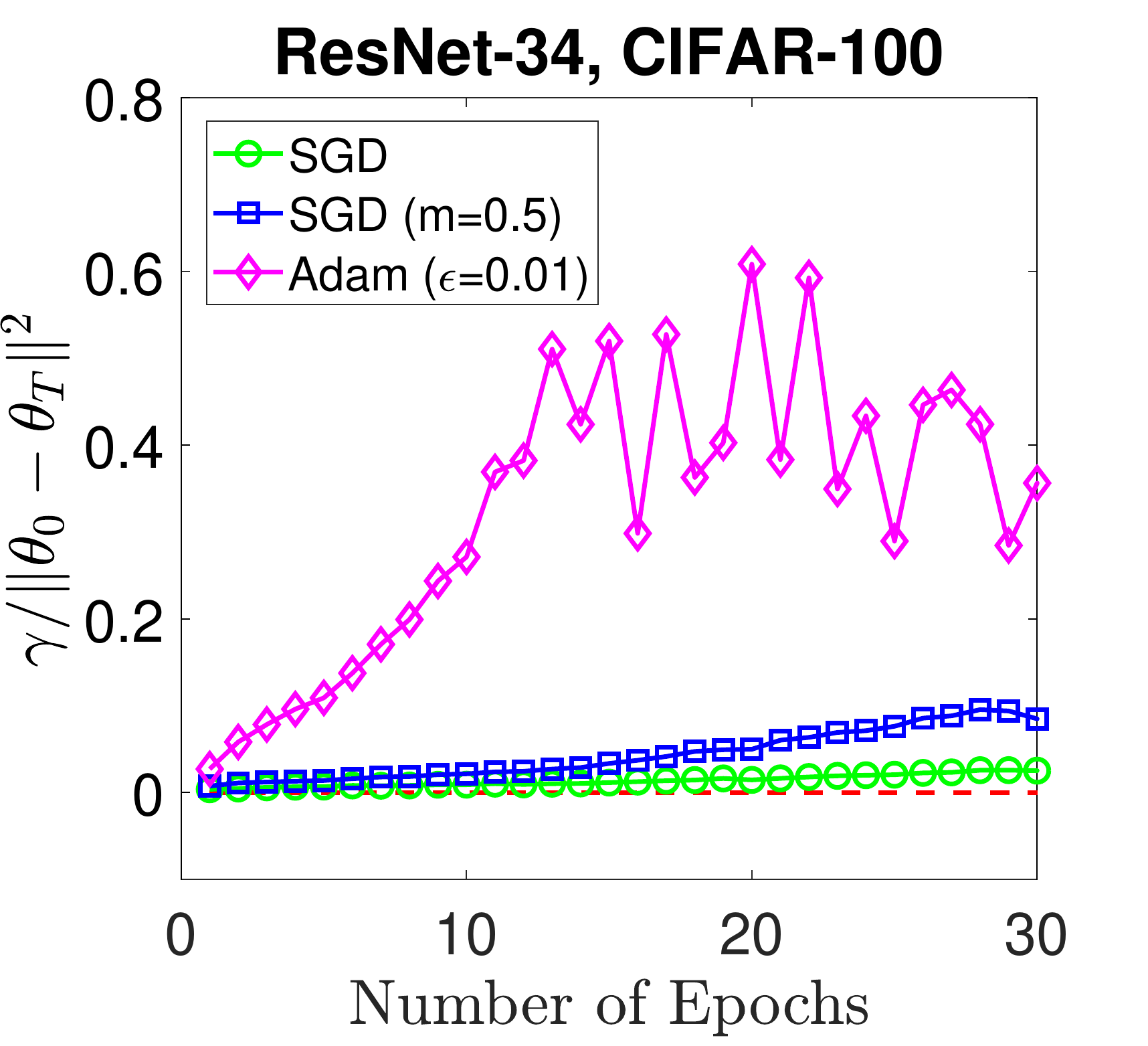}
	\vspace{-6mm}
	\caption{\small{Training ResNets with different optimizers on CIFAR-100.}}\label{fig: 7} 
\end{figure}

Fig. \ref{fig: 6} presents the ResNet training results on the CIFAR-10 dataset. 
In all these trainings, the Adam algorithm achieves the fastest convergence and is followed by the SGD with momentum, whereas the vanilla SGD achieves the slowest convergence. Furthermore, it can be seen that the optimization trajectories driven by Adam obey the regularity principle with the largest $\gamma$, and the trajectories driven by SGD types of algorithms is less regularized by the regularity principle. All these observations corroborate the theoretical implication of the regularity principle and show that advanced optimizers lead to well-regularized optimization trajectories. We further present the ResNet training results on the CIFAR-100 dataset in Fig. \ref{fig: 7} below and present all the VGG training results in \Cref{sec: app: 4}, where one can make the same observations and conclusions. 

%the $\gamma$ curves are above zero and therefore demonstrates the validity of the $\gamma$-optimization principle. Also, it can be seen that the SGD with momentum trainings achieve significantly faster convergence speed than that achieved by the SGD trainings, and the Adam trainings converge the fastest. Regarding the $\gamma$, it can be seen that the SGD trainings obey the optimization principle with the smallest $\gamma$, which is further enlarged in the SGD with momentum trainings. Moreover, the Adam trainings obey the optimization principle with the largest $\gamma$. These observations are consistent with item 3 of \Cref{thm: convergence}, where a larger $\gamma$ implies faster convergence of the training.

%The training results on the Cifar100 dataset are shown in \Cref{fig: 7}, where one can make very similar observations as those on Cifar10. In particular, for the Adam training of VGG-16 on Cifar100, the sign of $\gamma$ fluctuates after 50 epochs, which is possibly caused by the stochastic pre-conditioner in the Adam update. In addition to these experiments, we also visualize the optimization paths of different optimizers in \Cref{append: 3.2}, where one can see that the directions of SGD updates are very different from those of the updates generated by the SGD with momentum and Adam. These experiments show that the $\gamma$-optimization principle characterizes the impact of optimizers on DNN training. 

{\textbf{Experiment setup:}  Next, we train a U-Net on the CIFAR-10 and CIFAR-100 datasets using SGD, SGD with momentum and Adam, respectively, and use the MSE loss. We apply a fixed initialization, a constant learning rate $\eta=0.001$ and a mini-batch size 128 to all the optimizers. We set the momentum to be 0.5 for SGD with momentum and $\beta_1 = 0.9, \beta_2 = 0.999, \epsilon= 10^{-2}$ for the Adam.}
\begin{figure*}[htbp]%[bth]
	\centering
	\vspace{-2mm}	
	\includegraphics[width=0.24\textwidth,height=0.2\textwidth]{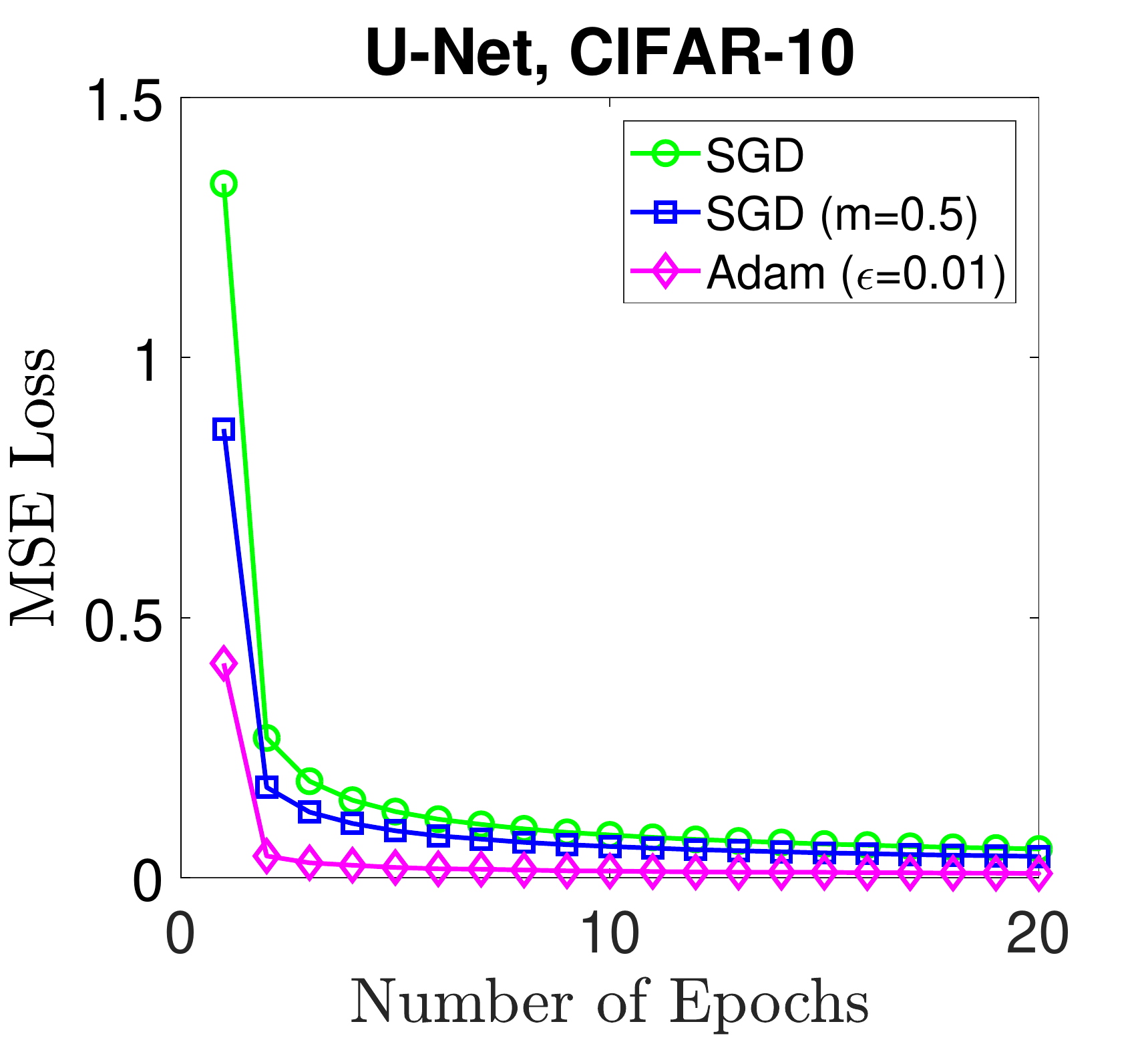}
	\includegraphics[width=0.24\textwidth,height=0.2\textwidth]{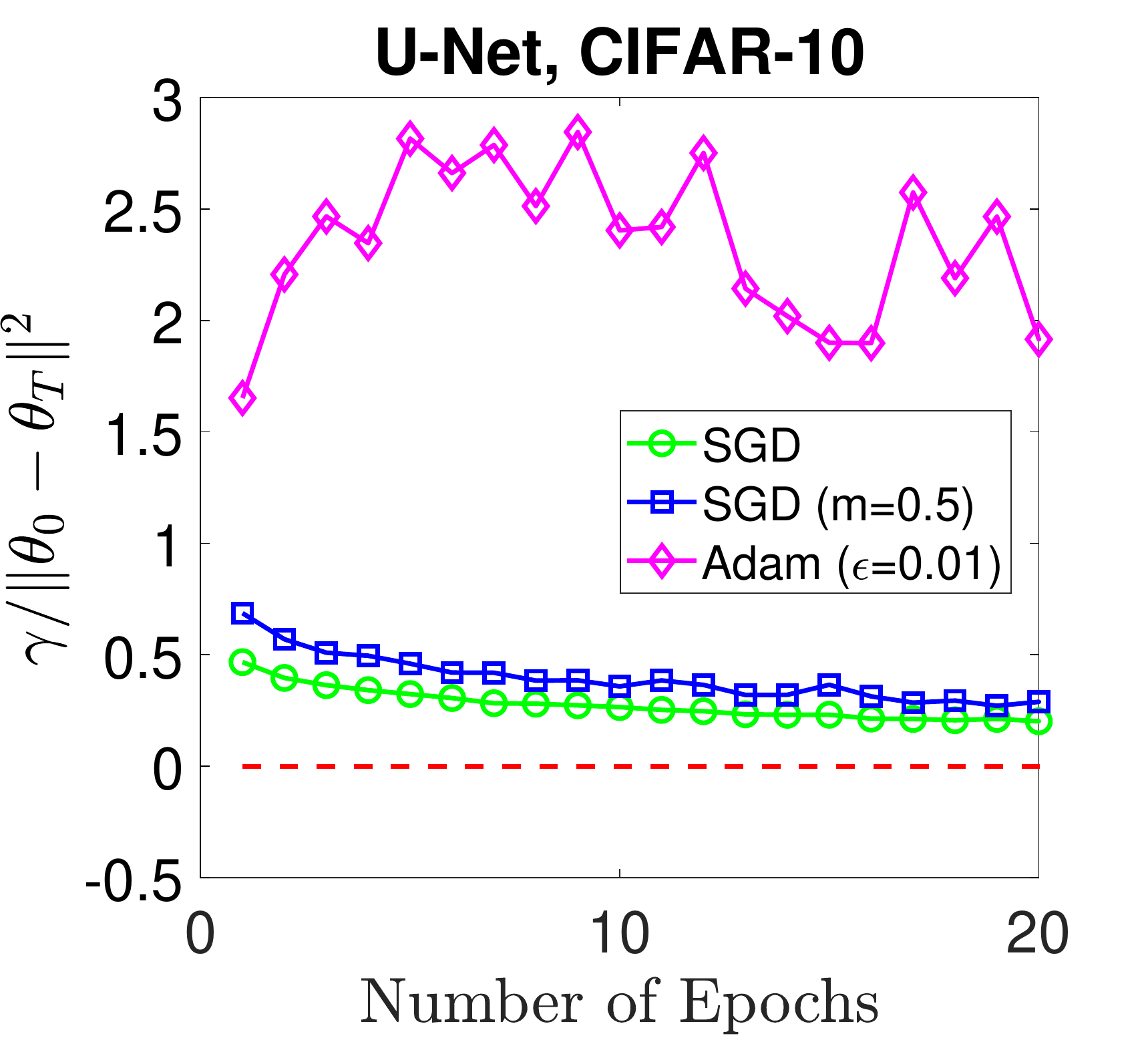}
	\includegraphics[width=0.24\textwidth,height=0.2\textwidth]{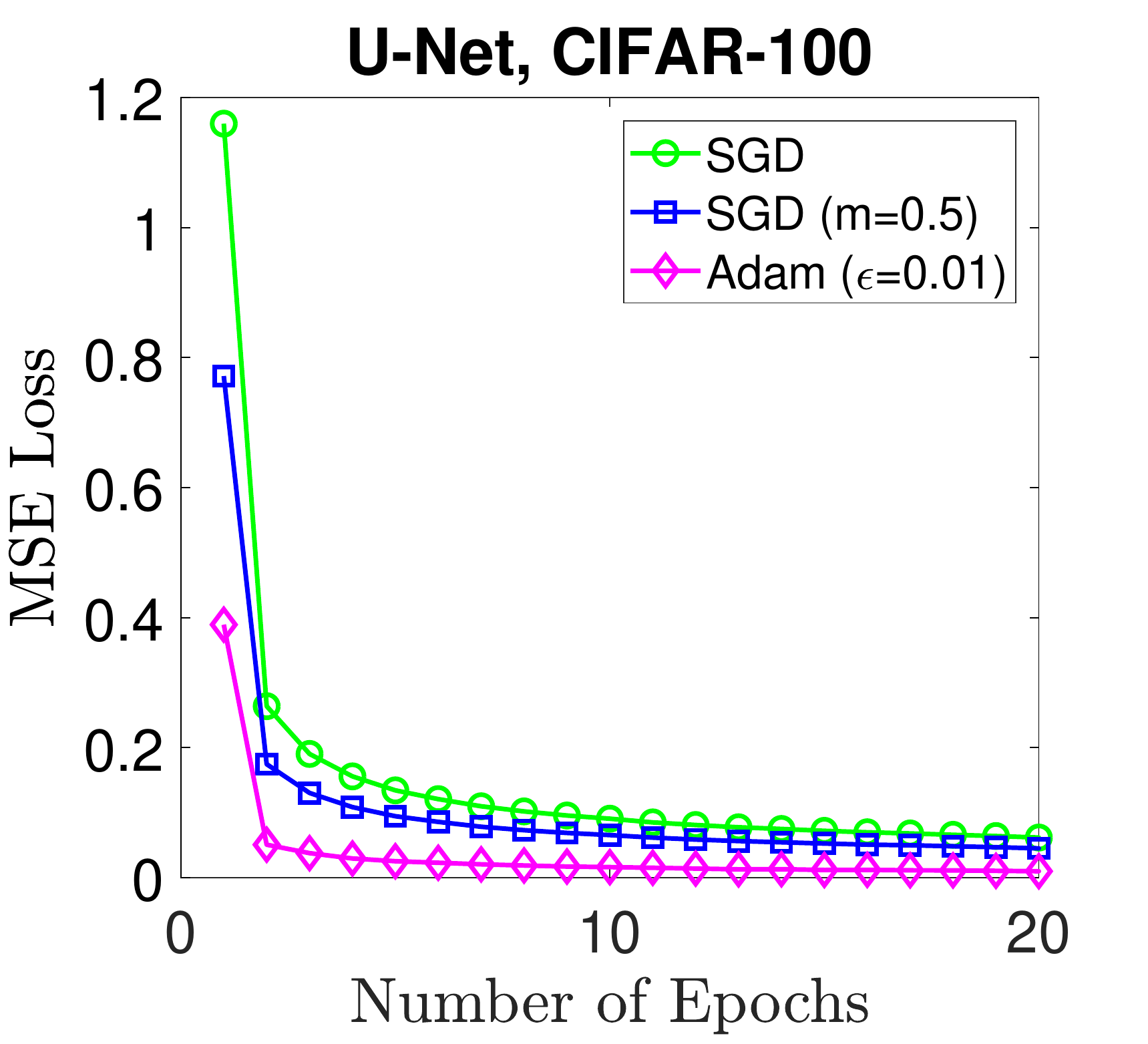}
	\includegraphics[width=0.24\textwidth,height=0.2\textwidth]{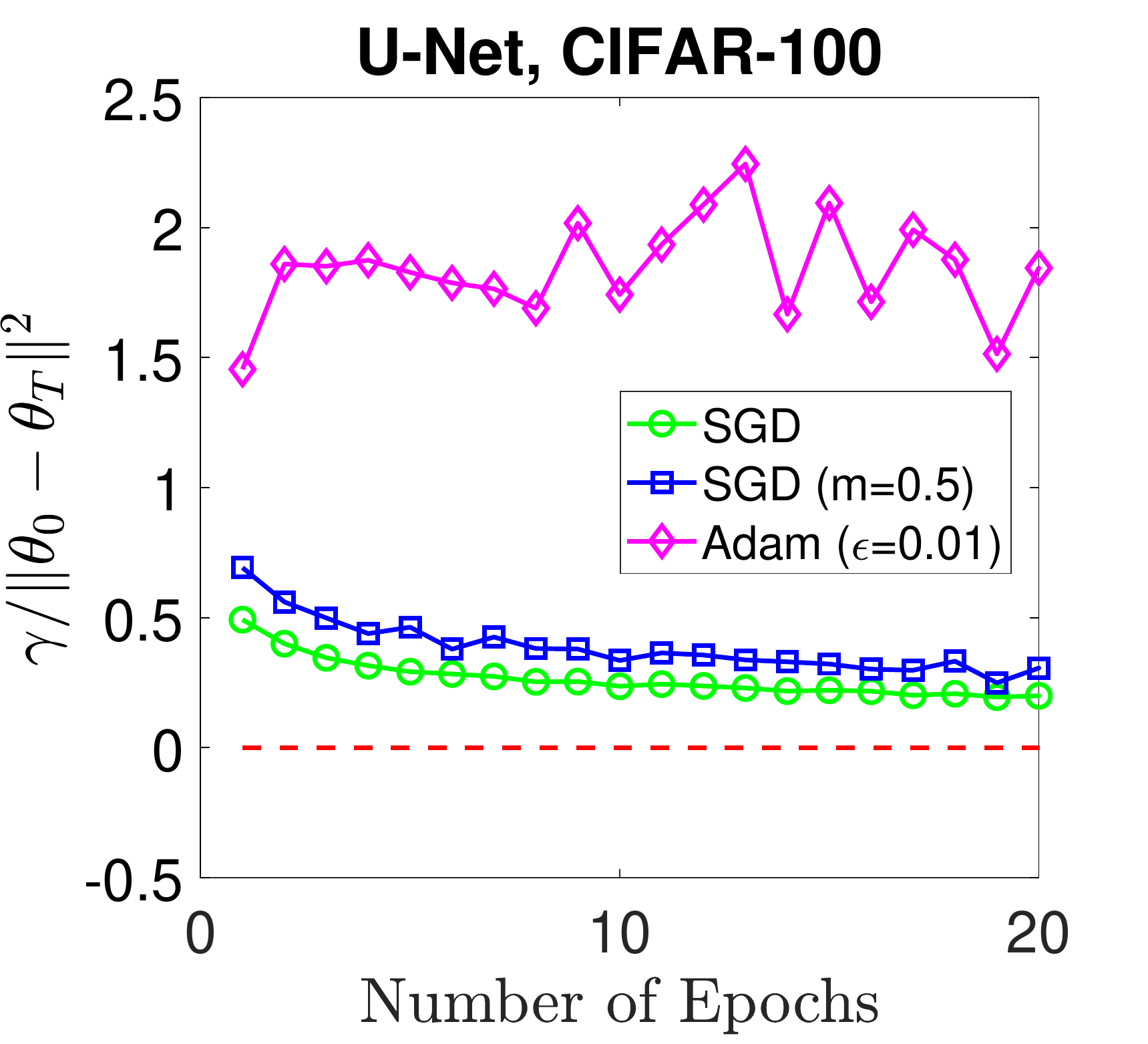}
	\vspace{-1mm}
	\caption{\small{Training U-Net with different optimizers.}}\label{fig: 104} 
	\vspace{-2mm}
\end{figure*}

Fig. \ref{fig: 104} shows the training results, where one can see that the Adam algorithm still achieves the fastest convergence and the corresponding optimization trajectories obey the regularity principle with the largest $\gamma$. As a comparison, SGD and SGD with momentum converge much slower and their optimization trajectories obey the regularity principle with a smaller $\gamma$. These experiments are also consistent with our theoretical understanding of the regularity principle. 

%In conclusion, the experiments in this subsection demonstrate that different stochastic optimizers regularize the DNN optimization trajectory at different levels, and they all obey the regularity principle.

%\vspace{-5pt}
\section{Conclusions}
%\vspace{-5pt}
In this paper, we propose a regularity principle for general stochastic algorithms in nonconvex optimization. The regularity principle provides convergence guarantees to stochastic optimization and achieves a sub-linear convergence rate that scales inverse proportionally to its regularization parameter $\gamma$. Through extensive DNN training experiments on a variety of deep models, we show that many practical DNN training trajectories obey the regularity principle reasonably well, and the regularization parameter $\gamma$ provides a metric that quantifies the effect of different training techniques on DNN optimization. In the future work, we expect that such an optimization-level regularity principle can be exploited to develop improved training techniques for deep learning and can help understand the generalization performance of DNN under different training techniques.

%\section*{Acknowledgment}

{%\small
	\bibliographystyle{abbrv}
	\bibliography{eccv}
}

%\begin{thebibliography}{00}
%\end{thebibliography}

%\appendix
\appendices
%\clearpage
%\onecolumn

\section{Proof of \Cref{thm: convergence}}
	Note that the SA update rule implies that
	\begin{align}
	\|\theta_{k+1} - \theta_T\|^2 &= \|\theta_{k} - \eta U(\theta_{k}; z_{\xi_k}) - \theta_T\|^2\nonumber\\
	&= \|\theta_{k} - \theta_T\|^2 + \eta^2 \|U(\theta_{k}; z_{\xi_k})\|^2 \nonumber\\
	&\quad- 2\eta\inner{\theta_k-\theta_T}{U(\theta_{k}; z_{\xi_k})} \nonumber\\
	&\le \|\theta_{k} - \theta_T\|^2 + \eta^2 \|U(\theta_{k}; z_{\xi_k})\|^2 \nonumber\\
	&\quad- \eta^2 \|U(\theta_{k}; z_{\xi_k})\|^2  \nonumber\\
	&\quad- 2\eta\gamma\big(\ell(\theta_k;z_{\xi_k}) - \inf_{\theta} \ell(\theta;z_{\xi_k})\big), \label{eq: 1}
	\end{align}
	where the last inequality follows from the regularity principle in \Cref{def: principle}. Note that the SA algorithm adopts the random sampling with reshuffle scheme. Summing \eqref{eq: 1} over the $B$-th epoch (i.e., $k=nB, nB+1,...,n(B+1)-1 $) gives that 
	\begin{align*}
	\|\theta_{n(B+1)} - &\theta_T\|^2 \le  \|\theta_{nB} - \theta_T\|^2 \nonumber\\
	&\quad- 2\eta\gamma\sum_{k=nB}^{n(B+1)-1}\big(\ell(\theta_k;z_{\xi_k}) - \inf_{\theta} \ell(\theta;z_{\xi_k})\big). 
	\end{align*} 
	Further telescoping over the epoch index, we obtain that for all $B$
	\begin{align}
	\|\theta_{nB} - &\theta_T\|^2 \le \|\theta_{0} - \theta_T\|^2 \nonumber\\
	&\quad-2\eta\gamma \sum_{P=0}^{B-1}\sum_{k=nP}^{n(P+1)-1}\big(\ell(\theta_k;z_{\xi_k}) - \inf_{\theta} \ell(\theta;z_{\xi_k})\big). \label{eq: 4}
	\end{align}
	Therefore, for all $T=nB$, the above inequality further implies that
	\begin{align}
	0 &\le \|\theta_{0} - \theta_T\|^2 -2\eta\gamma \sum_{k=0}^{T-1}\big(\ell(\theta_k;z_{\xi_k}) - \inf_{\theta} \ell(\theta;z_{\xi_k})\big) \nonumber\\
	&= \|\theta_{0} - \theta_T\|^2 -2\eta\gamma \Big( \sum_{k=0}^{T-1} \ell(\theta_k;z_{\xi_k}) - T \inf_{\theta} f(\theta; \mathcal{Z}) \Big),
	\end{align}
	where the last equality follows from the over-parameterization of the model. Rearranging the above inequality yields that
	\begin{align}
	\frac{1}{T}\sum_{k=0}^{T-1} \ell(\theta_k;z_{\xi_k}) - \inf_{\theta} f(\theta; \mathcal{Z})\le \frac{\|\theta_{0} - \theta_T\|^2}{2\eta\gamma T}. \nonumber
	\end{align}

%\newpage
\section{Additional Experiments}

\subsection{Effect of Neuron Activation Function on Regularity Principle}\label{sec: app: 1}
\begin{figure}[htbp]%[bth]
	\vspace{-2mm}
	\centering
	\includegraphics[width=0.24\textwidth,height=0.2\textwidth]{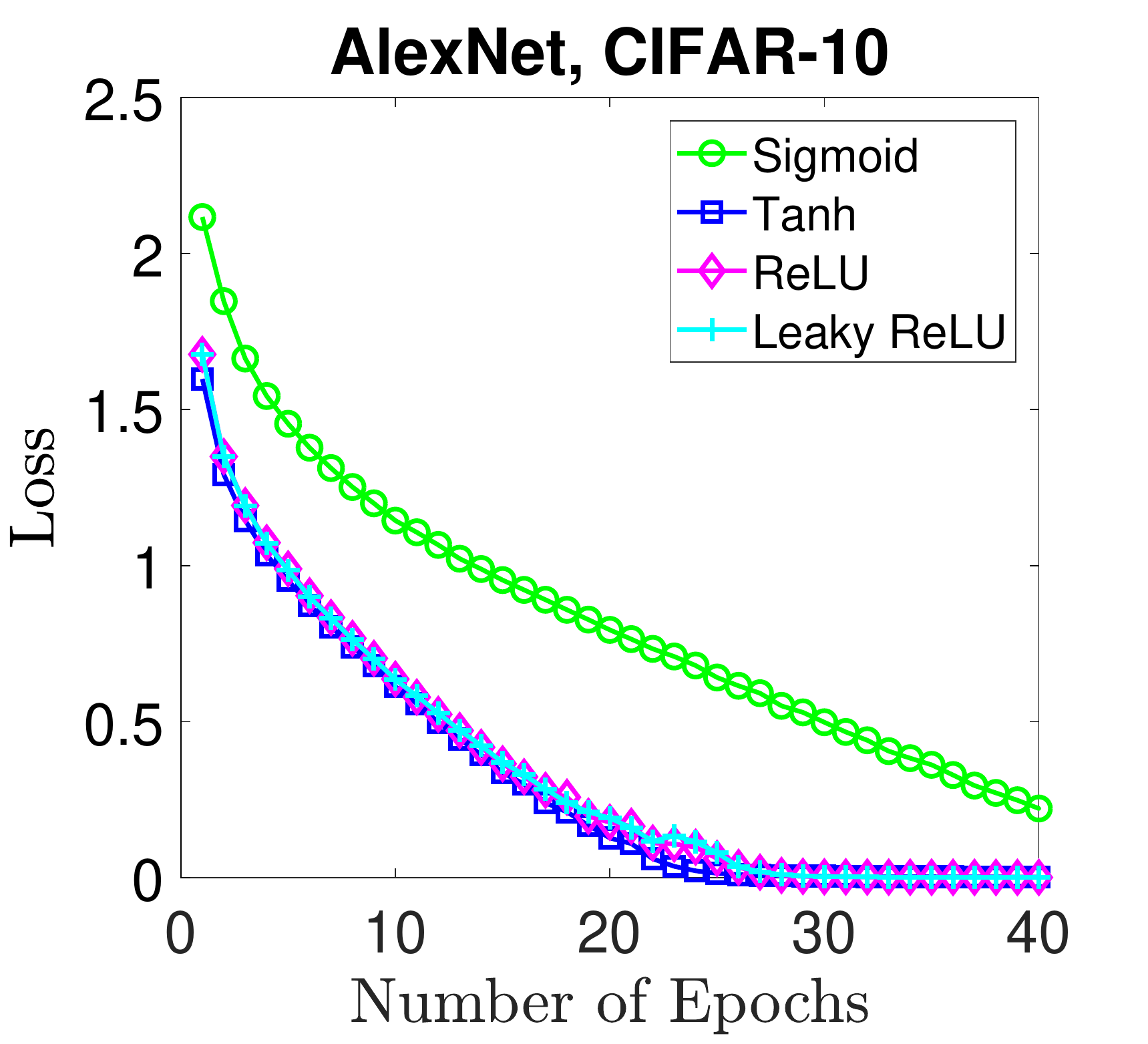}	
	\includegraphics[width=0.24\textwidth,height=0.2\textwidth]{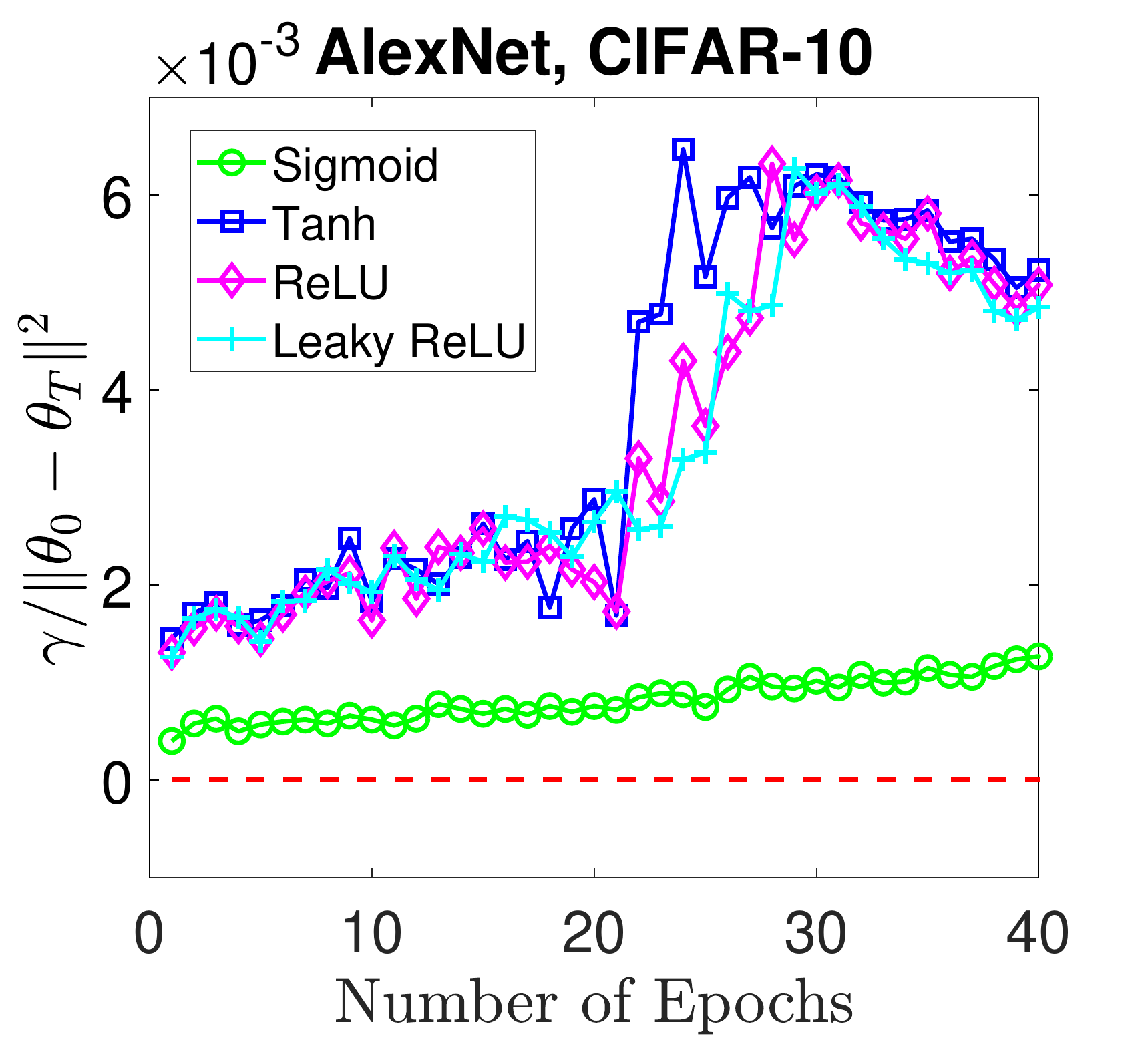}
	\includegraphics[width=0.24\textwidth,height=0.2\textwidth]{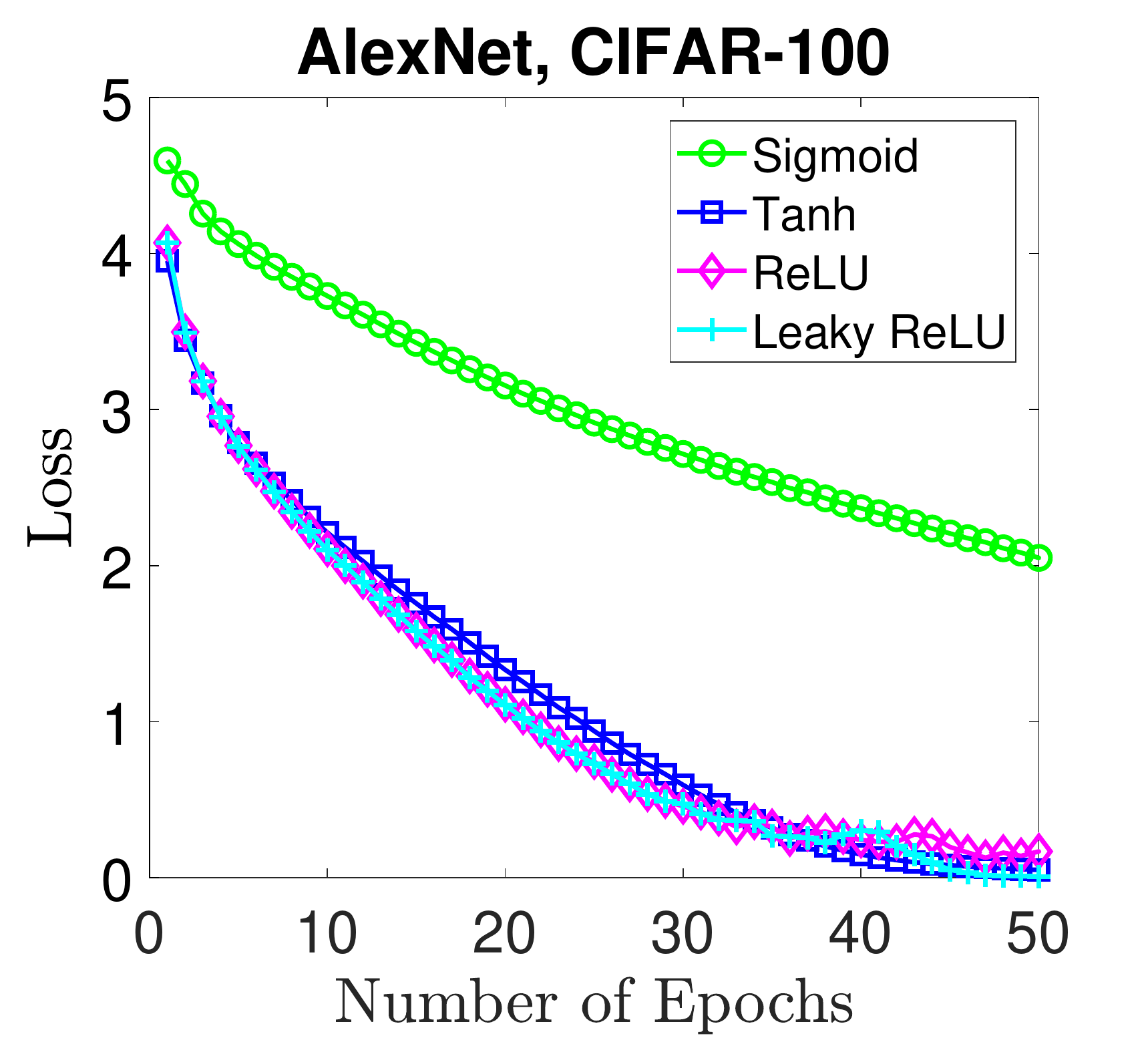}
	\includegraphics[width=0.24\textwidth,height=0.2\textwidth]{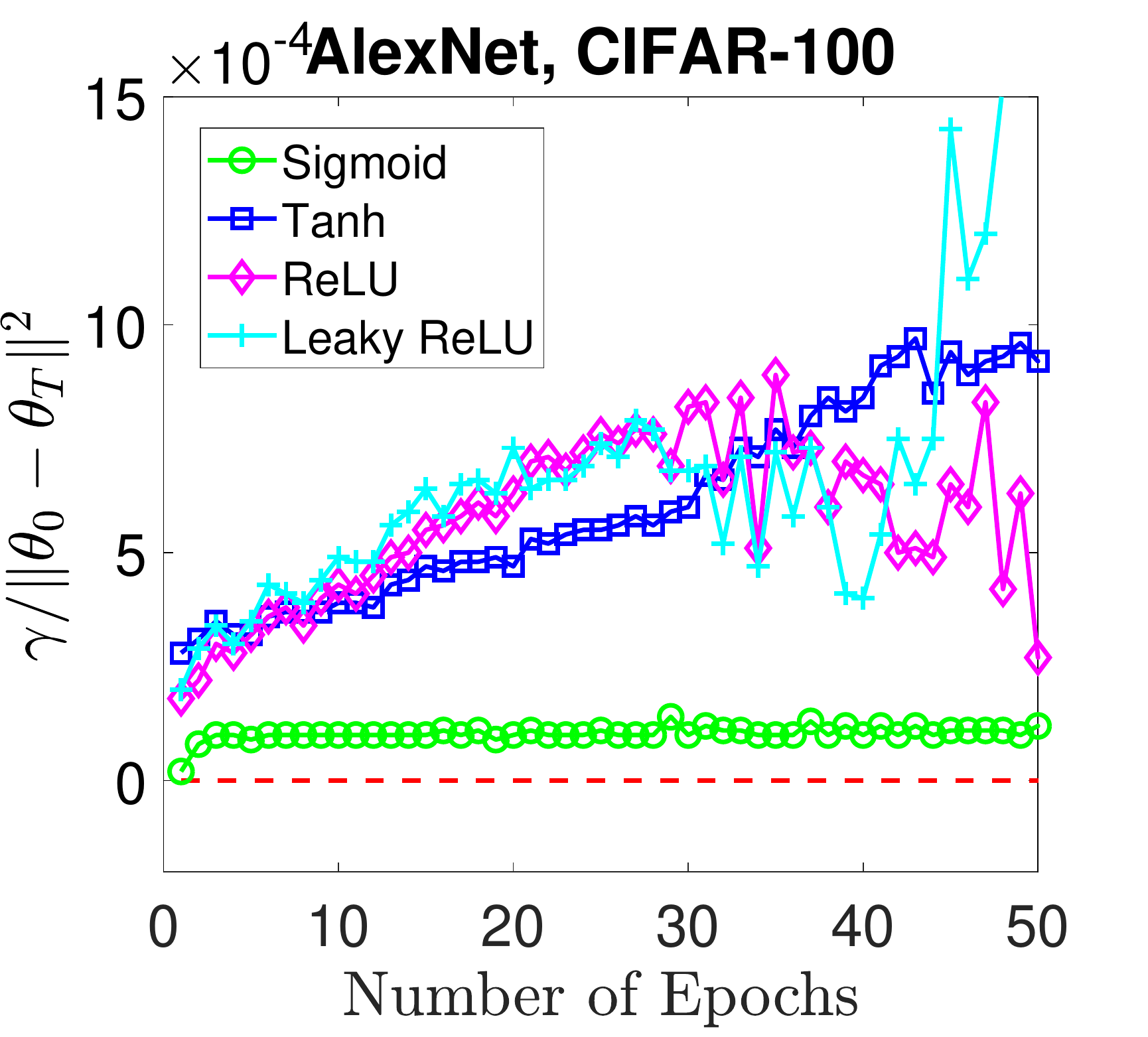}
		\vspace{-6mm}
	\caption{\small{Training AlexNet with different activation functions.}}\label{fig: app: 1} 
	\vspace{-4mm}
\end{figure}

Regarding the trainings of the AlexNet in Fig. \ref{fig: app: 1}, we observe that the trainings with the sigmoid activation function converge much slower than those with the other activation functions, and the parameter $\gamma$ of the regularity principle under the sigmoid activation is much smaller than those under the other activation functions. These observations are consistent with \Cref{thm: convergence}. 

\subsection{Effect of Batch Normalization on Regularity Principle}\label{sec: app: 2}
%On CIFAR-10 dataset, it can be seen that the trainings with all BN layers removed suffer from a significant convergence slow down, and  the corresponding optimization trajectories obey the regularity principle with a very small $\gamma$. 
%On the other hand, the trainings that keep the first BN layer in each block converge as fast as those that keep all the BN layers, and their optimization trajectories obey the regularity principle with a large $\gamma$. These empirical observations corroborate \Cref{thm: convergence}. 
\begin{figure}[htbp]%[bth]
	\vspace{-2mm}
	\centering
	\includegraphics[width=0.24\textwidth,height=0.2\textwidth]{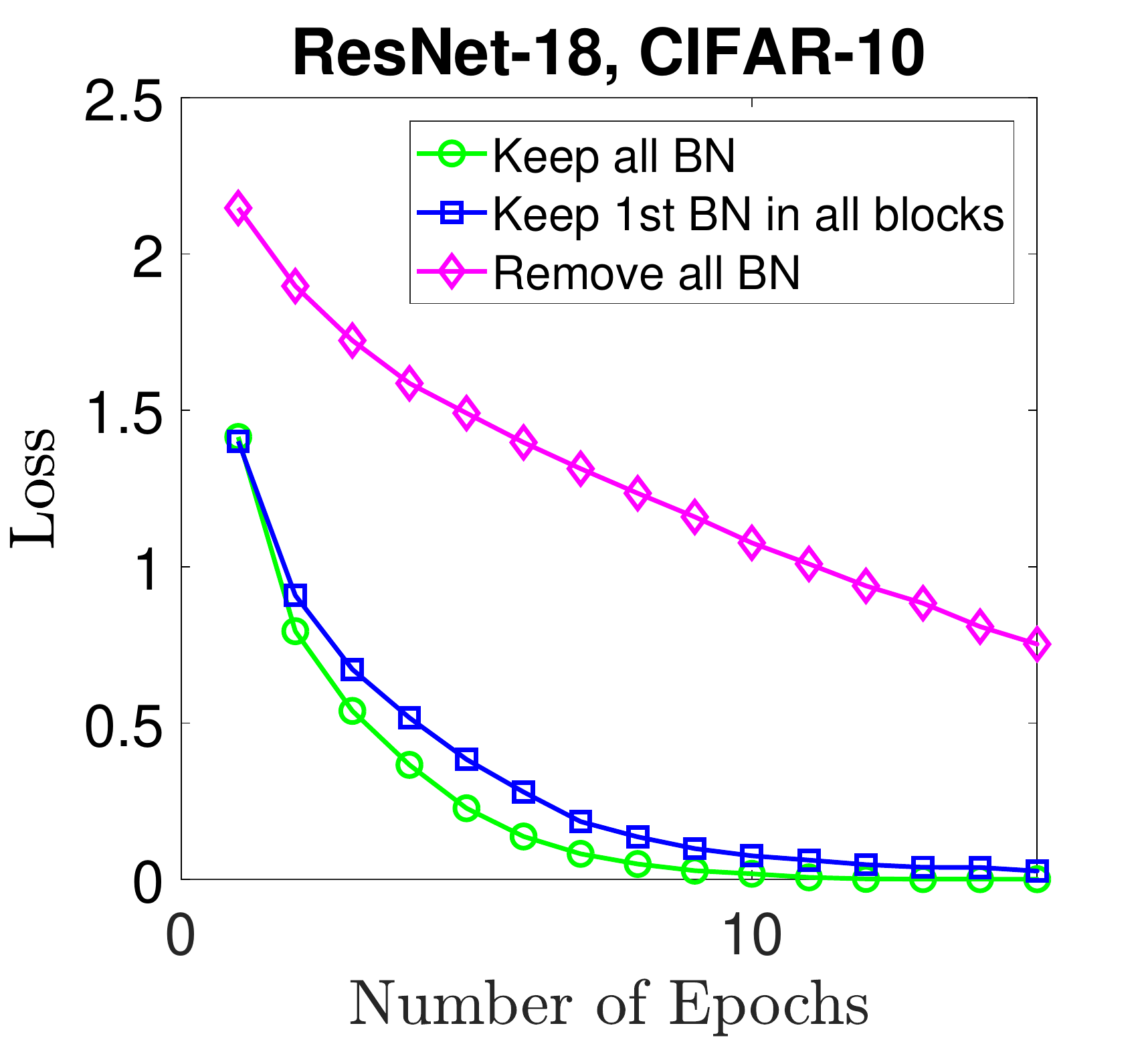}
	\includegraphics[width=0.24\textwidth,height=0.2\textwidth]{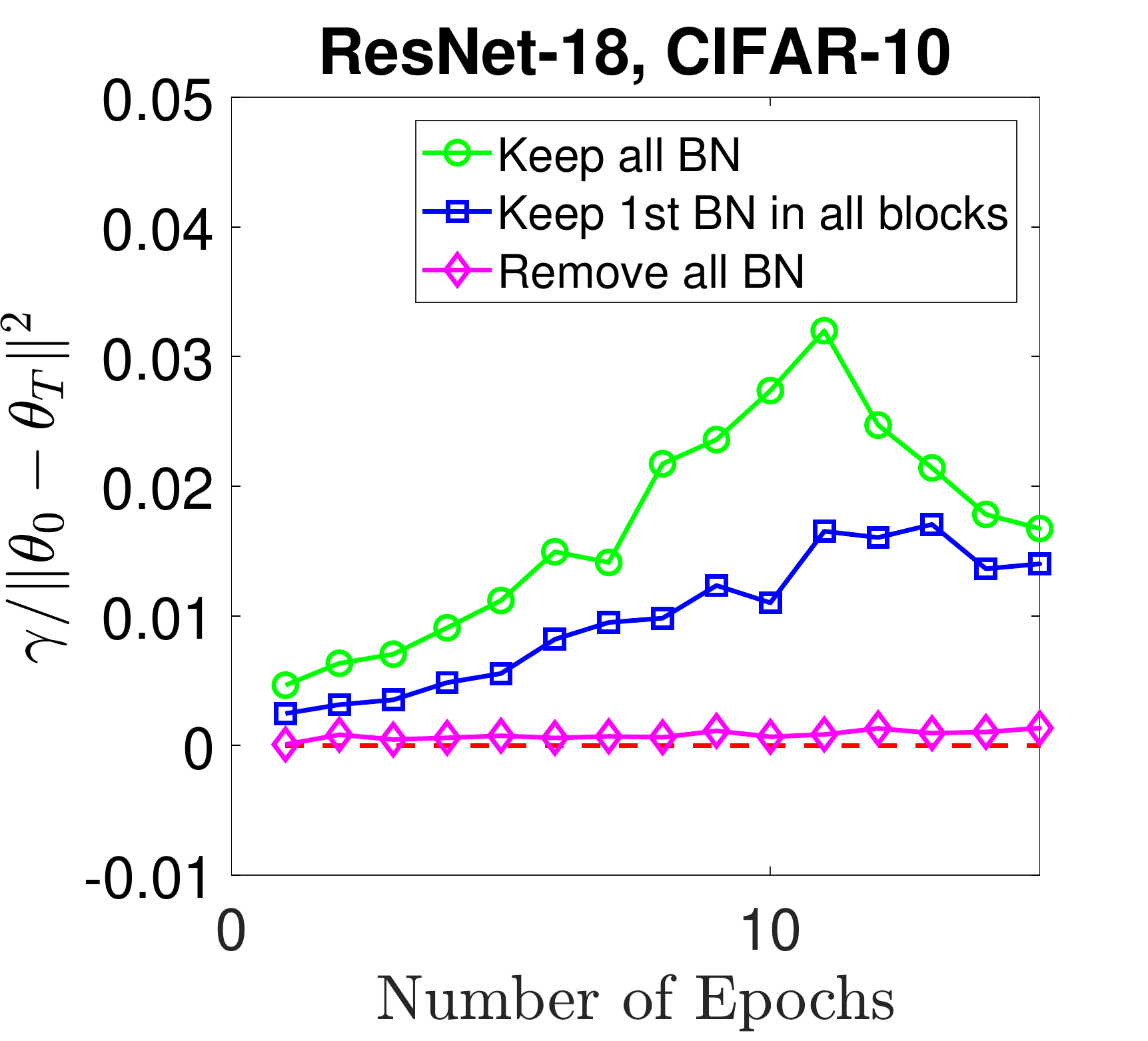}
	\includegraphics[width=0.24\textwidth,height=0.2\textwidth]{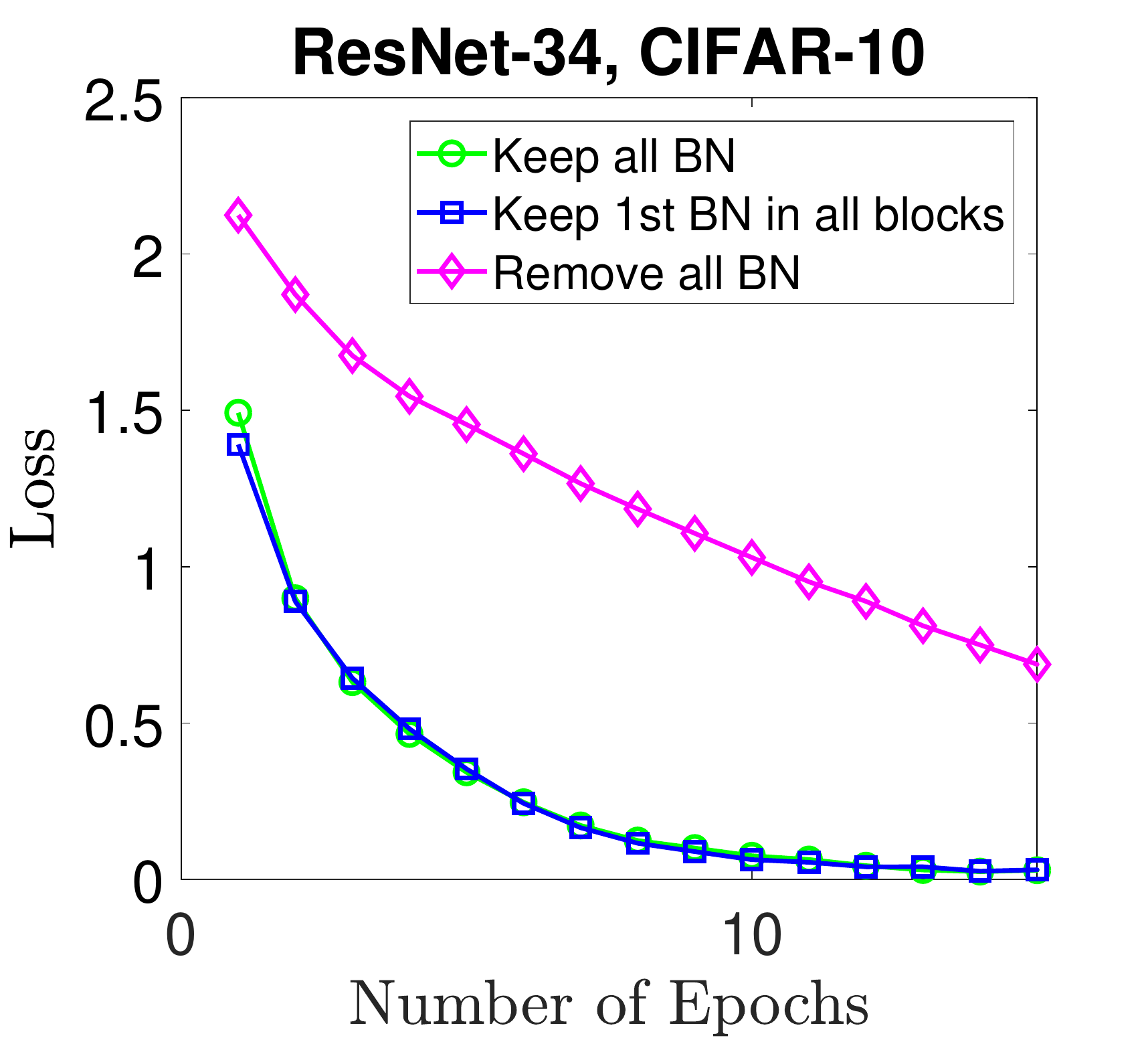}
	\includegraphics[width=0.24\textwidth,height=0.2\textwidth]{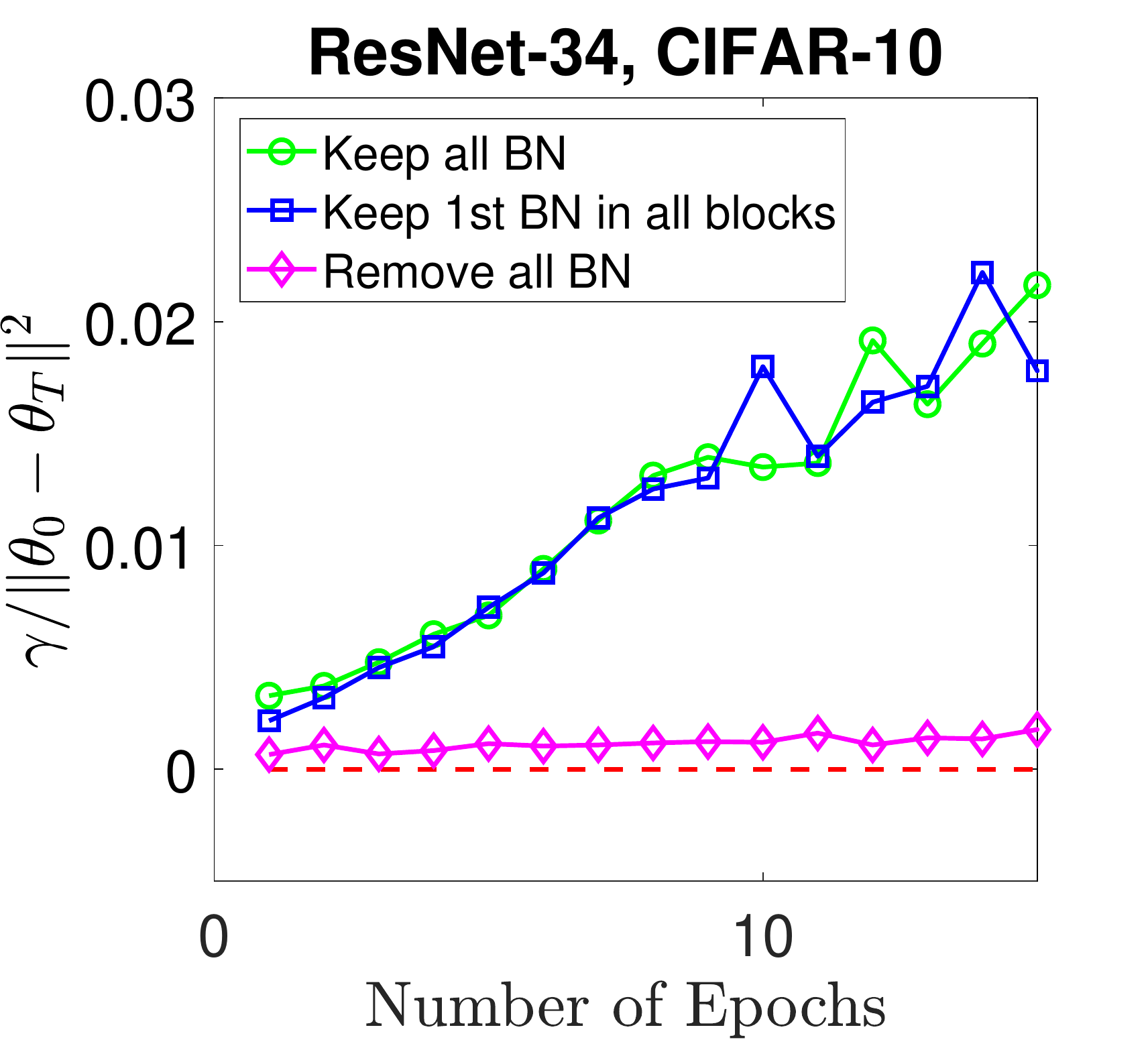}
	\vspace{-6mm}
	\caption{\small{Training ResNets with and without batch normalization on CIFAR-10.}}\label{fig: app: 3} 
		\vspace{-2mm}
\end{figure}

%On the CIFAR-100 dataset, it can be seen that the trainings with all BN layers removed suffer from a significant convergence slow down, and  the corresponding optimization trajectories obey the regularity principle with a very small $\gamma$. 
%On the other hand, the trainings that keep all the BN layers are slightly  faster than those that keep the first BN layers, and their optimization trajectories obey the regularity principle with the largest $\gamma$. These empirical observations corroborate the theoretical implication of the regularity principle in \Cref{thm: convergence}. 

\begin{figure}[htbp]%[bth]
	\vspace{-2mm}
	\centering
	\includegraphics[width=0.24\textwidth,height=0.2\textwidth]{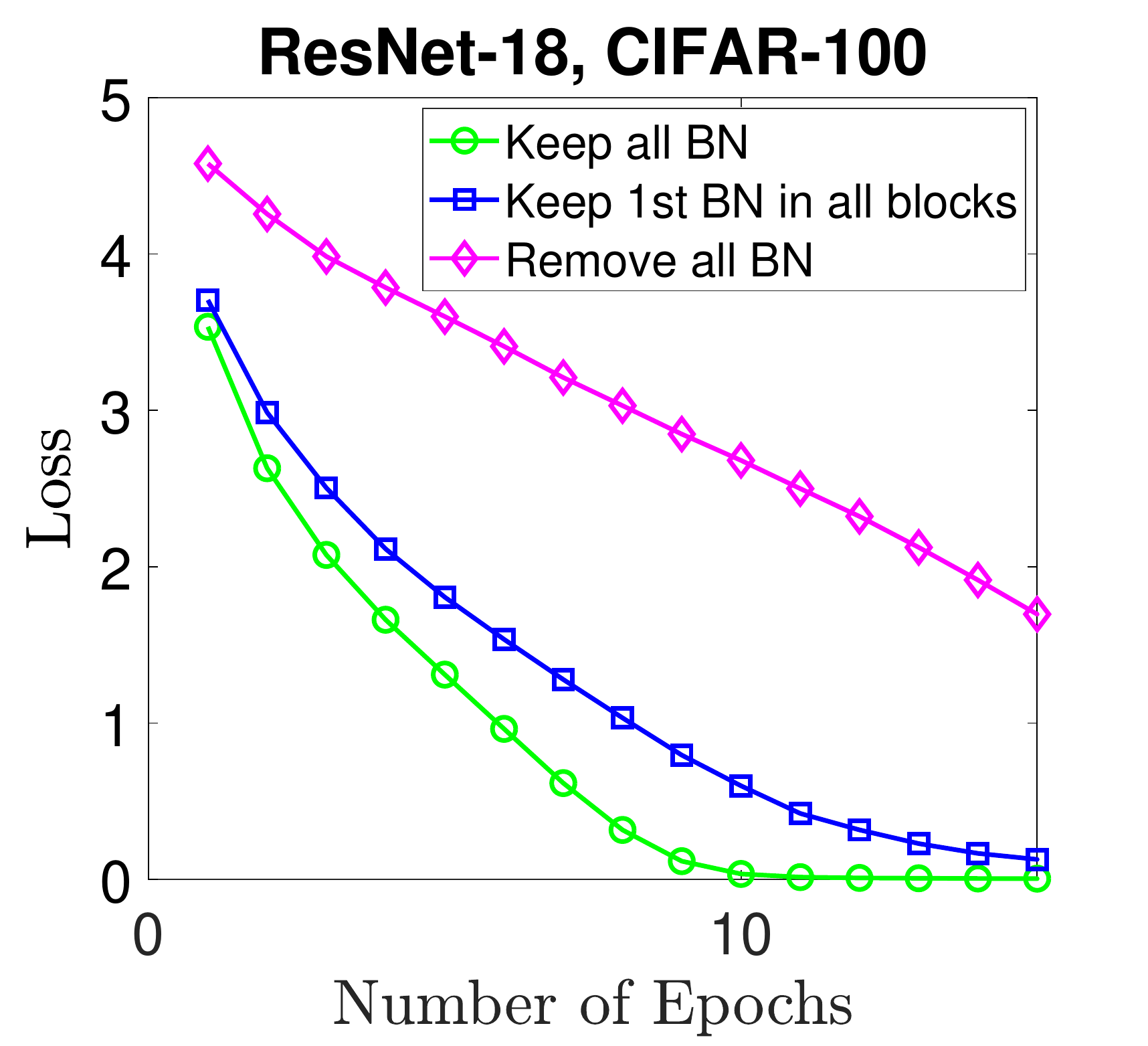}
	\includegraphics[width=0.24\textwidth,height=0.2\textwidth]{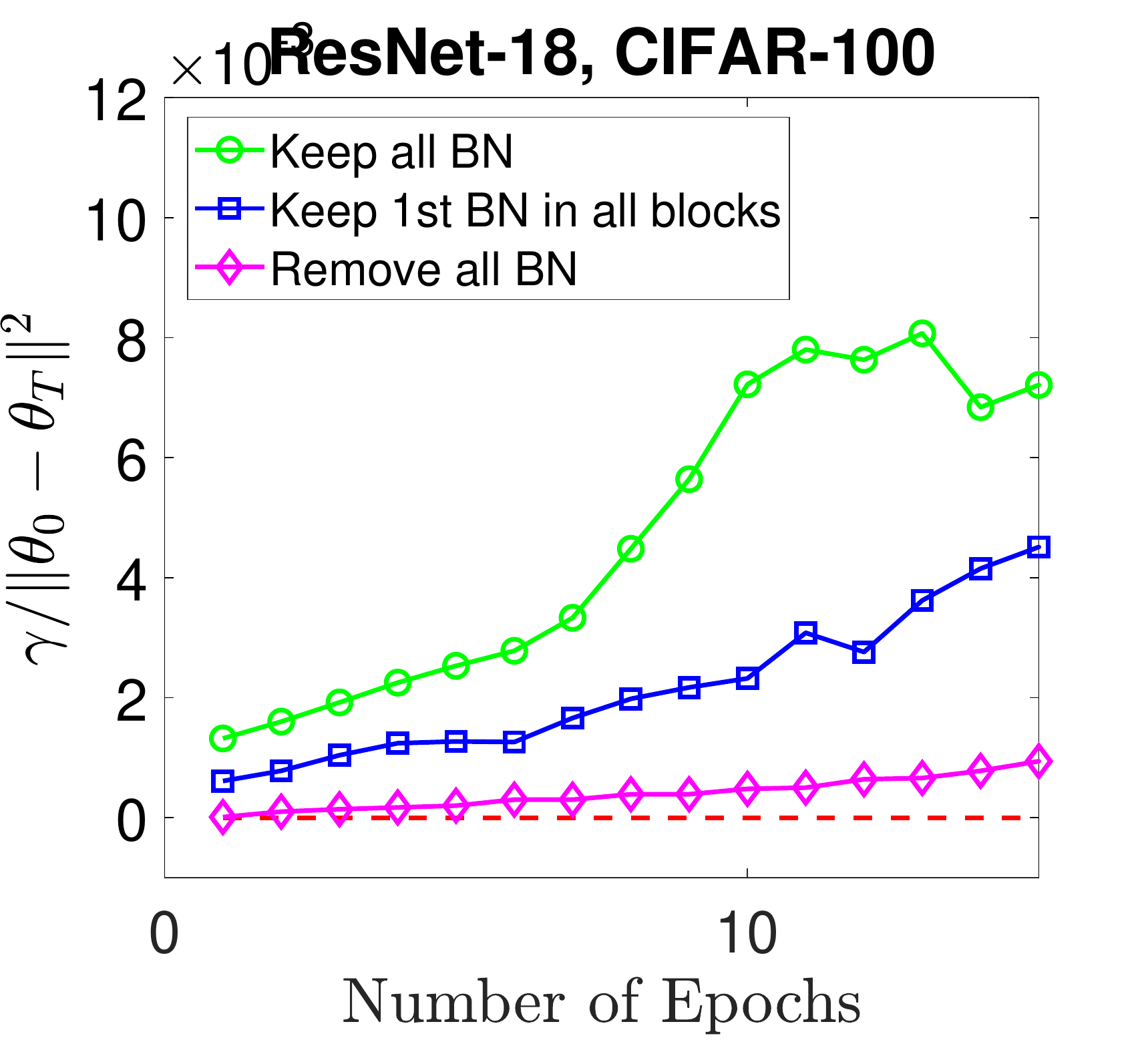}
	\includegraphics[width=0.24\textwidth,height=0.2\textwidth]{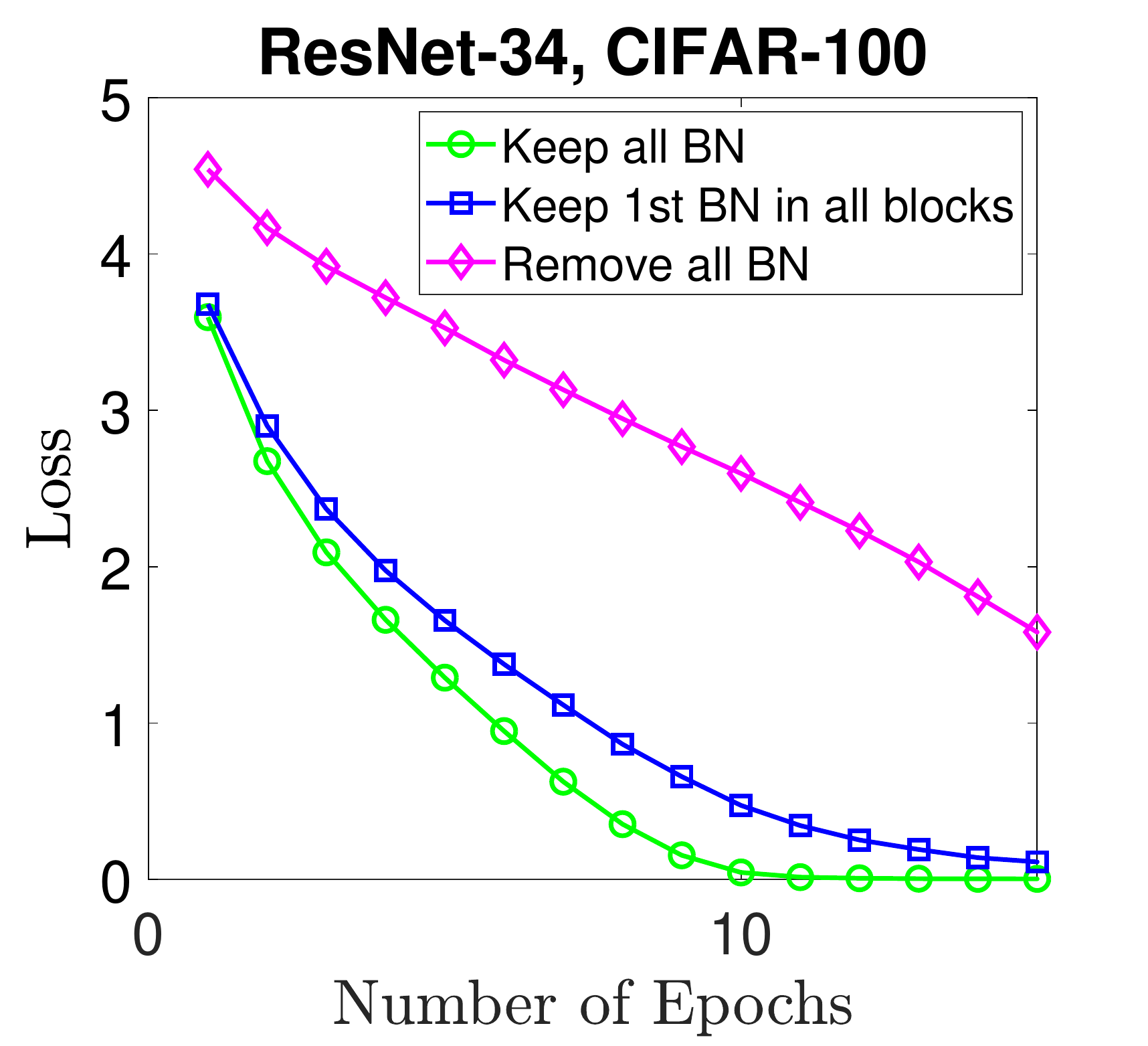}
	\includegraphics[width=0.24\textwidth,height=0.2\textwidth]{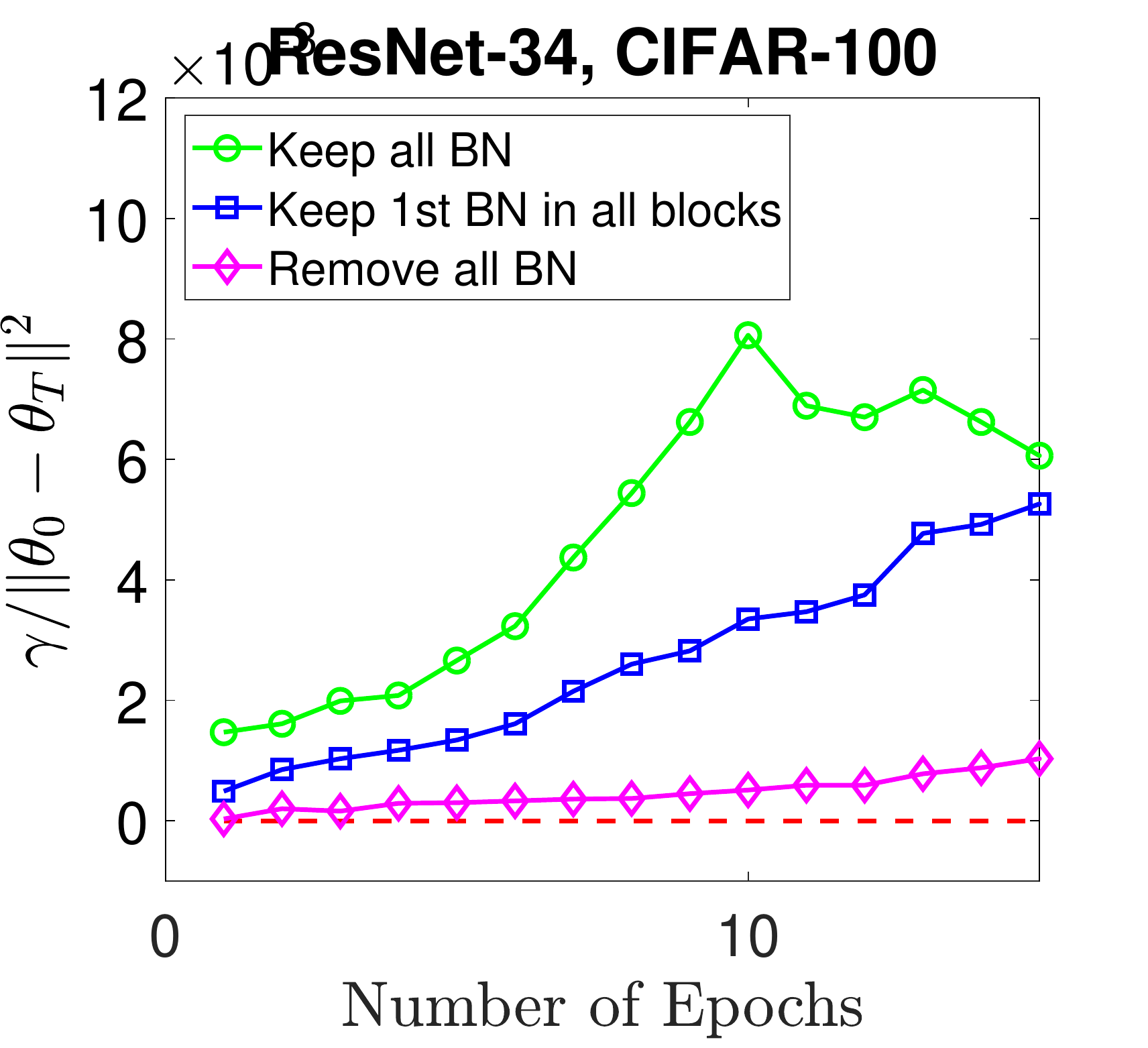}
	\vspace{-6mm}
	\caption{\small{Training ResNets with and without batch normalization on CIFAR-100.}}\label{fig: app: 4} 
		\vspace{-4mm}
\end{figure}

\newpage

\subsection{Effect of Skip-connection on Regularity Principle}\label{sec: app: 3}
%This subsection presents the additional experiments on the effect of skip-connection on the regularity Principle.
\begin{figure}[htbp]%[bth]
	\vspace{-2mm}
	\centering
	\includegraphics[width=0.24\textwidth,height=0.2\textwidth]{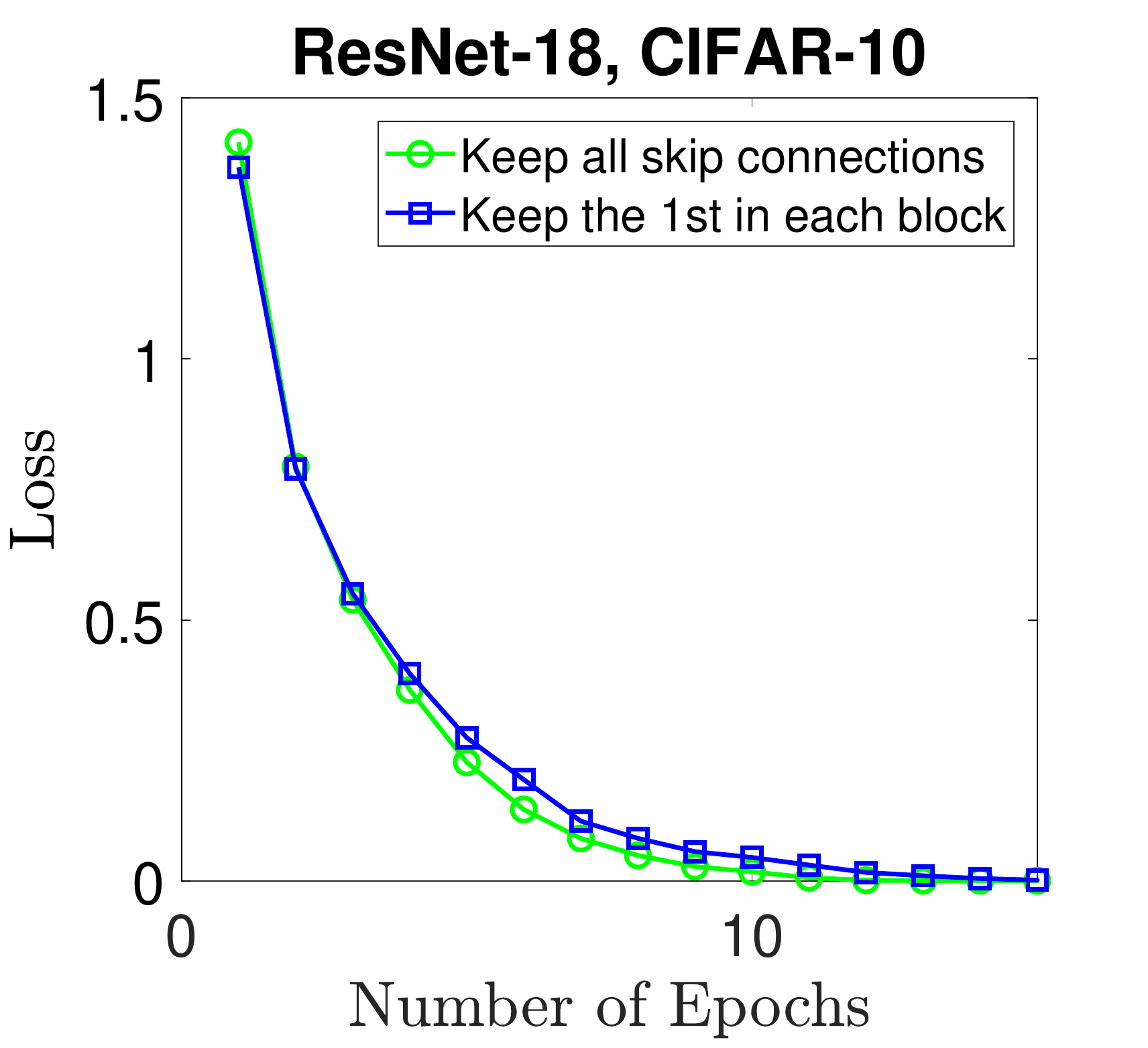}
	\includegraphics[width=0.24\textwidth,height=0.2\textwidth]{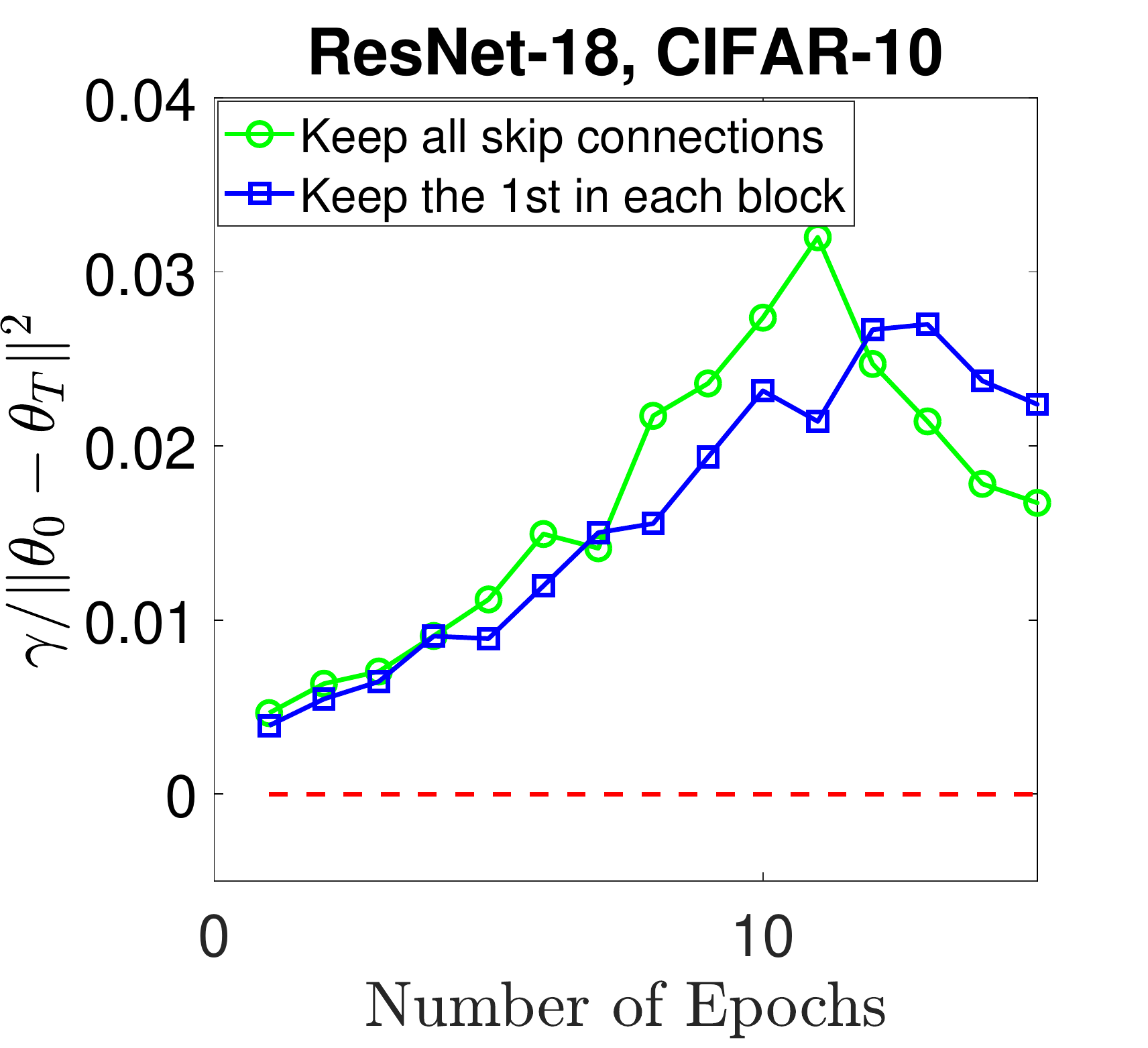}
	\includegraphics[width=0.24\textwidth,height=0.2\textwidth]{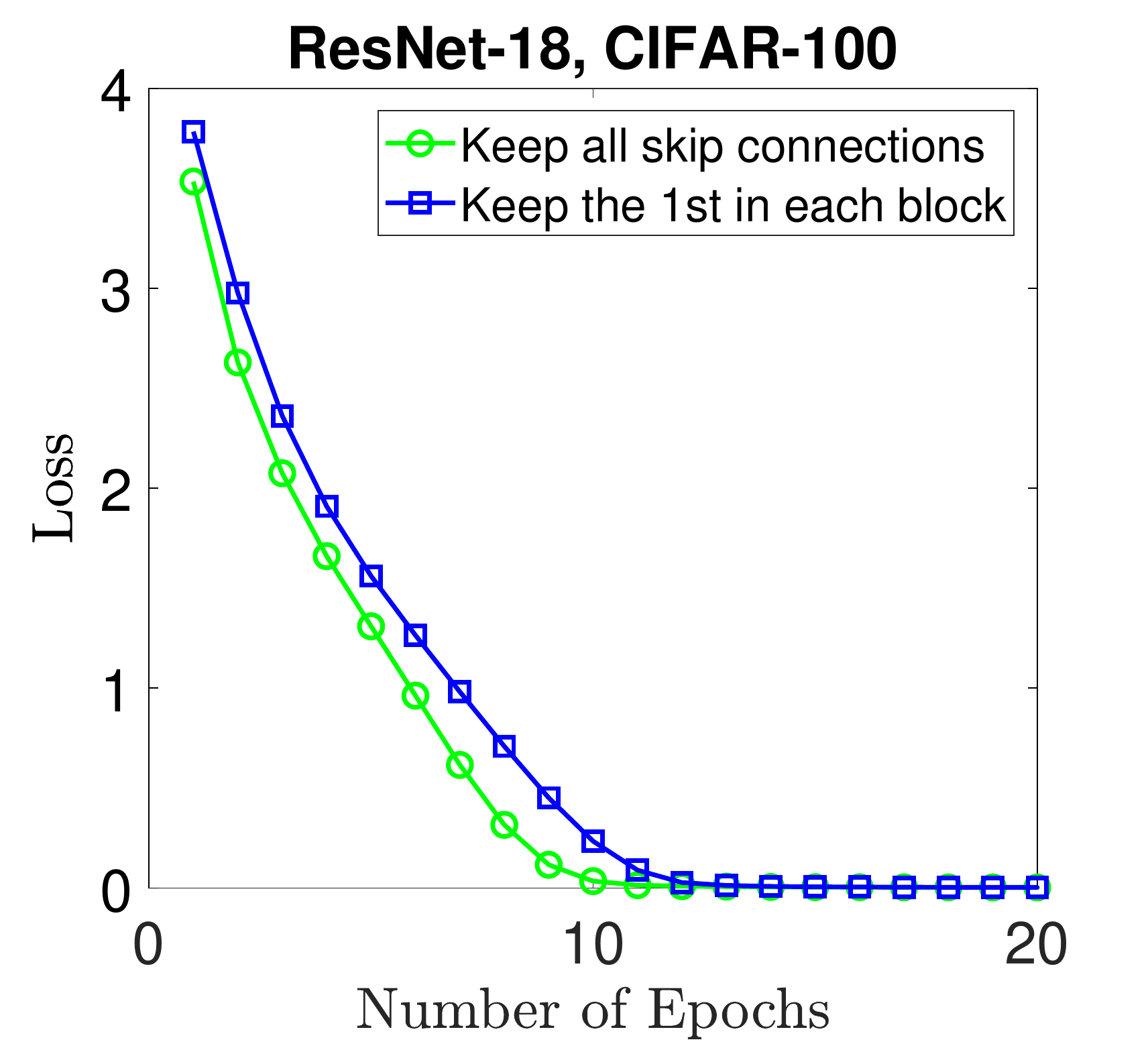}
	\includegraphics[width=0.24\textwidth,height=0.2\textwidth]{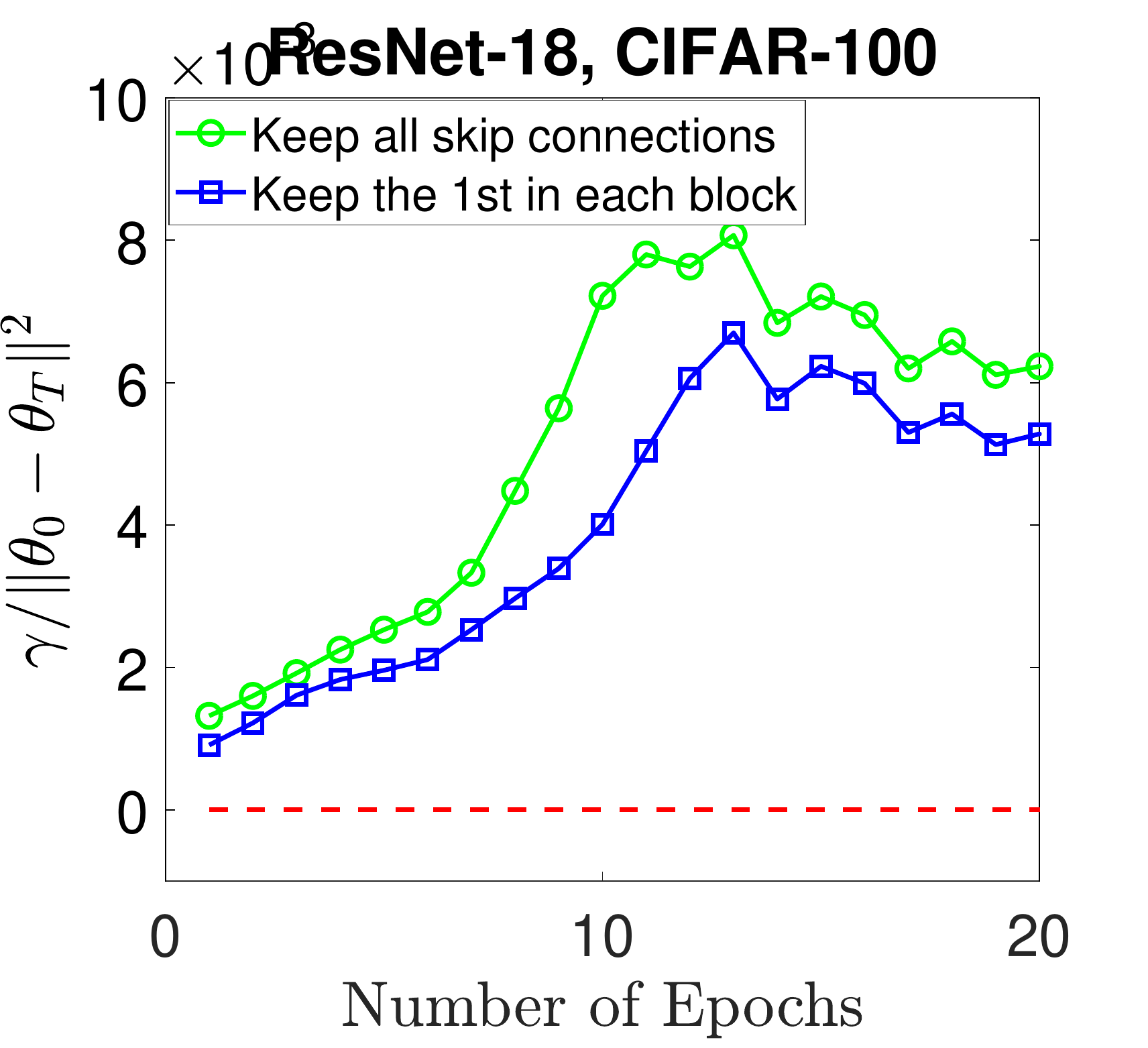}
	\vspace{-6mm}
	\caption{\small{Training ResNet-18 with and without skip-connections.}}\label{fig: app: 5} 
	\vspace{-4mm}
\end{figure}

\newpage
\subsection{Experiments on Optimization-level Training Techniques}\label{sec: app: 4}
%This subsection presents the additional experiments on optimization-level training techniques.
\begin{figure}[htbp]%[bth]
	\vspace{-2mm}
	\centering
	\includegraphics[width=0.24\textwidth,height=0.2\textwidth]{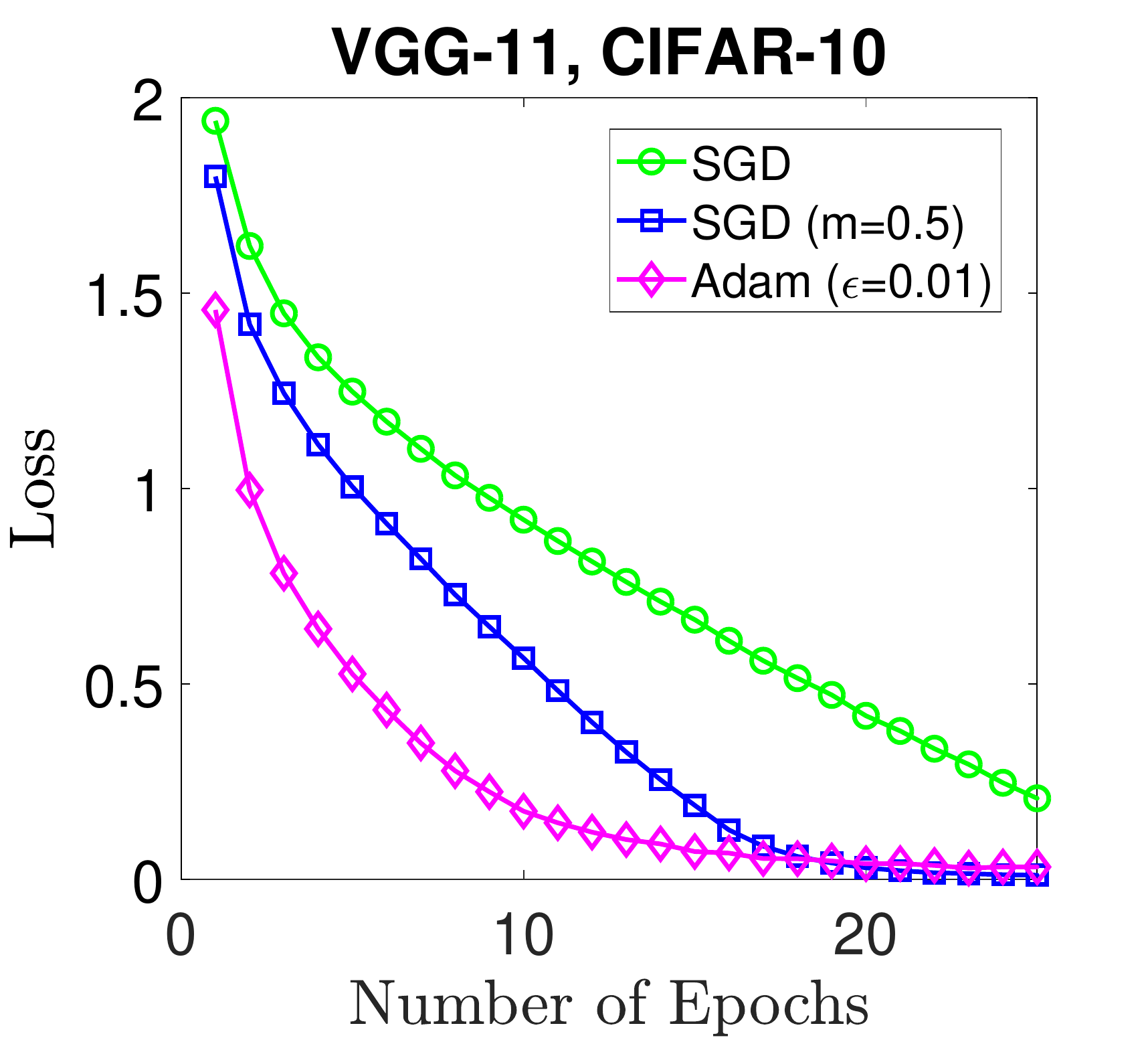}
	\includegraphics[width=0.24\textwidth,height=0.2\textwidth]{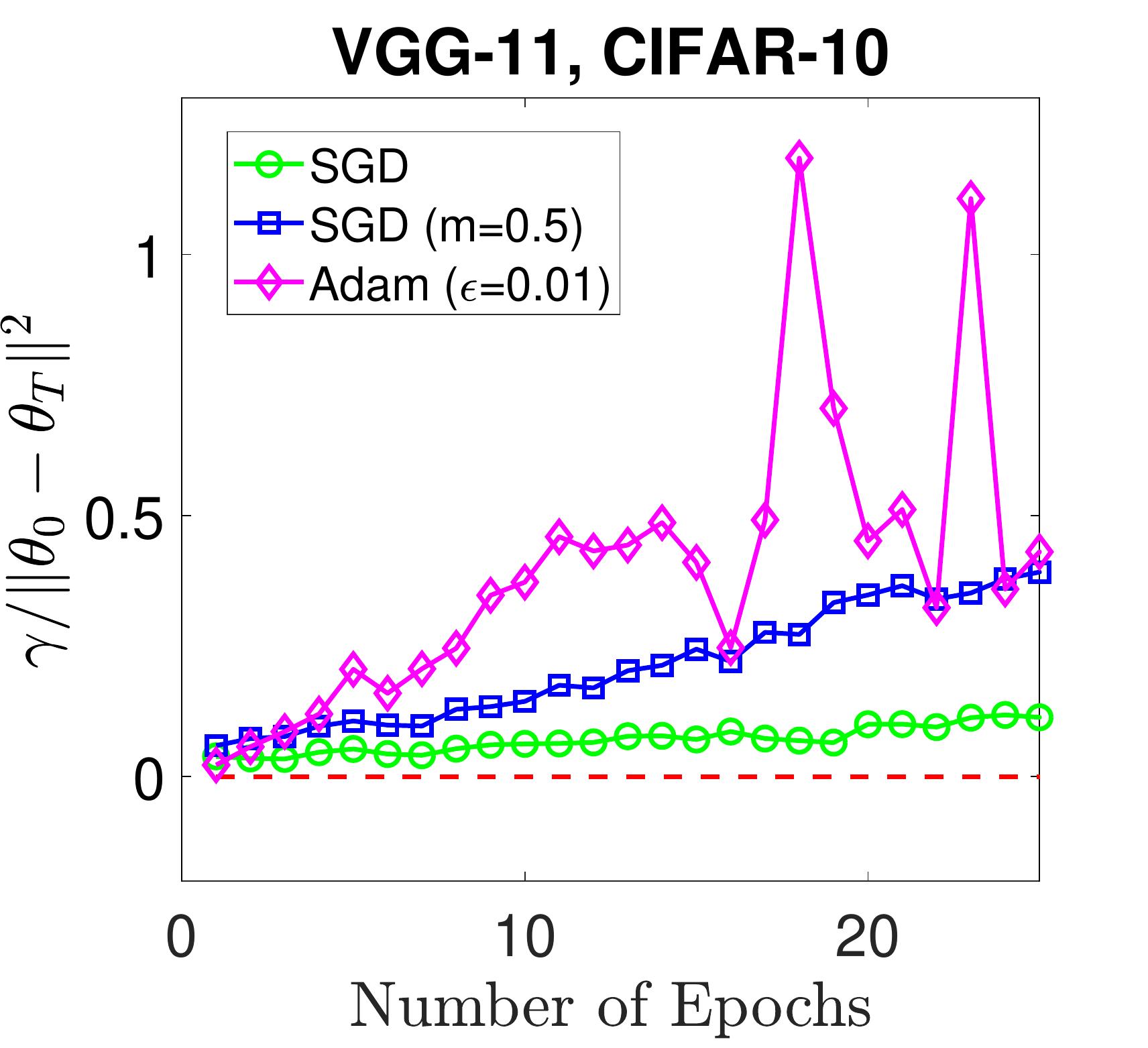}
	\includegraphics[width=0.24\textwidth,height=0.2\textwidth]{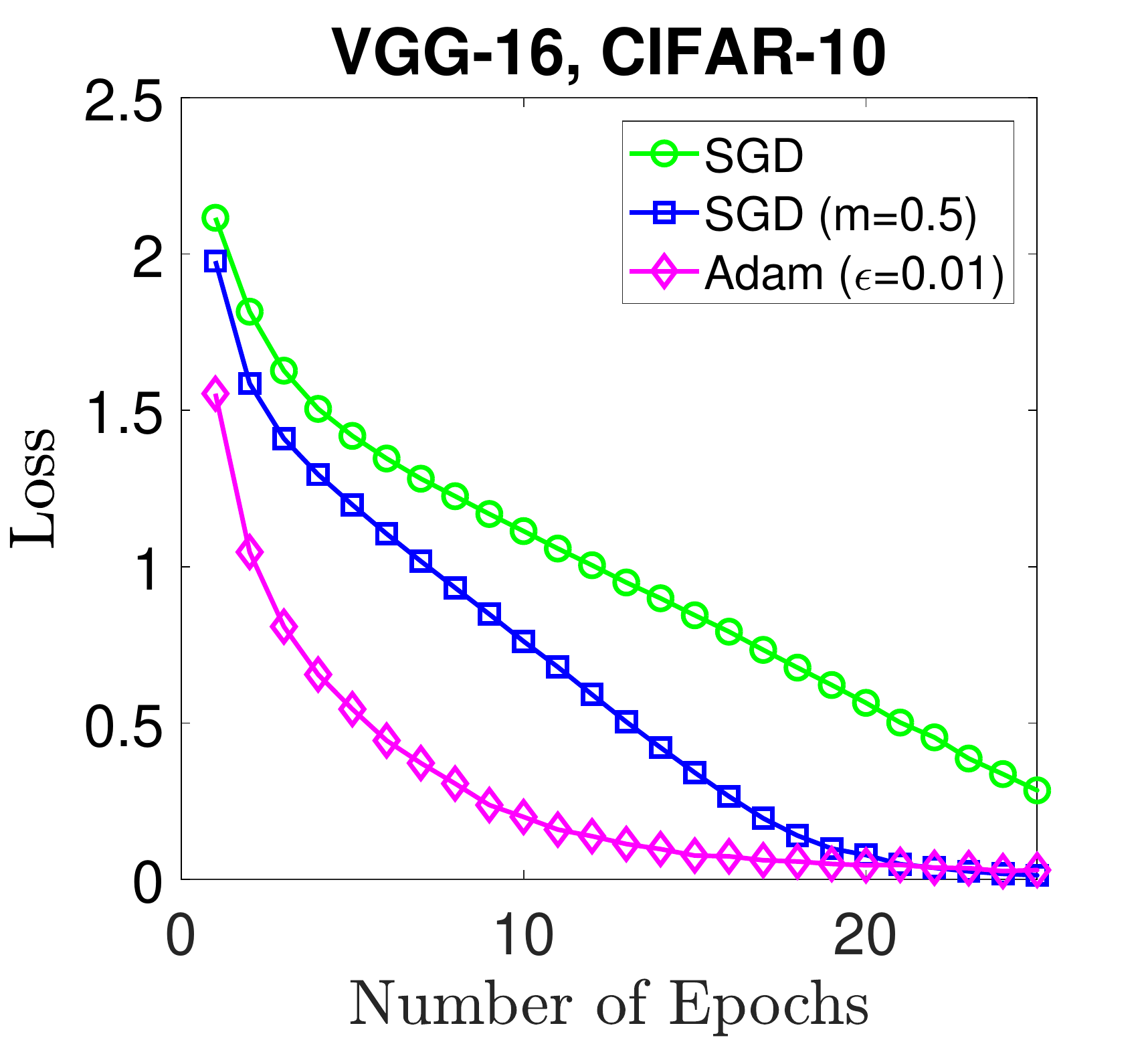}
	\includegraphics[width=0.24\textwidth,height=0.2\textwidth]{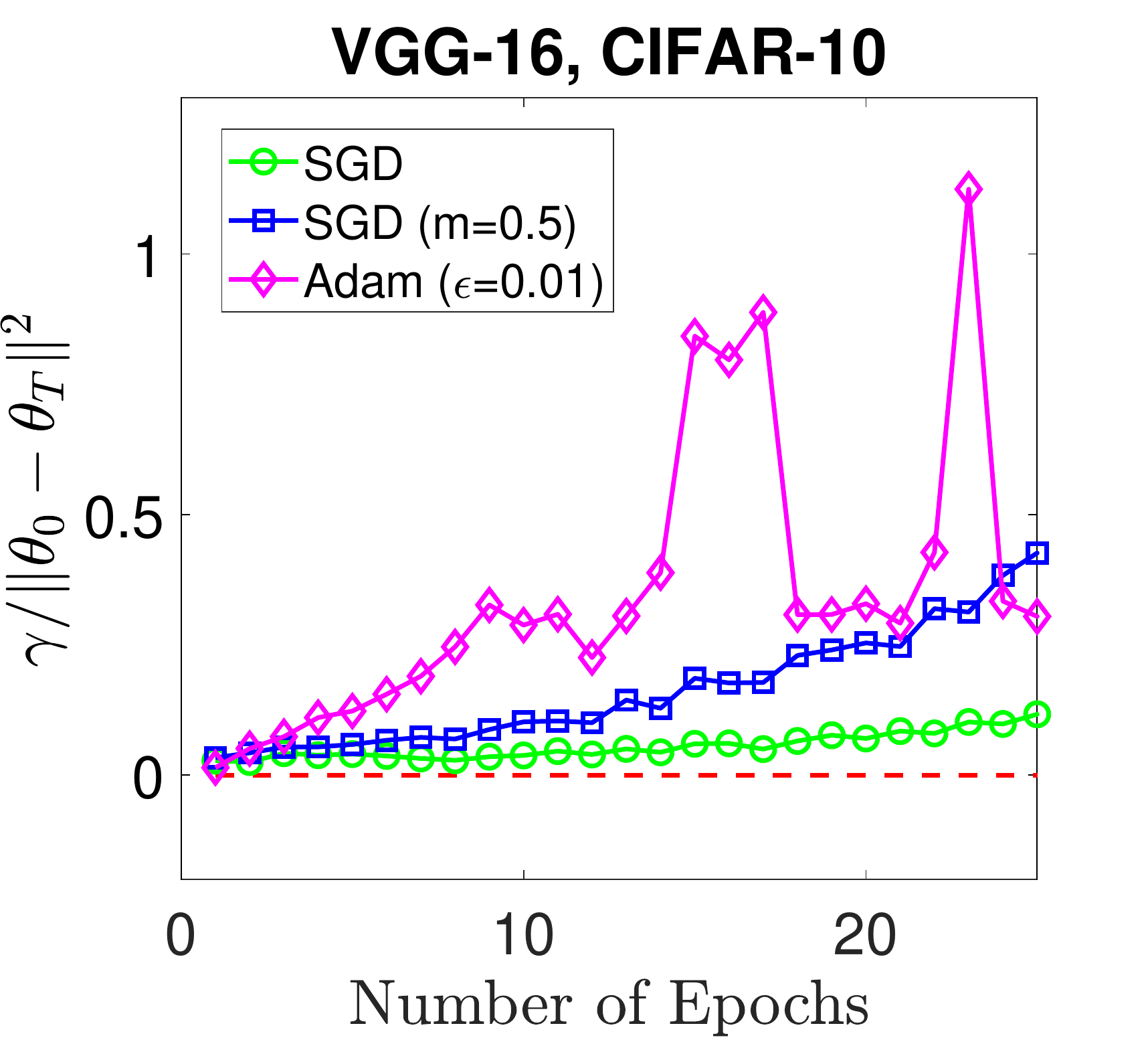}
	\vspace{-6mm}
	\caption{\small{Training VGGs with different optimizers on CIFAR-10.}}\label{fig: app: 6} 
\end{figure}

\begin{figure}[htbp]%[bth]
\vspace{-4mm}
	\centering
	\includegraphics[width=0.24\textwidth,height=0.2\textwidth]{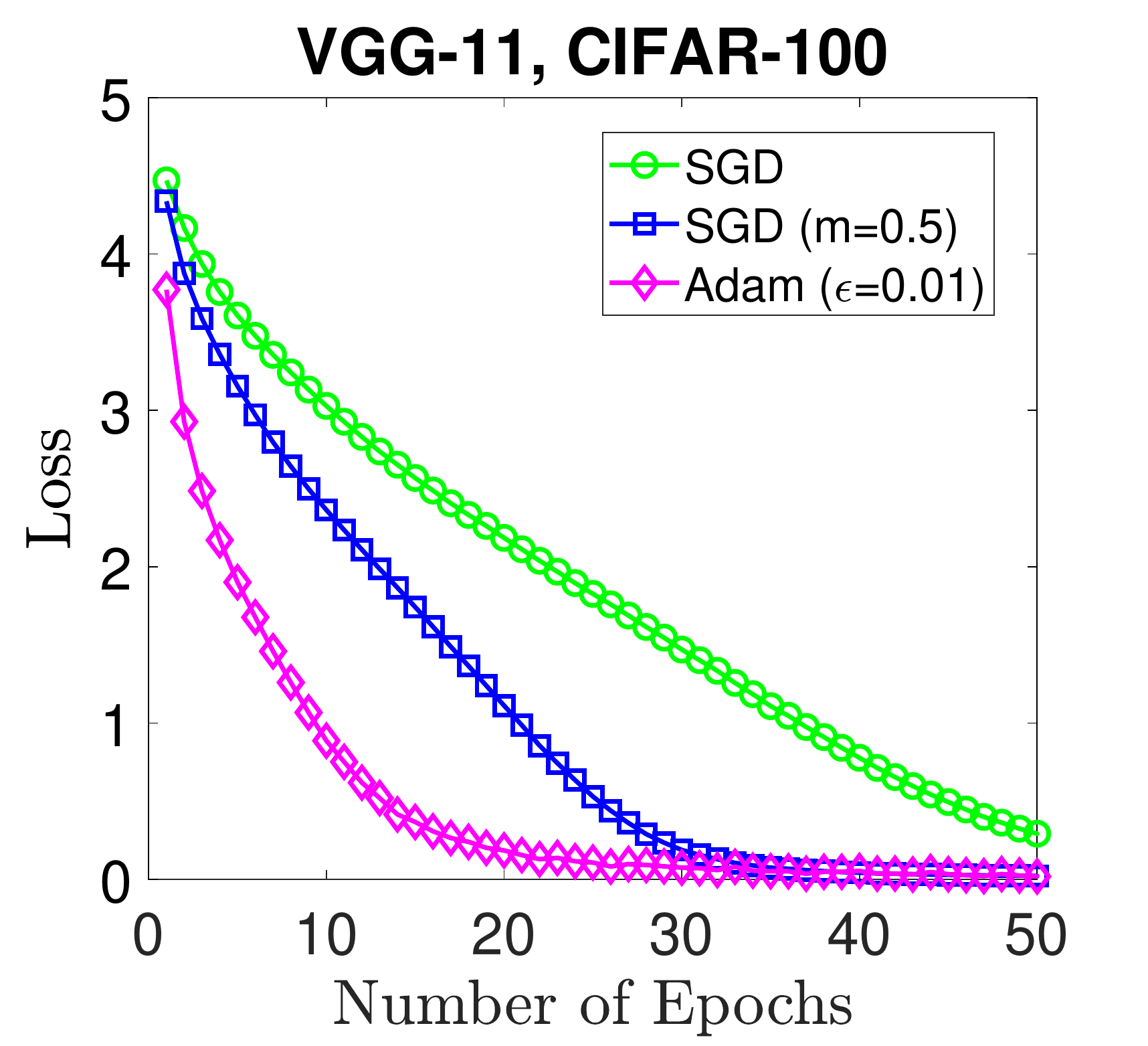}
	\includegraphics[width=0.24\textwidth,height=0.2\textwidth]{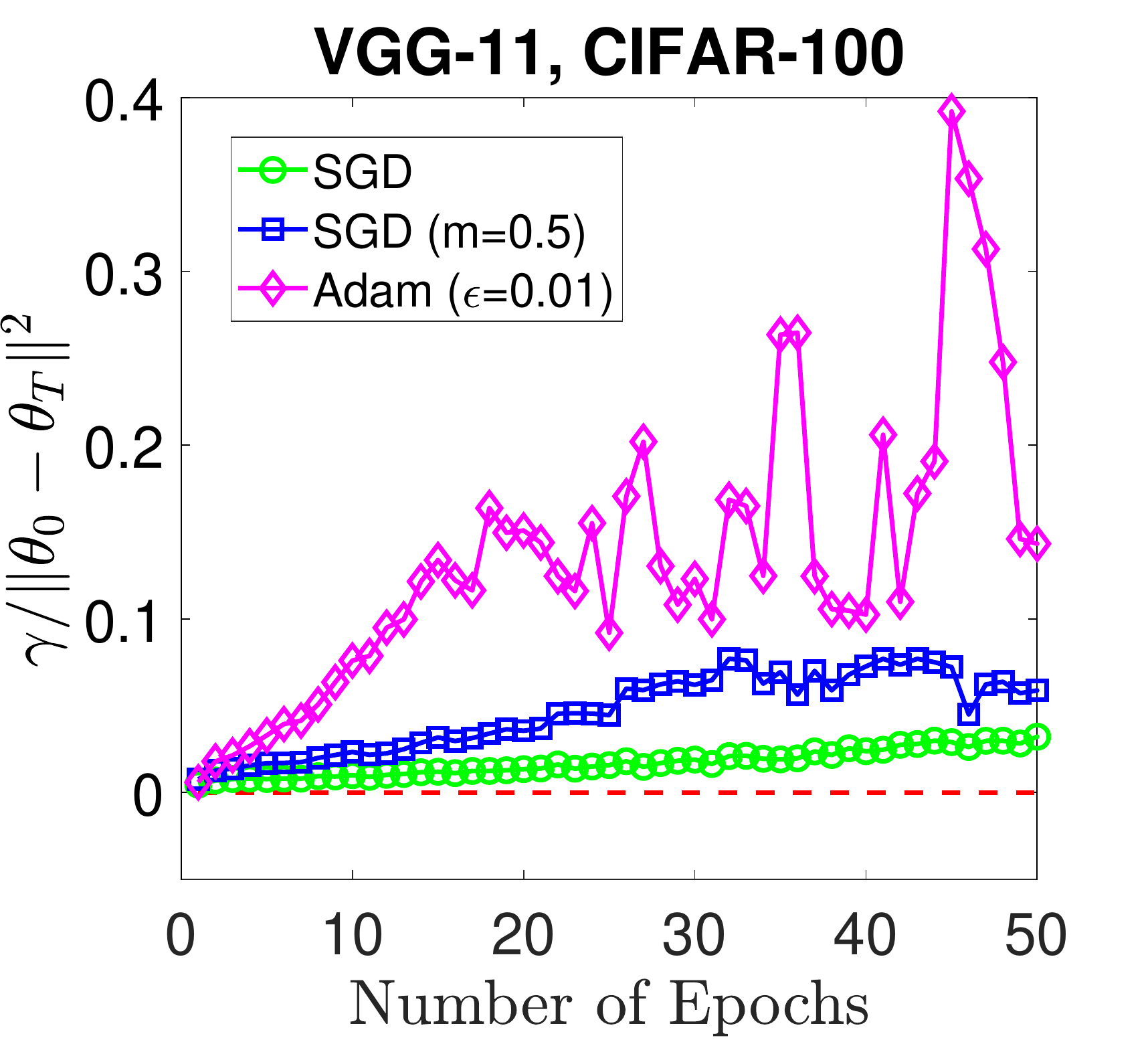}
	\includegraphics[width=0.24\textwidth,height=0.2\textwidth]{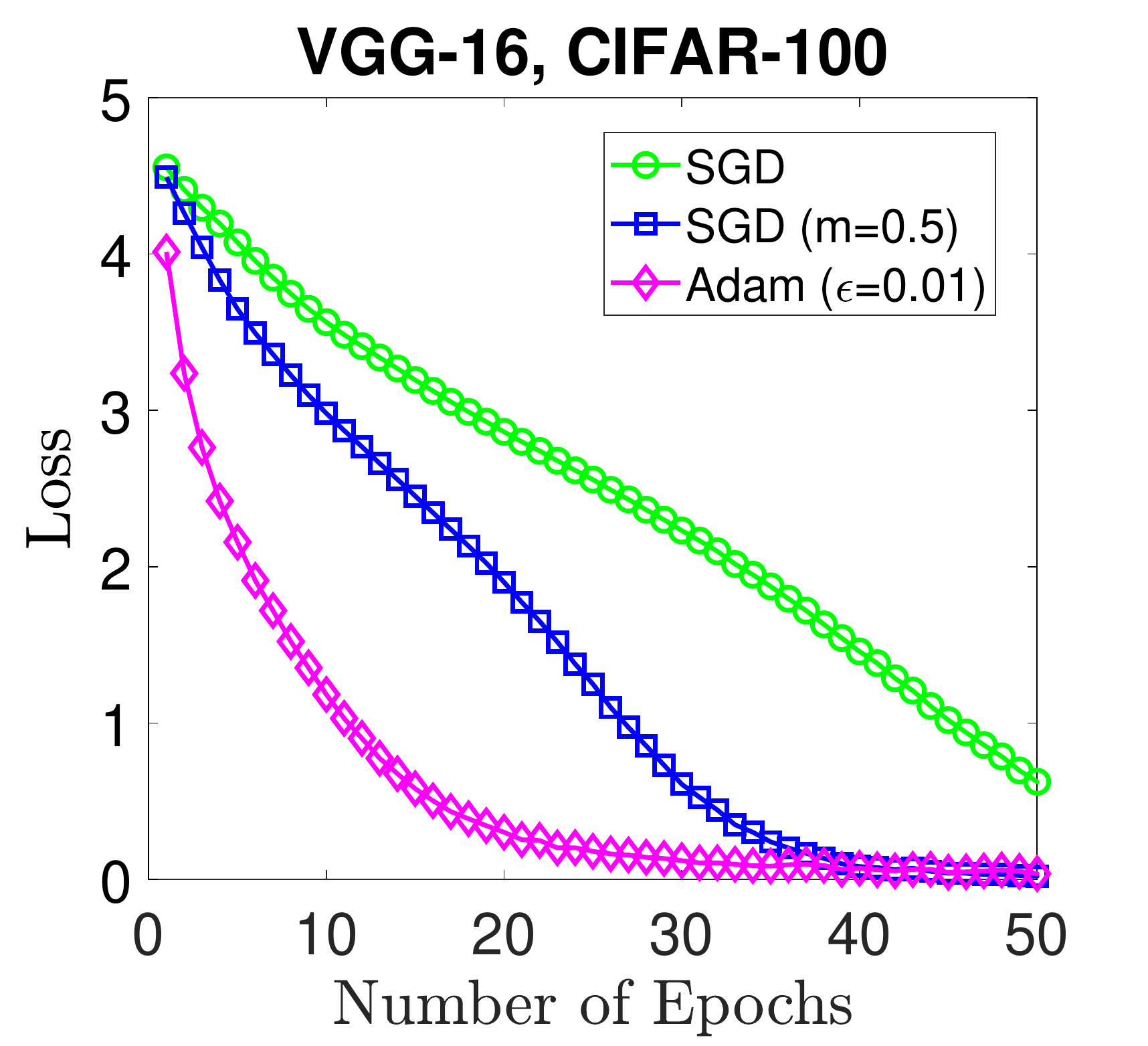}
	\includegraphics[width=0.24\textwidth,height=0.2\textwidth]{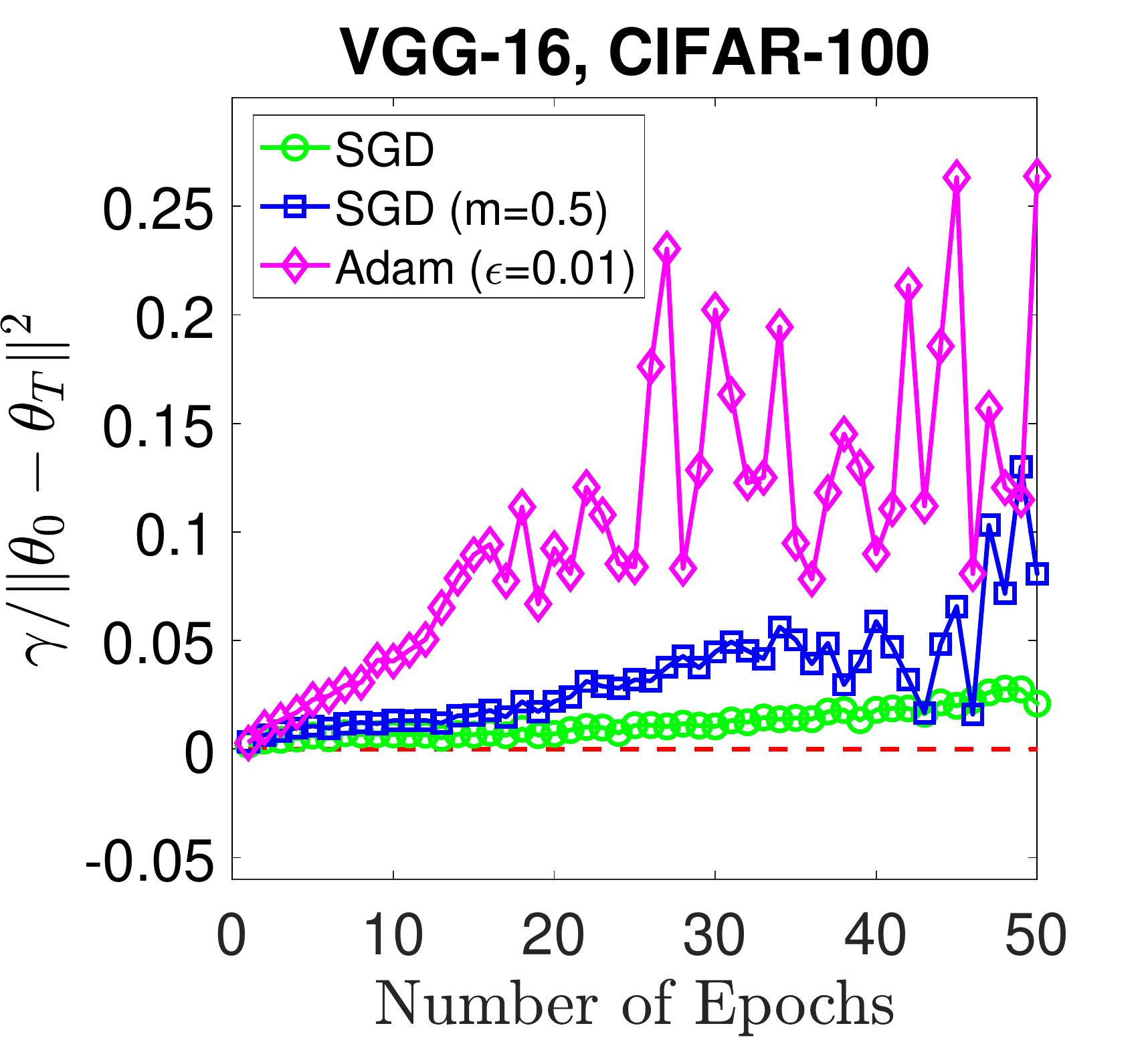}
\vspace{-6mm}
	\caption{\small{Training VGGs with different optimizers on CIFAR-100.}}\label{fig: app: 7} 
	%	\vspace{-2mm}
\end{figure}

\end{document}